\documentclass[pdflatex,sn-nature]{sn-jnl}

\usepackage{graphicx}%
\usepackage{multirow}%
\usepackage{amsmath,amssymb,amsfonts}%
\usepackage{amsthm}%
\usepackage{mathrsfs}%
\usepackage[title]{appendix}%
\usepackage{xcolor}%
\usepackage{textcomp}%
\usepackage{manyfoot}%
\usepackage{booktabs}%
\usepackage{algorithm}%
\usepackage{algorithmicx}%
\usepackage{algpseudocode}%
\usepackage{listings}%
\usepackage{hyperref}

\usepackage[T1]{fontenc}
\usepackage[misc]{ifsym}

\usepackage{siunitx}
\usepackage{bbm,mathtools,bm,pifont}
\usepackage{enumitem}
\usepackage{subcaption}
\usepackage{comment}
\usepackage{multirow}
\usepackage{tabularx} 
\usepackage{array} 
\usepackage{dsfont}
\usepackage{lipsum}
\usepackage[export]{adjustbox}

\usepackage{dirtree}

\usepackage{filecontents}  

\theoremstyle{thmstyleone}%
\theoremstyle{thmstyletwo}%

\theoremstyle{thmstylethree}%

\begin{document}

\title[Article Title]{SteelDS: A High-Resolution Video Dataset of E40 Steel Scrap for Object Detection and Instance Segmentation}


\author*[1]{\fnm{Melanie} \sur{Neubauer}}\email{melanie.neubauer@unileoben.ac.at}

\author[1]{\fnm{Christian} \sur{Rauch}}\email{christian.rauch@unileoben.ac.at}

\author[2]{\fnm{Gerald} \sur{Koinig}}\email{gerald.koinig@unileoben.ac.at}

\author[2]{\fnm{Alexia} \sur{Tischberger-Aldrian}}\email{alexia.tischberger-aldrian@unileoben.ac.at}

\author[2]{\fnm{Roland} \sur{Pomberger}}\email{roland.pomberger@unileoben.ac.at}

\author[1]{\fnm{Elmar} \sur{Rueckert}}\email{rueckert@unileoben.ac.at}

\affil*[1]{\orgdiv{Chair of Cyber-Physical-Systems}, \orgname{Technical University of Leoben}, \orgaddress{\street{Franz Josef-Straße 18}, \city{Leoben}, \postcode{8700}, \state{Styria}, \country{Austria}}}

\affil[2]{\orgdiv{Chair of Waste Processing Technology and Waste Management}, \orgname{Technical University of Leoben}, \orgaddress{\street{Franz Josef-Straße 18}, \city{Leoben}, \postcode{8700}, \state{Styria}, \country{Austria}}}


\abstract{
This dataset provides high-resolution, annotated video sequences of shredded E40-grade steel and copper scrap on a conveyor belt. 
Captured in a controlled laboratory environment, the data reflects the industrial post-magnetic sorting stage, where manual intervention is typically required to remove copper contaminants. 
The dataset comprises 24,297 labeled frames across five subsets, featuring 396 steel and 101 copper objects categorized by size. 
It supports the development of machine learning models for material classification, object detection, and instance segmentation. 
Variations in object spacing and density are included to simulate realistic industrial sorting conditions. 
Ground truth annotations include pixel-wise segmentation masks and material classes. 
This dataset serves as a benchmark for evaluating automated sorting algorithms aiming to identify copper impurities within complex, heterogeneous steel scrap streams.}


\maketitle

\section*{Background \& Summary}
Recycling metals, such as steel and copper, is critical for environmental sustainability and resource conservation. 
Automated sorting in recycling facilities depends on accurate material classification algorithms, which require realistic, annotated datasets for training and evaluation. 
Traditionally, industrial facilities rely on Sensor-Based Sorting (SBS) architectures equipped with highly specialized, capital-intensive modalities like X-ray Transmission (XRT) or Hyperspectral Imaging (HSI)~\cite{maier2024survey}. 
While highly effective, these physical sensor arrays are economically prohibitive for widespread deployment and offer lower spatial resolutions. 
In contrast, standard vision-based sorting approaches leveraging deep learning require only a single, low-cost RGB camera system~\cite{maier2024survey, han2021automatedMetal, SARSWAT2024107651}. 
By shifting material discrimination from expensive sensing hardware to advanced spatial algorithms, computer vision presents an accessible alternative. 
However, there is a lack of specialized resources tailored to industrial metal scrap recycling, particularly for steel-copper separation. 
This scarcity is fundamentally driven by the fact that industrial entities restrict public access to their operational data, preventing the publication of proprietary domain-specific datasets.

Waste datasets, such as TACO~\cite{proencca2020taco}, TrashNet~\cite{yang2016trashnet}, and ZeroWaste~\cite{zerowaste}, primarily focus on household or municipal solid waste. 
These datasets feature macroscopic, mostly intact objects like plastic bottles, paper, or consumer packaging. 
Recent efforts have expanded into domains with heavier objects, providing benchmark datasets for the segmentation of construction and demolition waste (CDW) in cluttered environments~\cite{sirimewan2025benchmark}. 
While these datasets address industrial materials, they primarily represent macroscopic, structurally intact waste rather than mechanically shredded, micro-scale scrap.
Furthermore, recent studies have explored assistive computer vision for estimating non-metallic content in bulk scrap metal~\cite{storonkin2026images}, though these approaches focus on image-level bulk assessment. 
While early efforts successfully leveraged deep networks to classify and detect shredded copper-containing metal scrap fractions on sorting streams~\cite{CHEN2021266,KOINIG2024520, koinig2025cnn}, they relied on coarse bounding boxes or image-level annotations. 
Such methods fail to delineate the highly irregular boundaries and fine strands typical of stochastically deformed scrap. 
This highlights the critical gap that SteelDS addresses by providing a publicly available, instance-level segmentation benchmark.

Industrial E40-grade steel shredder scrap~\cite{jrc_steel_scrap_2010} (a standard European classification for fragmented old steel scrap primarily sourced from end-of-life vehicles and white goods) undergoes aggressive mechanical processing and magnetic pre-sorting. 
This process results in a highly specific material stream characterized by severe plastic deformation, torn edges, and structural fragmentation~\cite{aboussouan1999steel}. 

The remaining impurities in this stream, specifically embedded or partially obscured copper components, typically necessitate manual post-sorting to meet stringent quality constraints in secondary steelmaking. 
Hills of mixed scrap on scrapyards, as seen in some prior visual datasets, only partially reflect the industrial complexity of this post-magnetic-sorting stage. 

DOES~\cite{schafer2023does} recently introduced multimodal image data of non-alloyed European scrap classes with varying oxidation states for variation in visual appearance. 
Instead of segmentation masks, they applied random crops to images without specific semantic object borders. 
Furthermore, real-world scrap streams present an open-world challenge. 
The material contains unpredictable, heavily deformed shapes, ranging from hazardous items~\cite{jurtsch2024machine} to unknown foreign alloys. 
Standard models usually fail when encountering these unexpected objects, misclassifying them into predefined categories with wrong, high confidence.

To address these baseline limitations and the systematic lack of representative data, we present SteelDS: a high-resolution, instance-segmented video and image dataset captured under controlled laboratory conditions. 
Our work focuses on mechanically shredded E40 steel for copper impurity detection and provides manually labeled, pixel-wise segmentation masks for complex perception tasks. 
The dataset targets this specific post-sorting stage, characterized by extreme intra-class variance. 
Due to the stochastic nature of industrial shredders, parts exhibit unique shapes, edges, and deformation patterns, making material classification highly complex.

By providing annotated video sequences with varying object arrangements, density conditions, and natural occlusions, the data supports multiple computer vision tasks, including material classification, object detection, instance segmentation, and multi-object tracking. 
Additionally, by establishing precise pixel-wise baselines of known targets within a heterogeneous stream, SteelDS provides the foundational topological data required to train and validate robust open-world sorting architectures. 
In summary, the contributions of SteelDS include a high-resolution benchmark of mechanically deformed and mixed post-sorting industrial E40 scrap, specifically designed for instance segmentation tasks, distinguishing it from existing municipal and static industrial waste datasets as detailed in Table \ref{tab:related_work_comparison}.

\begin{table*}
\centering
\caption{Comparison of dataset requirements and features between related work and the proposed SteelDS.}
\label{tab:related_work_comparison}
\resizebox{\textwidth}{!}{%
\begin{tabular}{lccccc}
\toprule
\textbf{Dataset} & \textbf{Domain} & \textbf{Material Focus} & \textbf{Material Condition} & \textbf{Data Format} & \textbf{Annotation Type} \\
\midrule
CDW-Seg \cite{sirimewan2025benchmark} & Construction & Demolition waste & Macroscopic & Image & Segmentation \\
TACO \cite{proencca2020taco} & Municipal/Household & General waste & Mostly intact & Image & Segmentation \\
TrashNet \cite{yang2016trashnet} & Municipal/Household & General waste & Mostly intact & Image & Classification \\
ZeroWaste \cite{zerowaste} & Municipal/Industrial & General waste & Macroscopic & Image & Segmentation \\
DOES \cite{schafer2023does} & Industrial & Unalloyed scrap & Shredded & Image/Video & Random Crops \\
SteelDS (Ours) & Industrial & E40 Steel \& Copper & Shredded & Image/Video & Segmentation \\
\bottomrule
\end{tabular}%
}
\end{table*}


\begin{figure}[t]
  \centering
  \begin{subfigure}[b]{0.24\linewidth}
    \centering
    \includegraphics[width=\linewidth]{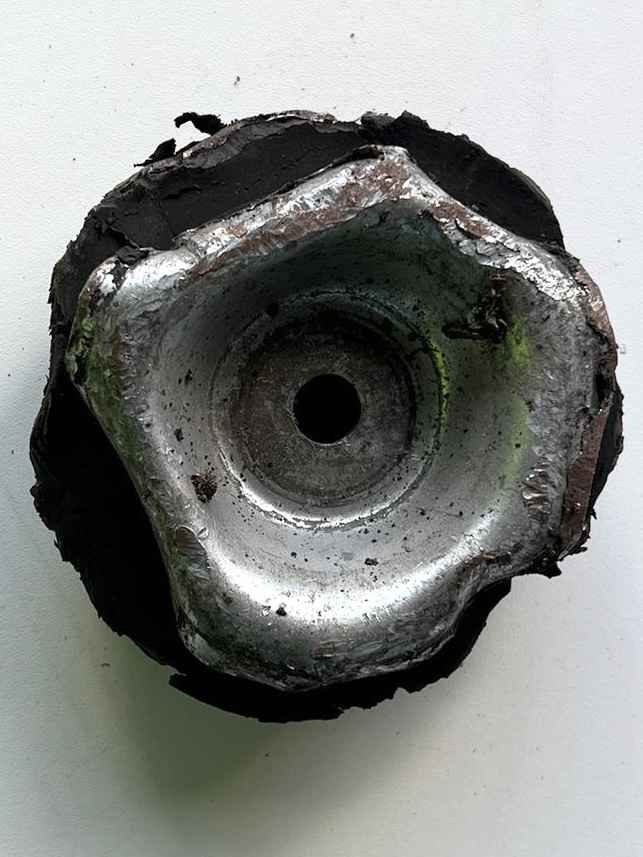}
  \end{subfigure}\hfill
  \begin{subfigure}[b]{0.24\linewidth}
    \centering
    \includegraphics[width=\linewidth]{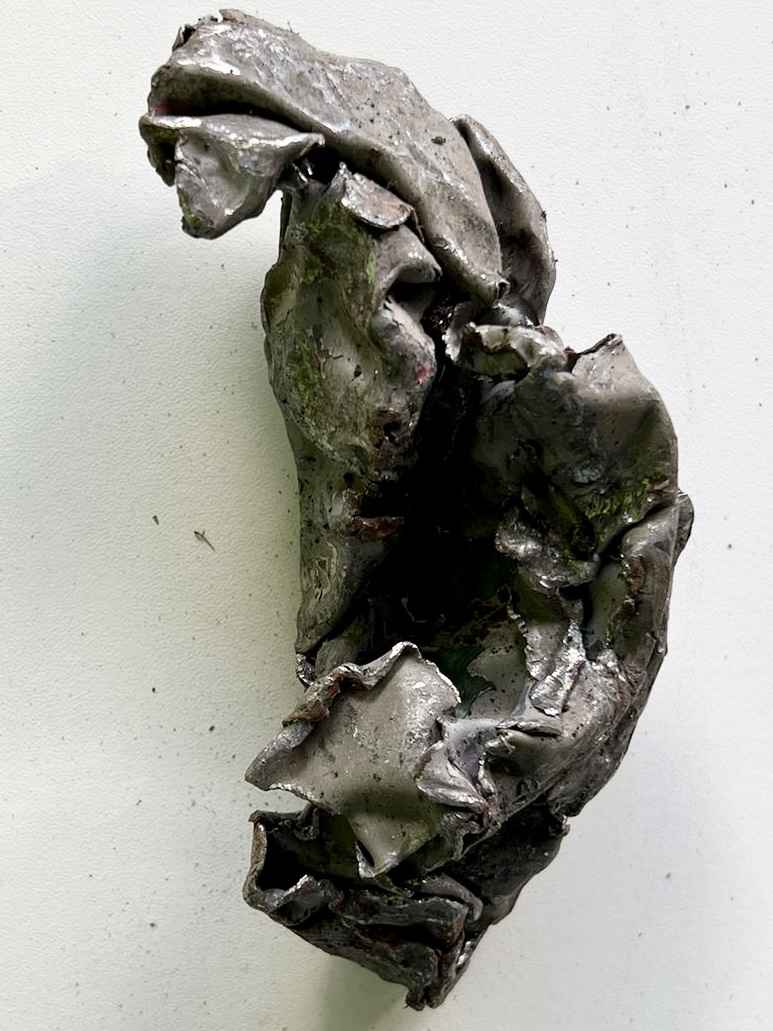}
  \end{subfigure}\hfill
  \begin{subfigure}[b]{0.24\linewidth}
    \centering
    \includegraphics[width=\linewidth]{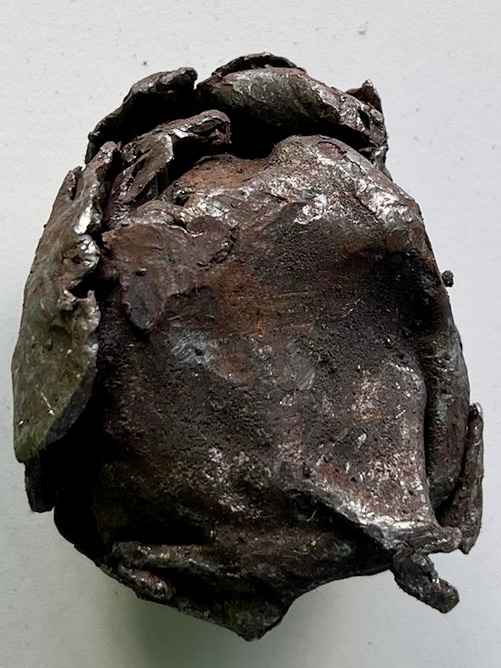}
  \end{subfigure}
  \begin{subfigure}[b]{0.24\linewidth}
    \centering
    \includegraphics[width=\linewidth]{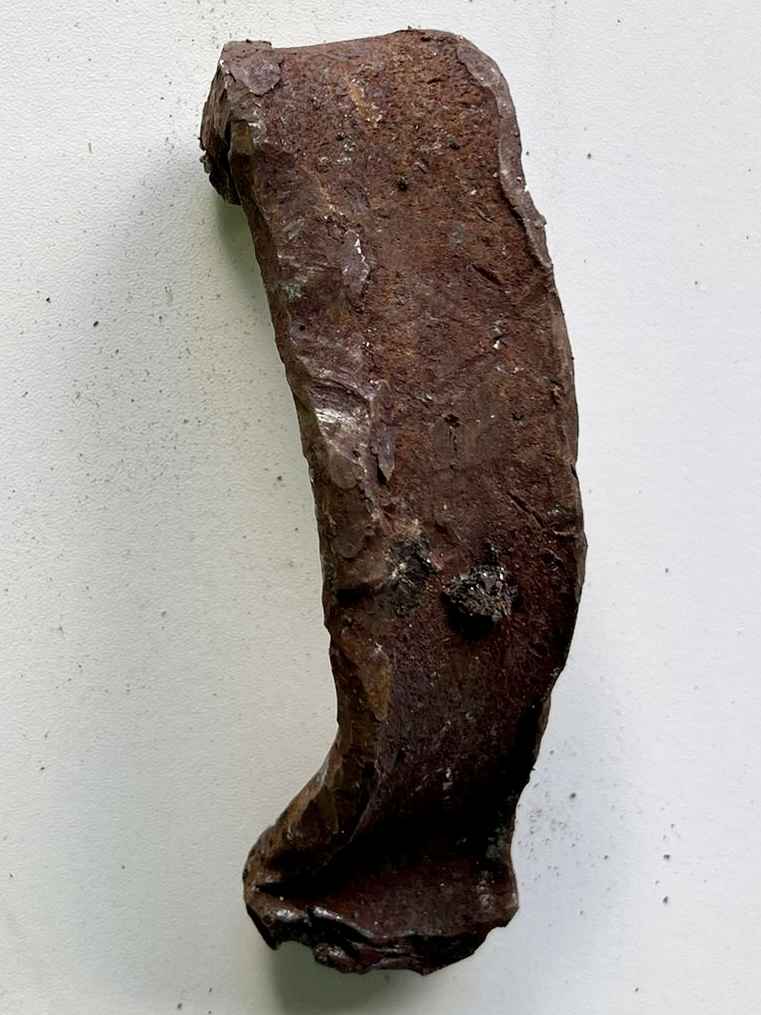}
  \end{subfigure}

  \vspace{0.2cm} 

  \begin{subfigure}[b]{0.24\linewidth}
    \centering
    \includegraphics[width=\linewidth]{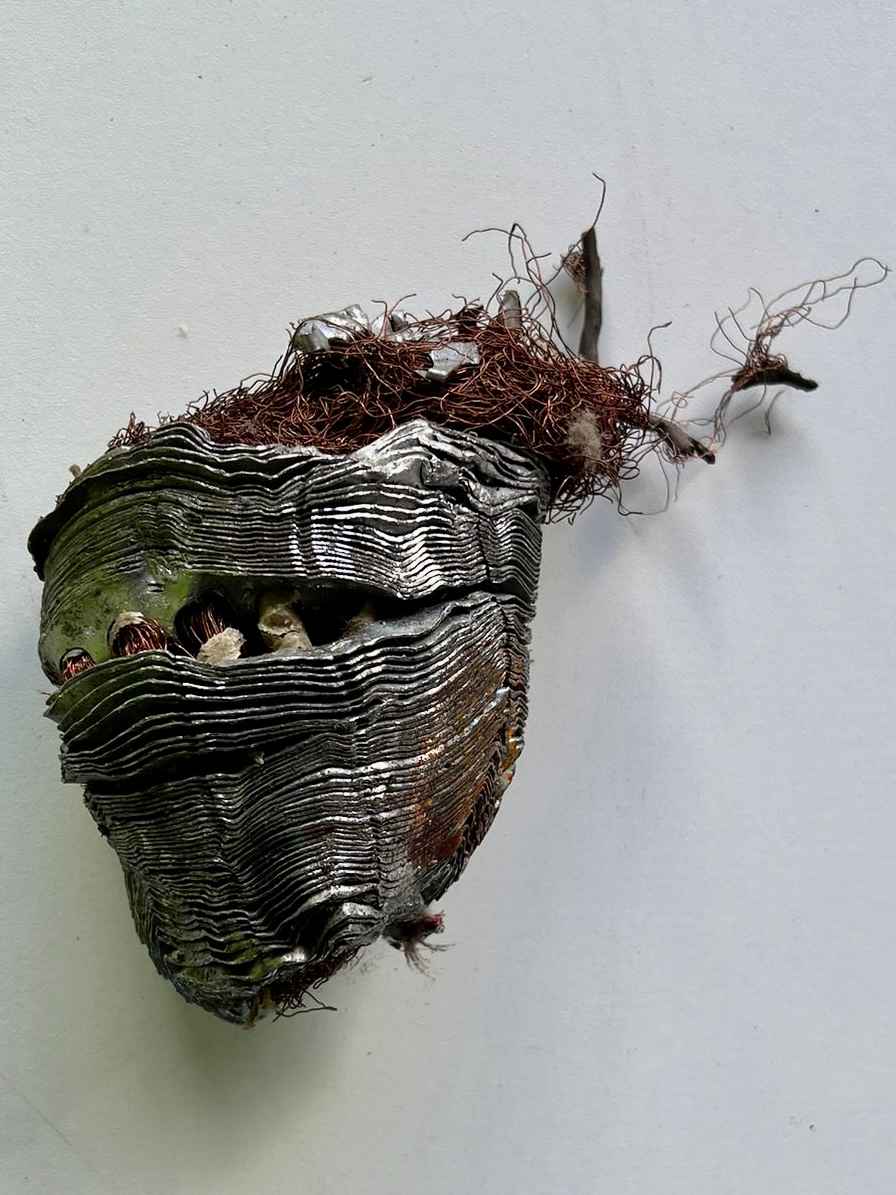}
  \end{subfigure}\hfill
  \begin{subfigure}[b]{0.24\linewidth}
    \centering
    \includegraphics[width=\linewidth]{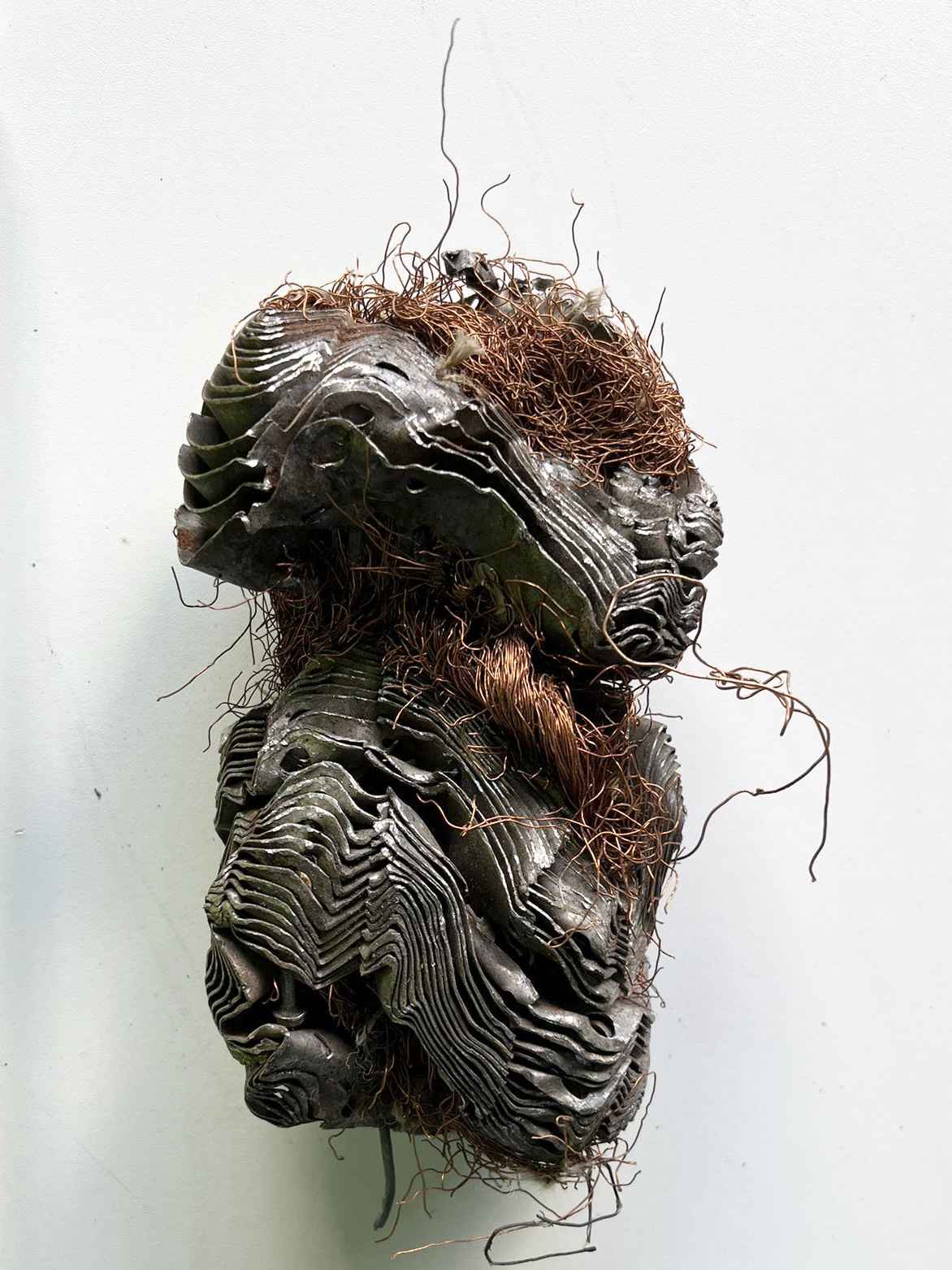}
  \end{subfigure}\hfill
  \begin{subfigure}[b]{0.24\linewidth}
    \centering
    \includegraphics[width=\linewidth]{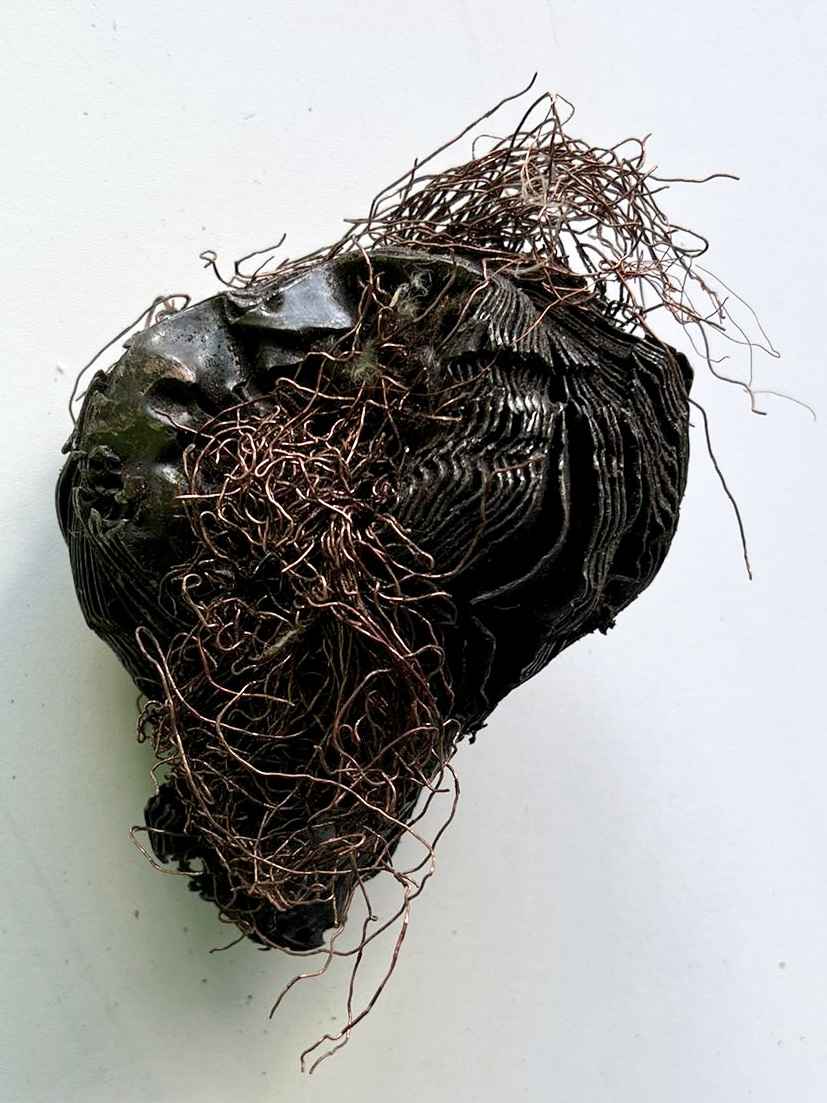}
  \end{subfigure}
  \begin{subfigure}[b]{0.24\linewidth}
    \centering
    \includegraphics[width=\linewidth]{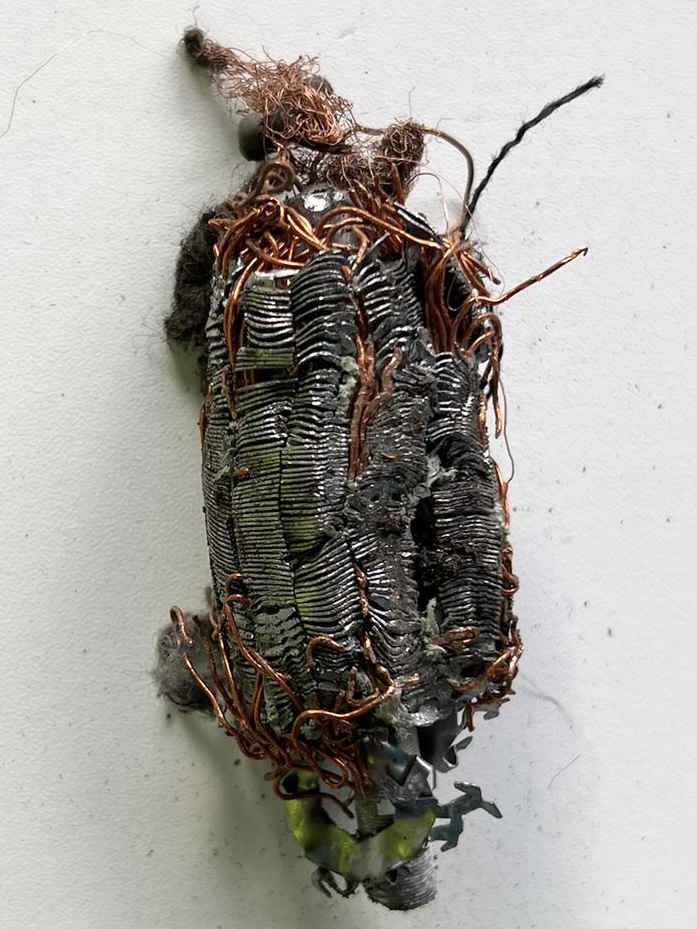}
  \end{subfigure}

  \caption{Randomly sampled steel objects (top row) and copper objects (bottom row) from the complete dataset.}
  \label{fig:sample_objects}
\end{figure}

\section*{Materials and Methods}
This section outlines the procedural workflow for material sampling, optical data acquisition, and computational pre-processing.

\subsection*{Material Specification and Sampling}

The dataset consists of 497 shredded metal objects sampled from an industrial E40-grade steel scrap fraction, provided from Scholz Austria GmbH in Laxenburg.
The material underwent standard industrial preprocessing, including mechanical shredding and magnetic separation. 
The sampled objects specifically represent the post-magnetic fraction that typically requires manual sorting to remove residual impurities to meet secondary steelmaking quality standards. 
To preserve the authentic industrial properties, no object cleaning, surface preparation, or pre-labeling was performed prior to data acquisition.
Objects were manually assigned to two primary material classes based on visual inspection: 
\begin{itemize}
    \item \textbf{Steel:} Purely ferrous components showing no visual trace of other metals (Fig.~\ref{fig:sample_objects} top row). 
    Coloration ranges from reflective or matte silver to dark gray, frequently exhibiting reddish hues due to surface oxidation. 
    Morphologies are highly heterogeneous, encompassing intact structural elements (e.g., screws), heavily deformed or compacted scrap (e.g., crushed metal agglomerates), and fragmented sheet metal.
    \item \textbf{Copper:} Objects containing any visible copper components, either isolated or embedded within a ferrous matrix (Fig.~\ref{fig:sample_objects} bottom row). 
    Surfaces predominantly exhibit characteristic copper coloration, though frequently obscured by industrial residue. 
    Typical topologies include exposed copper wires, isolated coils, and compound objects where copper windings are partially or fully enclosed within steel cavities.
\end{itemize}

\begin{table*}[t]
\caption{Material composition of the dataset, showing the categorization of steel (S1–S5) and copper (C1–C3) scrap based on size, weight, and quantity.}

\label{tab:comparison}
\begin{center}

\begin{minipage}{0.51\textwidth}
\centering
\subcaption{Steel-material-breakdown into groups from S1 to S5.} 
\resizebox{\textwidth}{!}{%
\begin{tabular}{cccc}
\toprule
\textbf{Group} & \textbf{Size} & \textbf{Weight [kg]} & \textbf{Quantity} \\
\midrule
\textbf{S1} & Large & 9.0 & 45\\
\textbf{S2} & Medium-Large & 20.0 & 95\\ 
\textbf{S3} & Medium & 9.0 & 72\\
\textbf{S4} & Medium-Small & 6.0 & 80\\
\textbf{S5} & Small& 2.5 & 104\\
\midrule
\textbf{Total} & & 46.5 & 396\\
\bottomrule
\end{tabular}%
}
\end{minipage}
\hfill
\begin{minipage}{0.45\textwidth}
\centering
\subcaption{Copper-material-breakdown into groups from C1 to C3} 
\resizebox{\textwidth}{!}{%
\begin{tabular}{cccc}
\toprule
\textbf{Group} & \textbf{Size} & \textbf{Weight [kg]} & \textbf{Quantity} \\
\midrule
\textbf{C1} & Large & 15.0 & 17 \\
 &  & & \\ 
\textbf{C2} & Medium & 8.5 & 47\\
 &  & & \\
\textbf{C3} & Small& 1.0 & 37 \\
\midrule
\textbf{Total} & & 24.5 & 101\\
\bottomrule
\end{tabular}%
}
\end{minipage}
\label{tab:material}
\end{center}
\end{table*}

Beyond material classification, the objects were subdivided into qualitative size categories (small, medium, large), resulting in groups S1–S5 for steel and C1–C3 for copper (see Table~\ref{tab:material}). 
While the captured image data yields a 2D object area in pixels corresponding to a physical range of $3.9\,\text{cm}^2$ to $160\,\text{cm}^2$ (spatial resolution $\approx 0.026\,\text{cm/pixel}$), exact 3D volumetric measurements of individual particles were not conducted. 
The highly irregular, deformed, and partially hollow nature of post-shredder scrap makes precise volumetric quantification operationally prohibitive. 
Consequently, the physical size categorization relies exclusively on visual inspection and practical handling impressions (e.g., perceived bulk and weight) during the sorting process. 
This heuristic approach introduces subjective variance but accurately simulates the cognitive evaluation performed by human workers during manual sorting, where strict dimensional boundaries are impractical.

During the sampling and handling process, minor fragmentation of highly brittle objects occurred. 
These fragments were retained and subsequently treated as independent instances within the dataset. 
Because persistent global object identifiers were not assigned across different recording sessions, this natural fragmentation does not affect the structural integrity of the frame-level instance segmentation.

\subsection*{Experimental Setup and Data Acquisition}

\begin{figure}[t]
  \centering
  \begin{subfigure}[b]{0.55\linewidth}
    \centering
    \includegraphics[height=5cm, keepaspectratio]{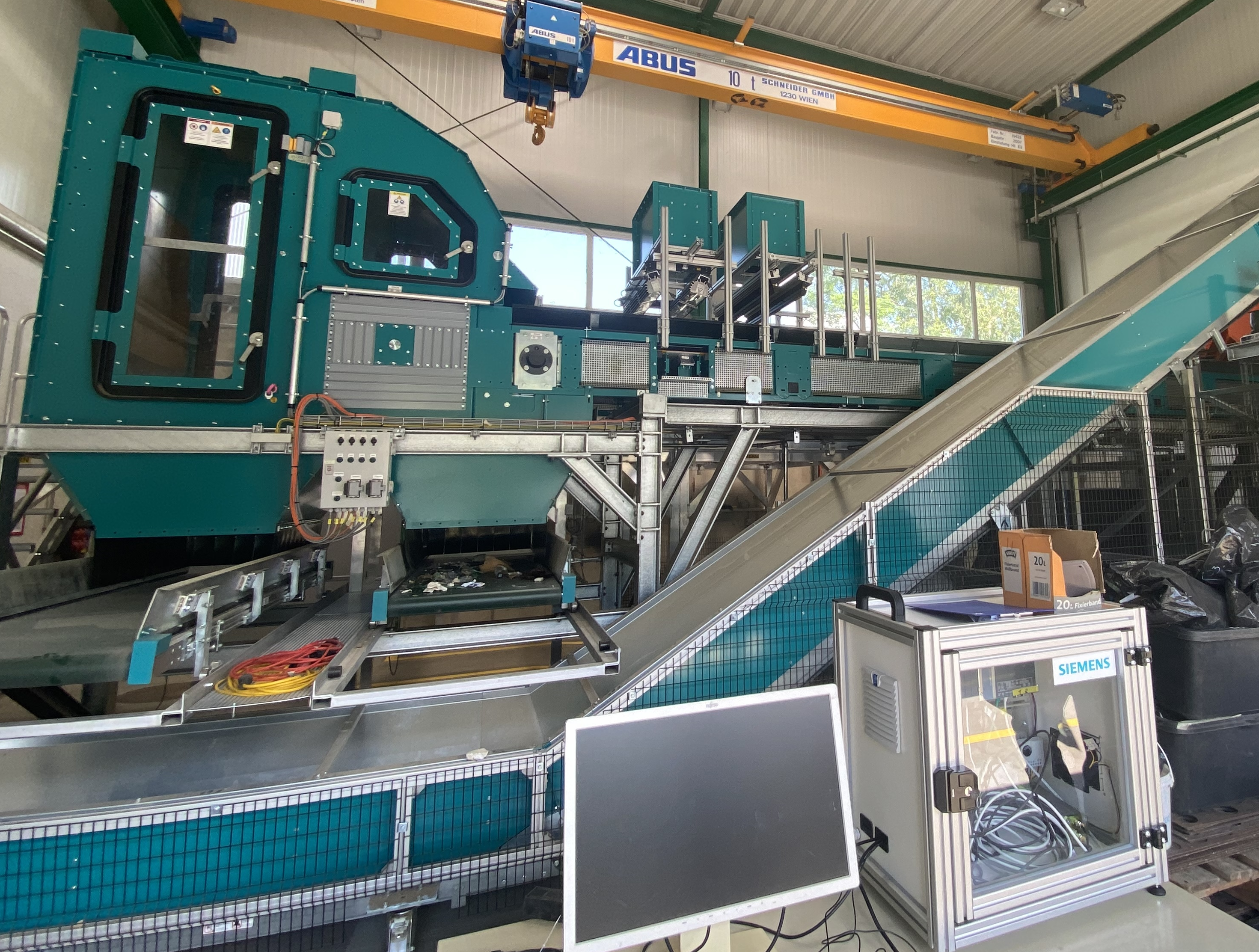}
    \caption{Digital Waste Research Lab at the Technical University of Leoben.}
    \label{fig:dwrl}
  \end{subfigure}
  \hspace{0.01\linewidth} 
  \begin{subfigure}[b]{0.35\linewidth}
    \centering
    \includegraphics[height=5cm, keepaspectratio]{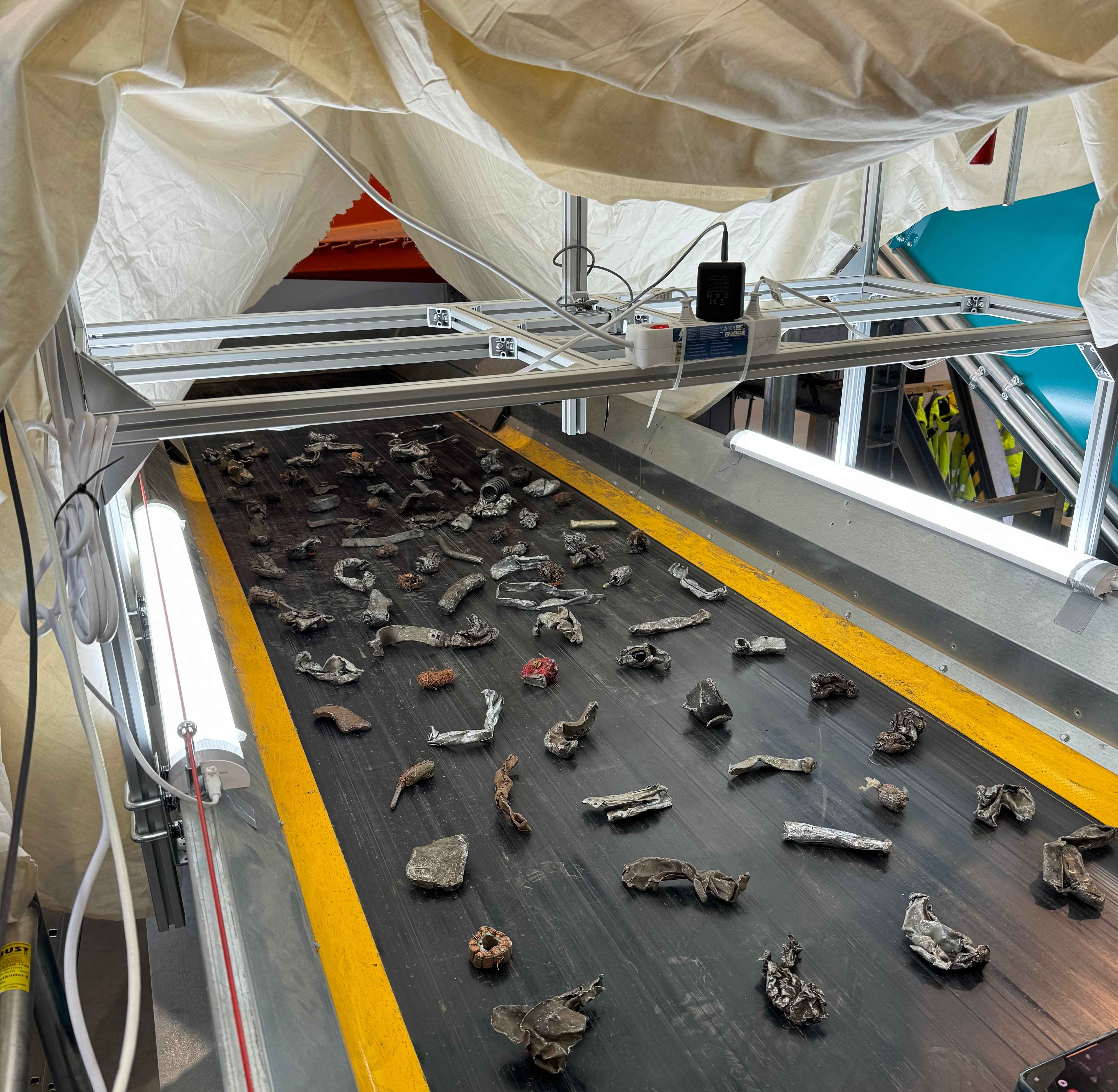}
    \caption{Experimental camera and LED setup.}
    \label{fig:dwrl_2}
  \end{subfigure}
  \caption{Recordings at the Digital Waste Research Lab.}
  \label{fig:dwrl_overall}
\end{figure}

\begin{figure}[t]
    \centering
    \includegraphics[width=0.8\linewidth]{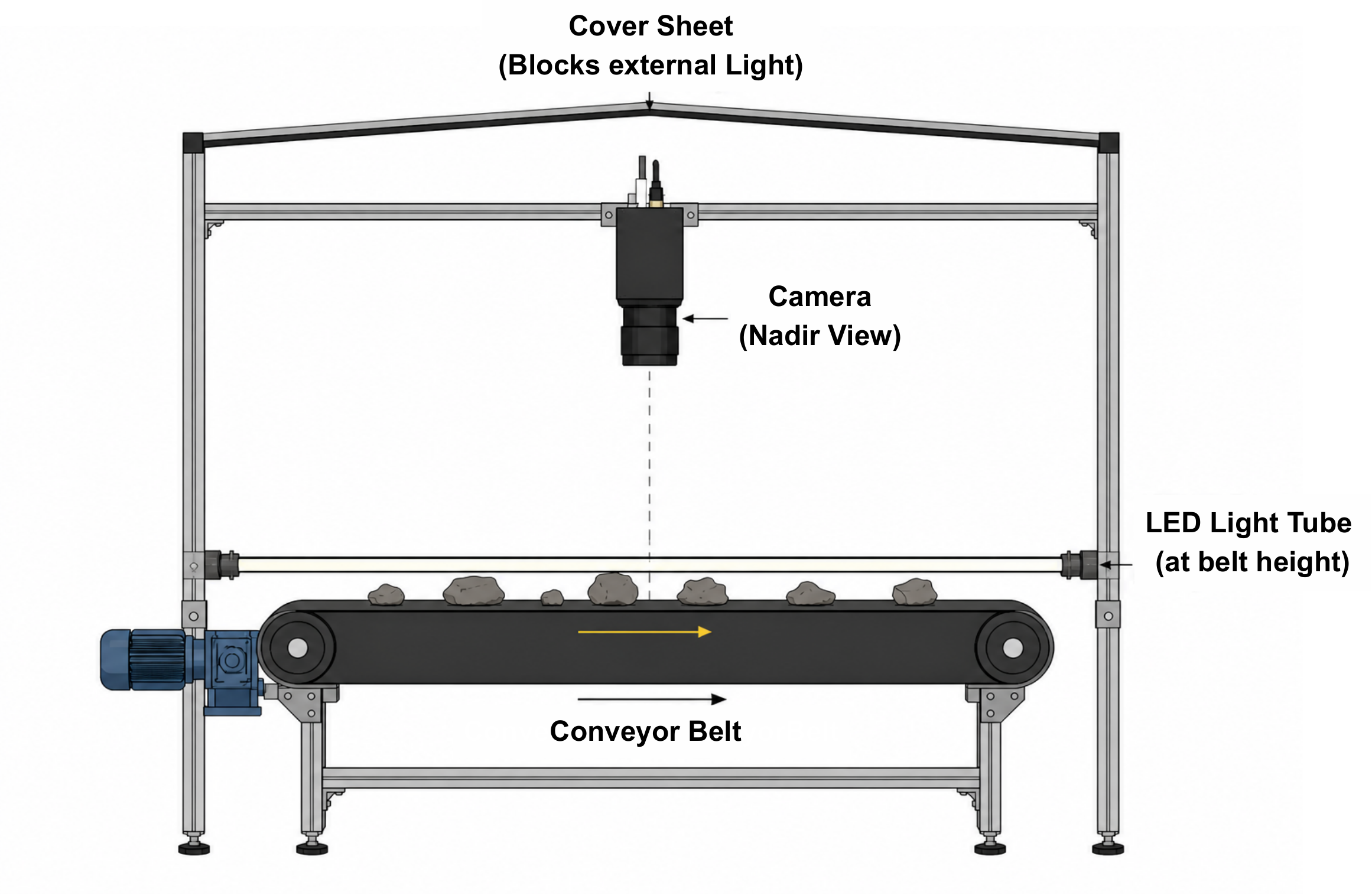}
    \caption{Schematic representation of the experimental data acquisition setup. The configuration features a top-mounted camera for nadir video capture, bilateral diffuse LED panels for uniform material illumination, and an opaque enclosure to prevent ambient light interference and surface reflections.}
    \label{fig:setup}
\end{figure}

Data collection was conducted at the Digital Waste Research Lab of the Technical University of Leoben (Fig.~\ref{fig:dwrl_overall}). 
The physical configuration was designed to replicate industrial conveyor belt inspection systems while maintaining controlled laboratory conditions to ensure data consistency. 
The setup utilized a 1.0~m wide conveyor belt operating continuously at speeds between 0.4~m/s and 0.5~m/s. 
Scrap objects were manually arranged on the moving belt in two spatial configurations to simulate varying industrial material densities: isolated particles with a minimum inter-particle spacing of 1~cm, and dense clusters with no forced spacing, which allowed for natural collisions and partial occlusions.

Optical data was acquired using a GoPro Hero 11 camera mounted 0.5~m directly above the belt center in a strict nadir perspective. Video streams were captured at 4K Ultra HD resolution (3680 $\times$ 2160 pixels) at 100 frames per second. A fixed shutter speed of 1/100~s and in-camera image sharpening algorithms were applied to minimize motion blur induced by the conveyor speed. To manage highly reflective metallic surfaces, diffuse LED panel lights were arranged symmetrically around the capture zone to provide uniform illumination and reduce harsh shadows (see Fig.~\ref{fig:setup}). Furthermore, the immediate recording area was physically enclosed to block ambient light and overhead window glare, ensuring temporal lighting consistency across all recording sessions.

A fundamental limitation of this single-view 2D optical setup is the restriction to top-facing surfaces. Consequently, copper components located exclusively on the underside of a scrap particle are systematically occluded and cannot be recorded or annotated. This remains a standard constraint in industrial 2D optical sorting systems.

\subsection*{Computational Processing and Annotation Protocol}
This section outlines the comprehensive pipeline used to transform raw video recordings into a high-fidelity, machine-learning-ready dataset. 
It details the technical procedures for frame extraction, the semi-automated instance segmentation workflow, and the quality assurance protocols during annotation.

\paragraph*{Video Pre-Processing:} 
Raw video files were recorded at 100~fps. 
To reduce temporal redundancy and optimize the dataset for machine learning applications, the frame rate was downsampled to 20~fps. 
Frame extraction was executed on an Ubuntu 24.04 environment using \texttt{ffmpeg}. 
To ensure lossless conversion and prevent secondary compression artifacts, frames were extracted directly to the PNG format without rescaling.

\paragraph*{Annotation Tool:} 
Instance-level segmentations were generated utilizing the custom-built \textit{Semi-Autonomous Fast Object Segmentation and Tracking Tool for Industrial Applications} (FOST)~\cite{neubauer2024semi}. 
The software generates semi-automated mask proposals, which were subsequently manually corrected and verified by human annotators.

\paragraph*{Annotation Rules:}
To maintain high data quality and consistency across all subsets, strict pixel-level annotation guidelines were enforced. 
Specifically, polygon vertices were required to strictly trace the physical boundaries of the metal particles, explicitly excluding any cast shadows on the conveyor belt. 
In instances where minor motion blur occurred on object edges despite the fast shutter speed, masks were drawn along the estimated solid boundary to omit diffuse motion trails. 
Furthermore, any physical fragments resulting from material breakage during transport were systematically labeled as entirely independent object instances.

\paragraph*{Quality Assurance:} 
A two-stage verification pipeline was implemented. 
Following the primary annotation phase, a secondary review was conducted on all frames across all subsets. 
This review specifically targeted and corrected systemic anomalies, such as the accidental inclusion of background textures or inconsistent handling of highly reflective copper surfaces.

\paragraph*{Data Export Format:} 
All segmentation masks and bounding box coordinates were normalized and exported into standard YOLO segmentation format to ensure immediate compatibility with contemporary computer vision frameworks.

\subsection*{Dataset Structuring and Split Methodology}

The final dataset is structured into five distinct subsets (a1–a5), encompassing approximately 30 video sequences and 24,297 labeled frames. 
The structuring follows a cumulative complexity approach:
\begin{itemize}
    \item \textbf{Subsets a1 to a3:} Cumulative in object inclusion (e.g., a2 contains all objects from a1 plus additional objects; a3 contains a2 plus new ones). These subsets contain objects with a minimum of 1~cm inter-particle spacing.
    \item \textbf{Subset a4:} Introduces high complexity through dense clusters (no forced spacing, resulting in occlusions), utilizing the complete object pool from a3.
    \item \textbf{Subset a5:} Serves as a holistic evaluation set, combining all materials (S1–S5, C1–C3) and both spacing configurations.
\end{itemize}

\begin{figure}[t]
  \centering
  \begin{subfigure}[b]{0.32\linewidth}
    \centering
    \includegraphics[width=\linewidth]{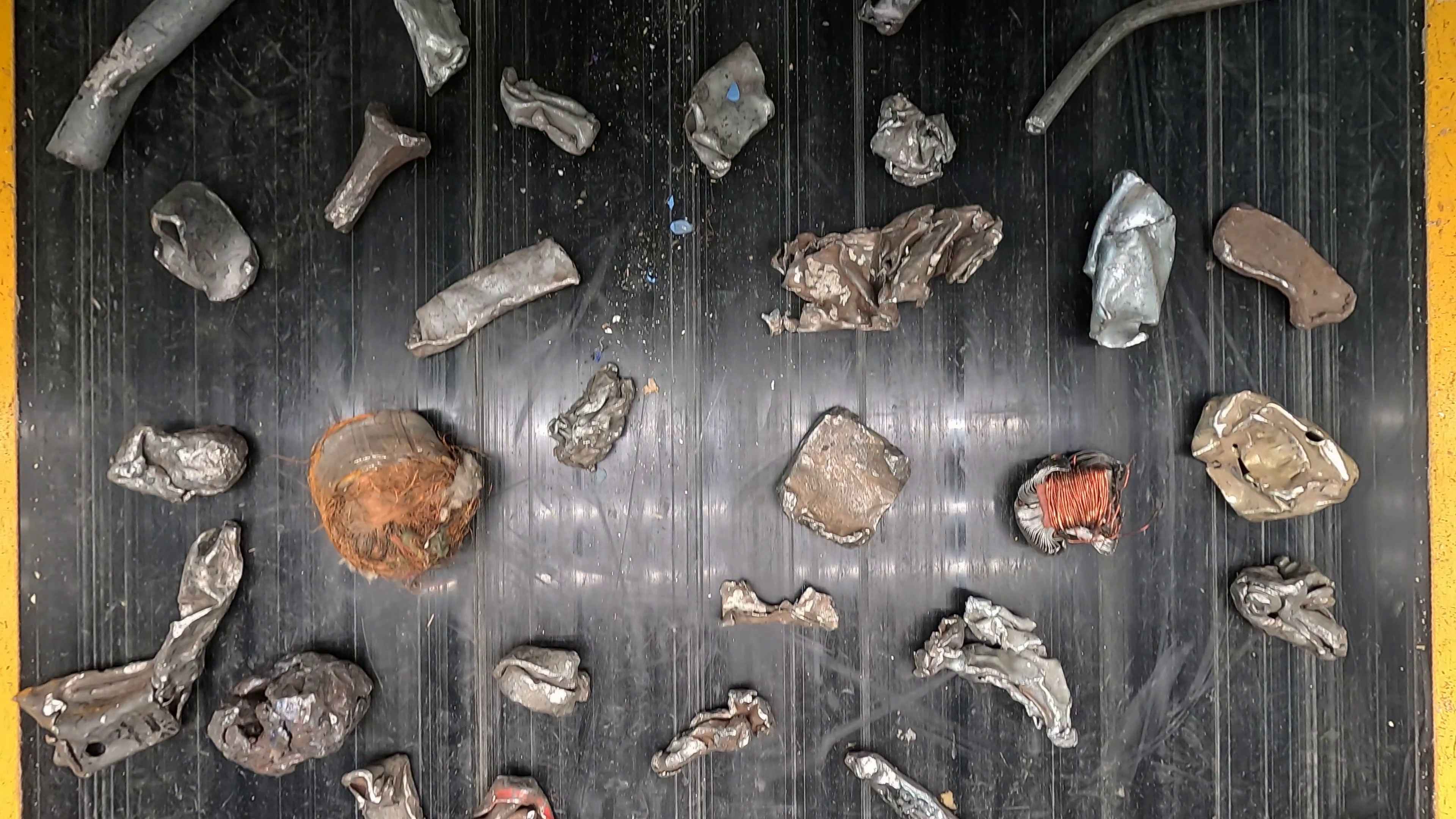}
    \caption{Subset a1}
  \end{subfigure}\hfill
  \begin{subfigure}[b]{0.32\linewidth}
    \centering
    \includegraphics[width=\linewidth]{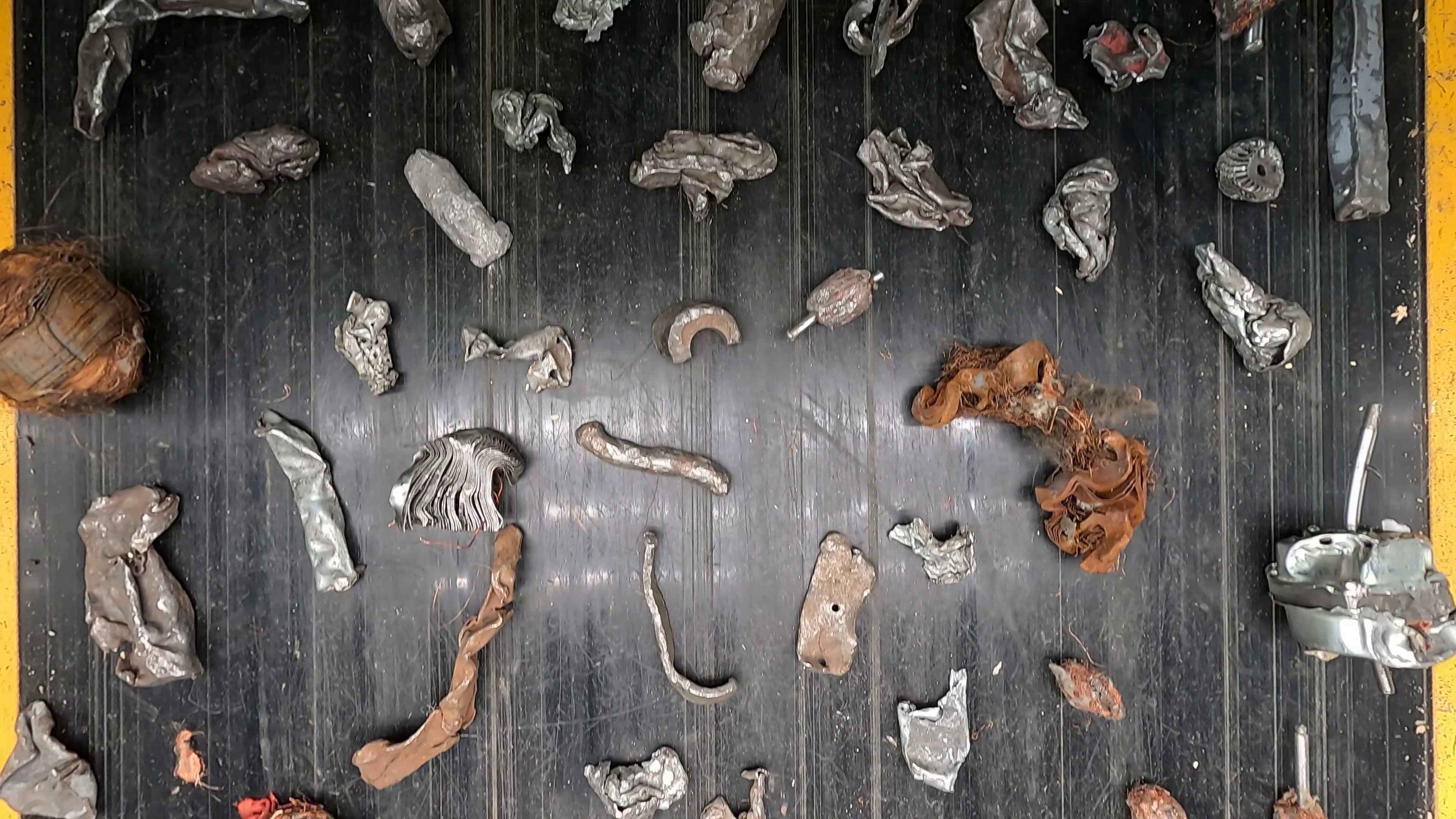}
    \caption{Subset a2}
    \label{fig:ds2_orig}
  \end{subfigure}\hfill
  \begin{subfigure}[b]{0.32\linewidth}
    \centering
    \includegraphics[width=\linewidth]{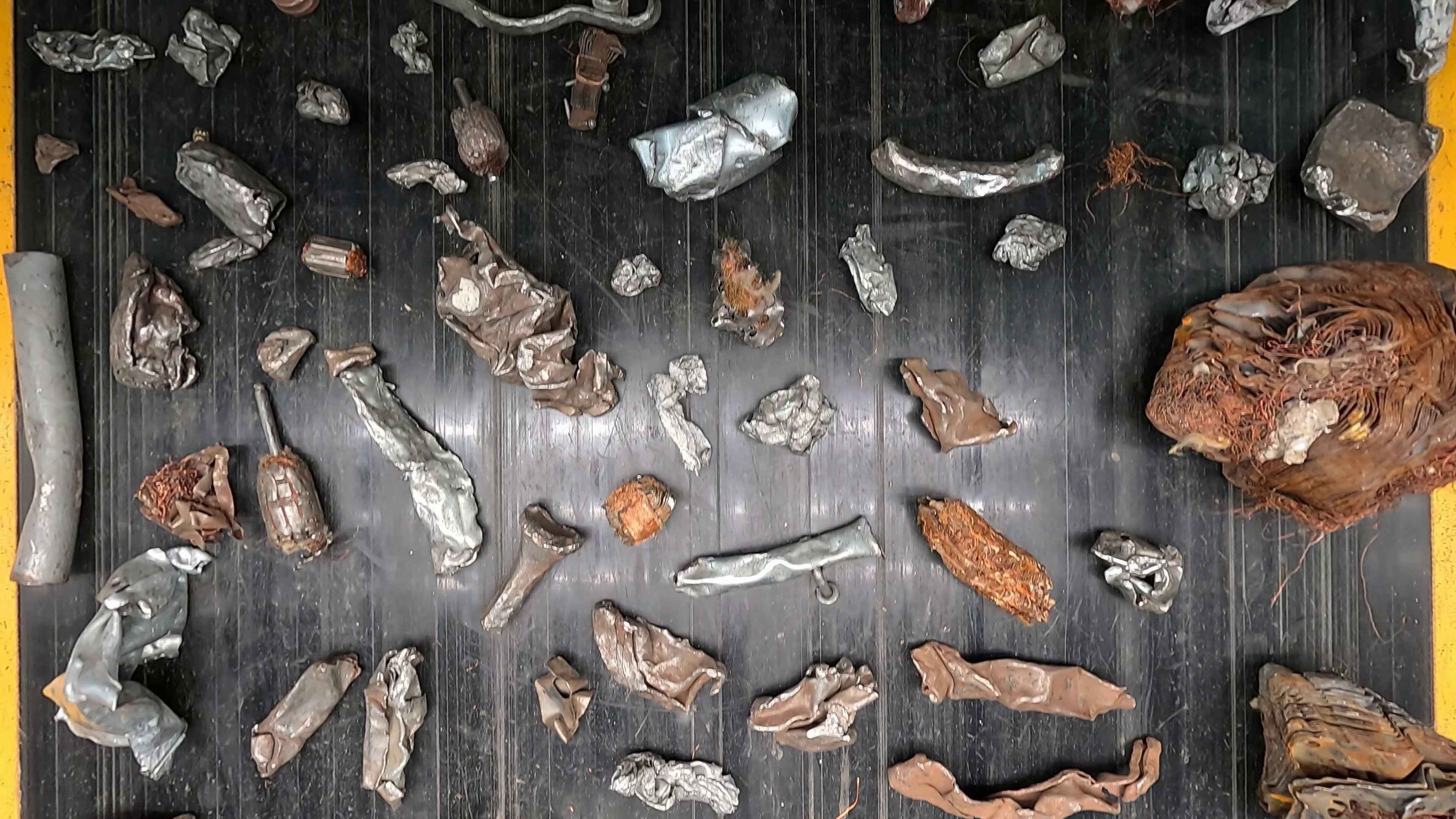}
    \caption{Subset a3}
    \label{fig:ds3_orig}
  \end{subfigure}

  \begin{subfigure}[b]{0.32\linewidth}
    \centering
    \includegraphics[width=\linewidth]{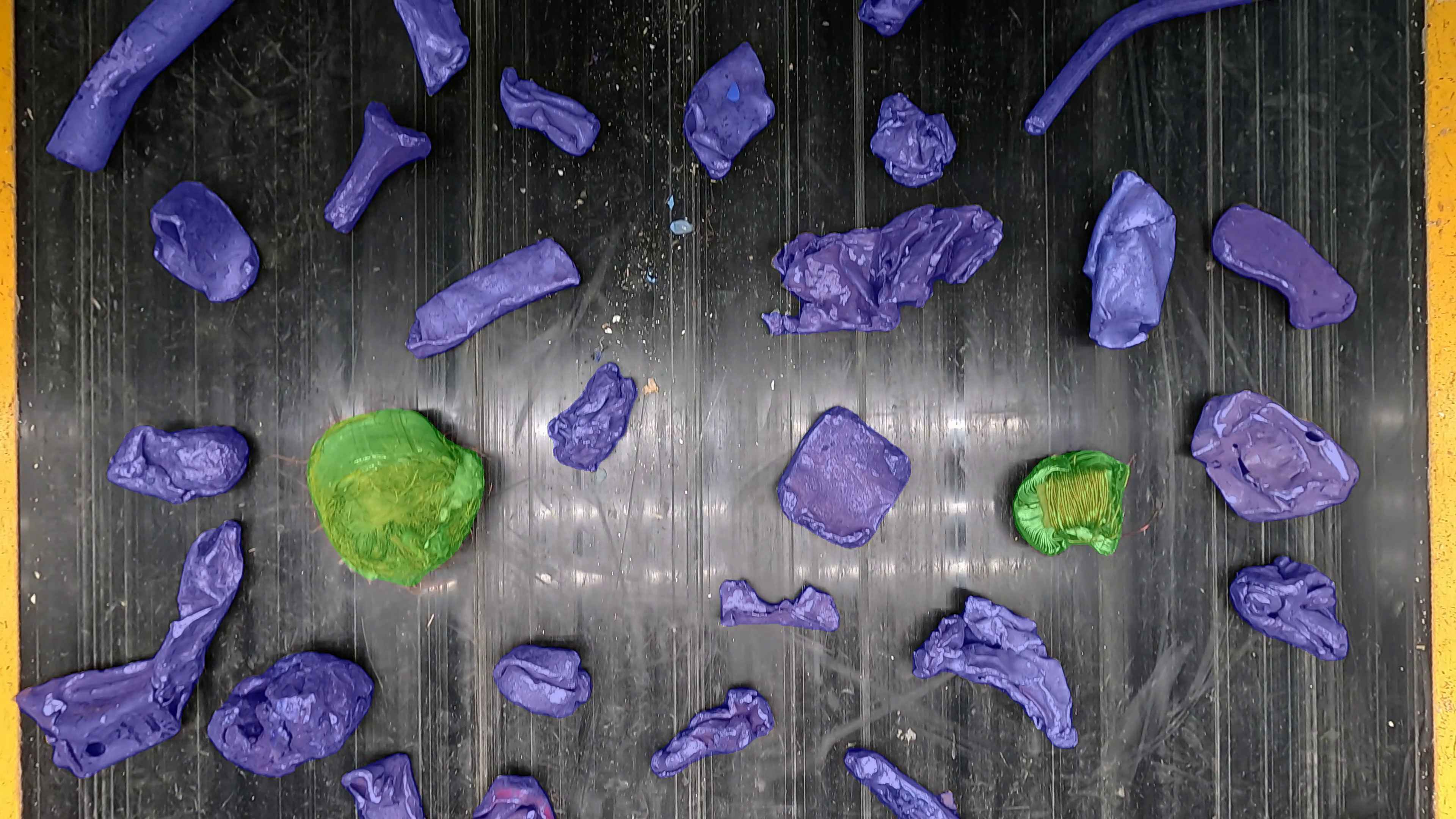}
    \caption{Ground Truth a1}
    \label{fig:ds1_gt}
  \end{subfigure}\hfill
  \begin{subfigure}[b]{0.32\linewidth}
    \centering
    \includegraphics[width=\linewidth]{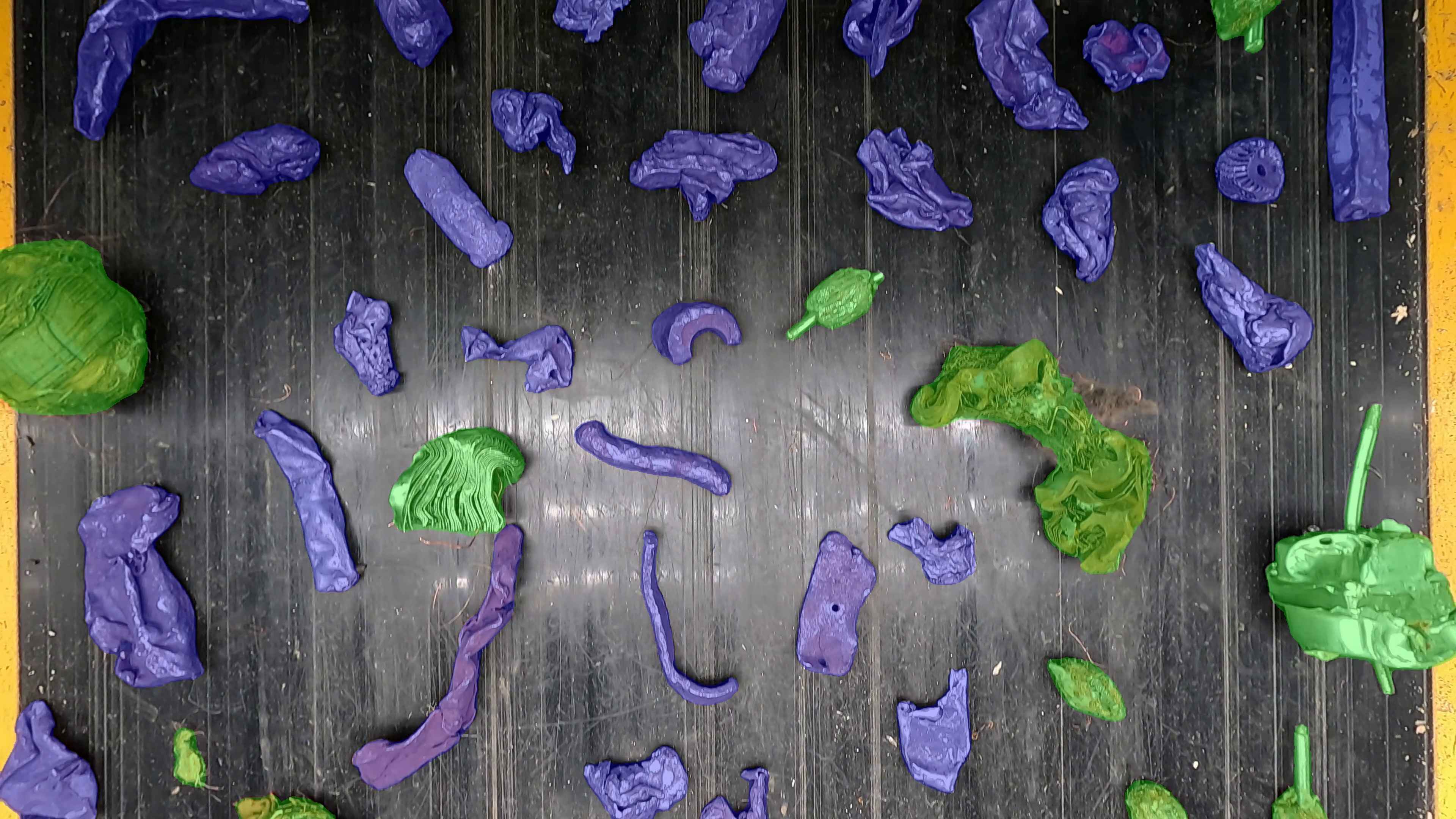}
    \caption{Ground Truth a2}
    \label{fig:ds2_gt}
  \end{subfigure}\hfill
  \begin{subfigure}[b]{0.32\linewidth}
    \centering
    \includegraphics[width=\linewidth]{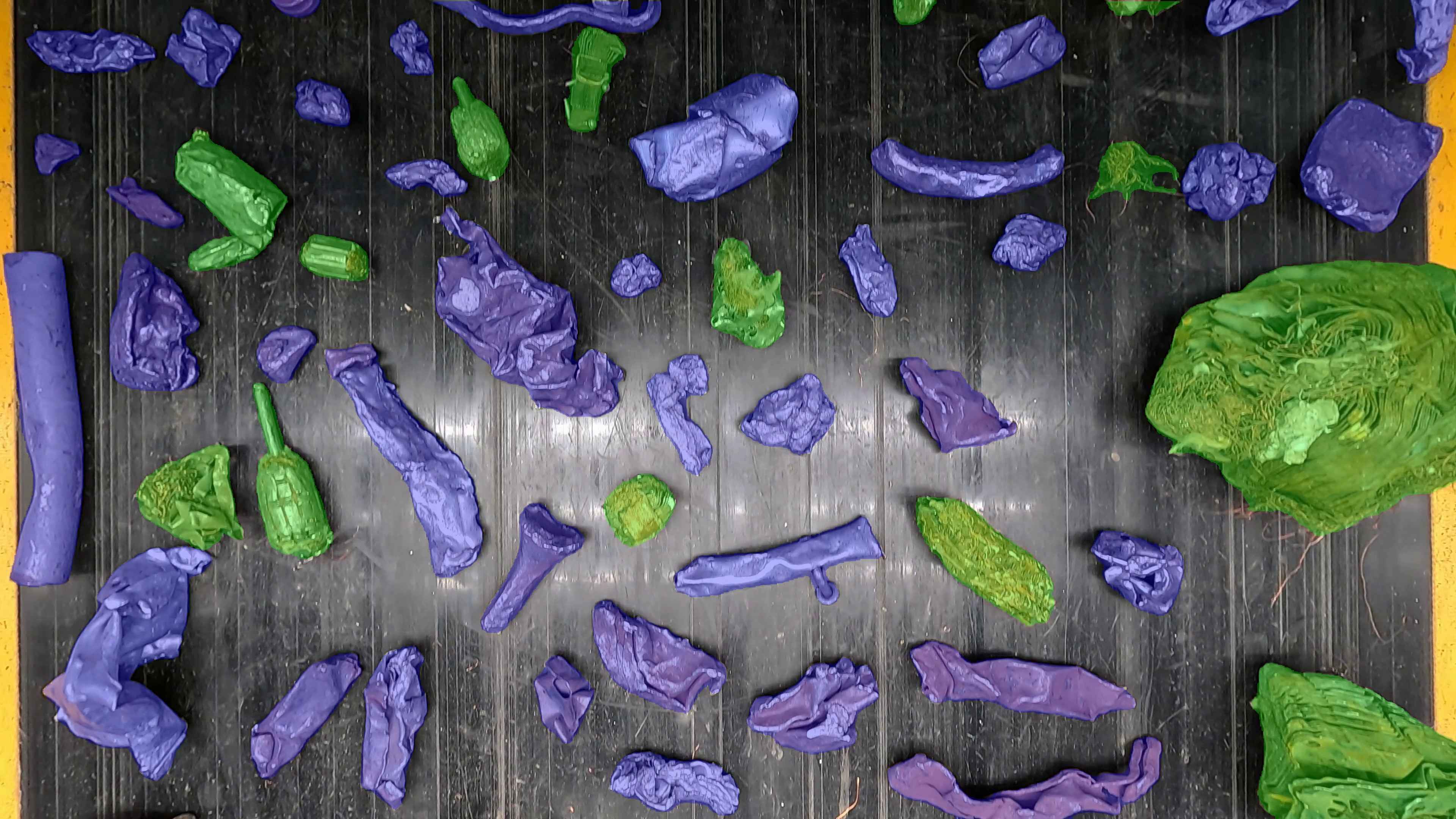}
    \caption{Ground Truth a3}
    \label{fig:ds3_gt}
  \end{subfigure}

  \caption{Comparison of the original images (top row) with the corresponding ground-truth segmentations (bottom row) from three different datasets. In the segmentation masks, green indicates copper objects and blue indicates steel objects.}
  \label{fig:dataset_examples1}
\end{figure}

\begin{figure}[t]
  \centering
  \begin{subfigure}[b]{0.32\linewidth}
    \centering
    \includegraphics[width=\linewidth]{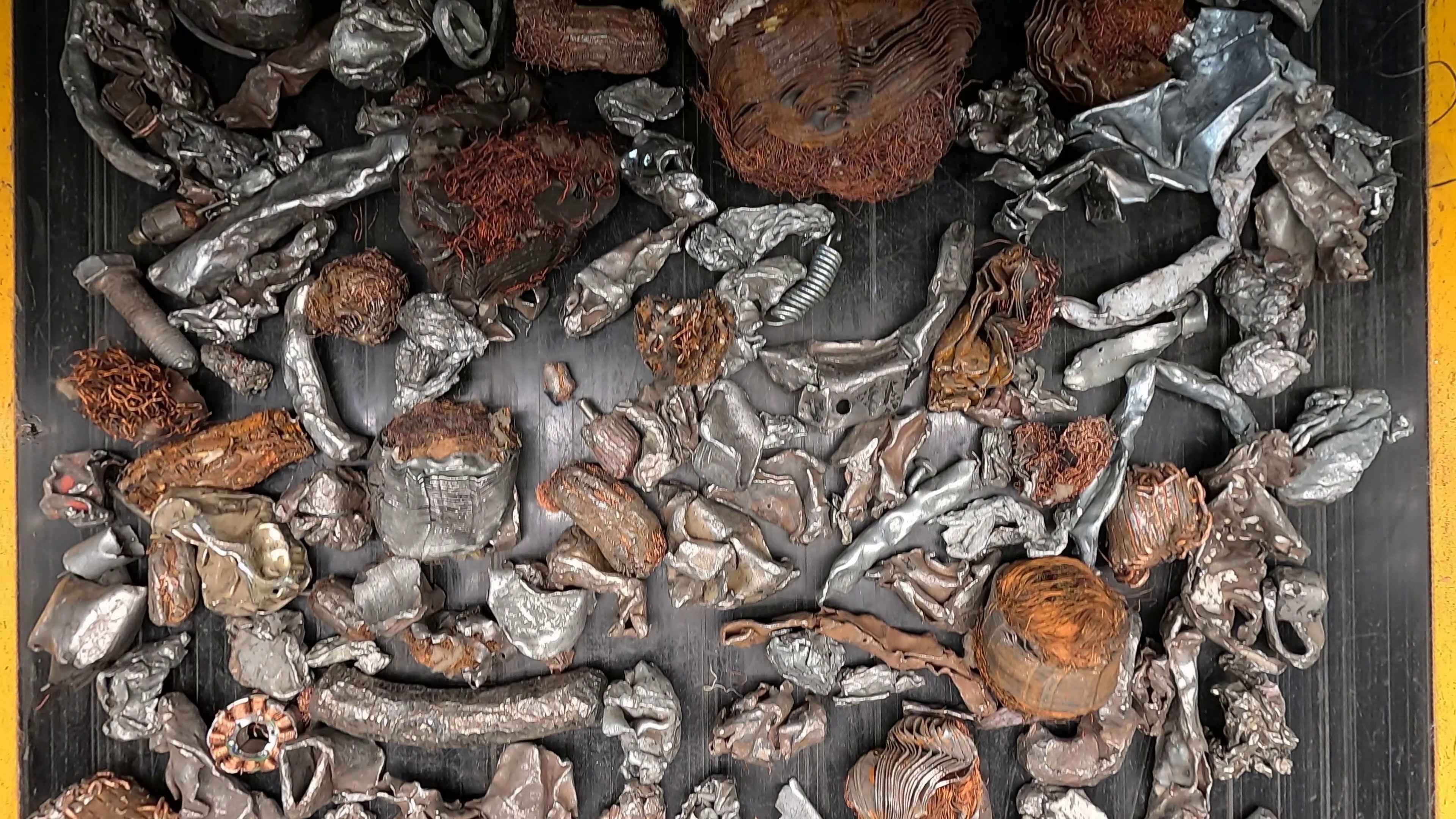}
    \caption{Subset a4}
  \end{subfigure}\hfill
  \begin{subfigure}[b]{0.32\linewidth}
    \centering
    \includegraphics[width=\linewidth]{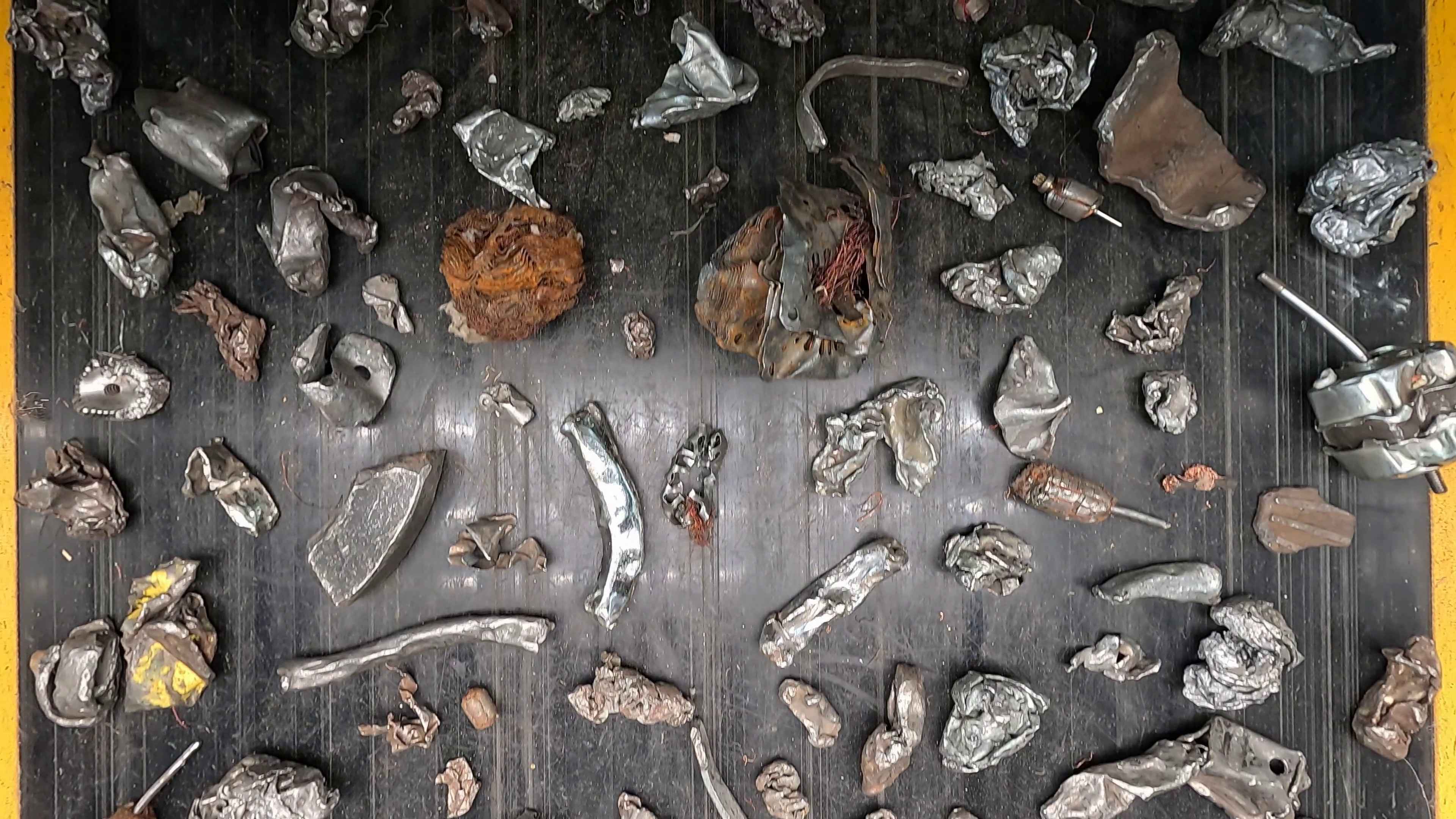}
    \caption{Subset a5 - Test 1}
  \end{subfigure}\hfill
  \begin{subfigure}[b]{0.32\linewidth}
    \centering
    \includegraphics[width=\linewidth]{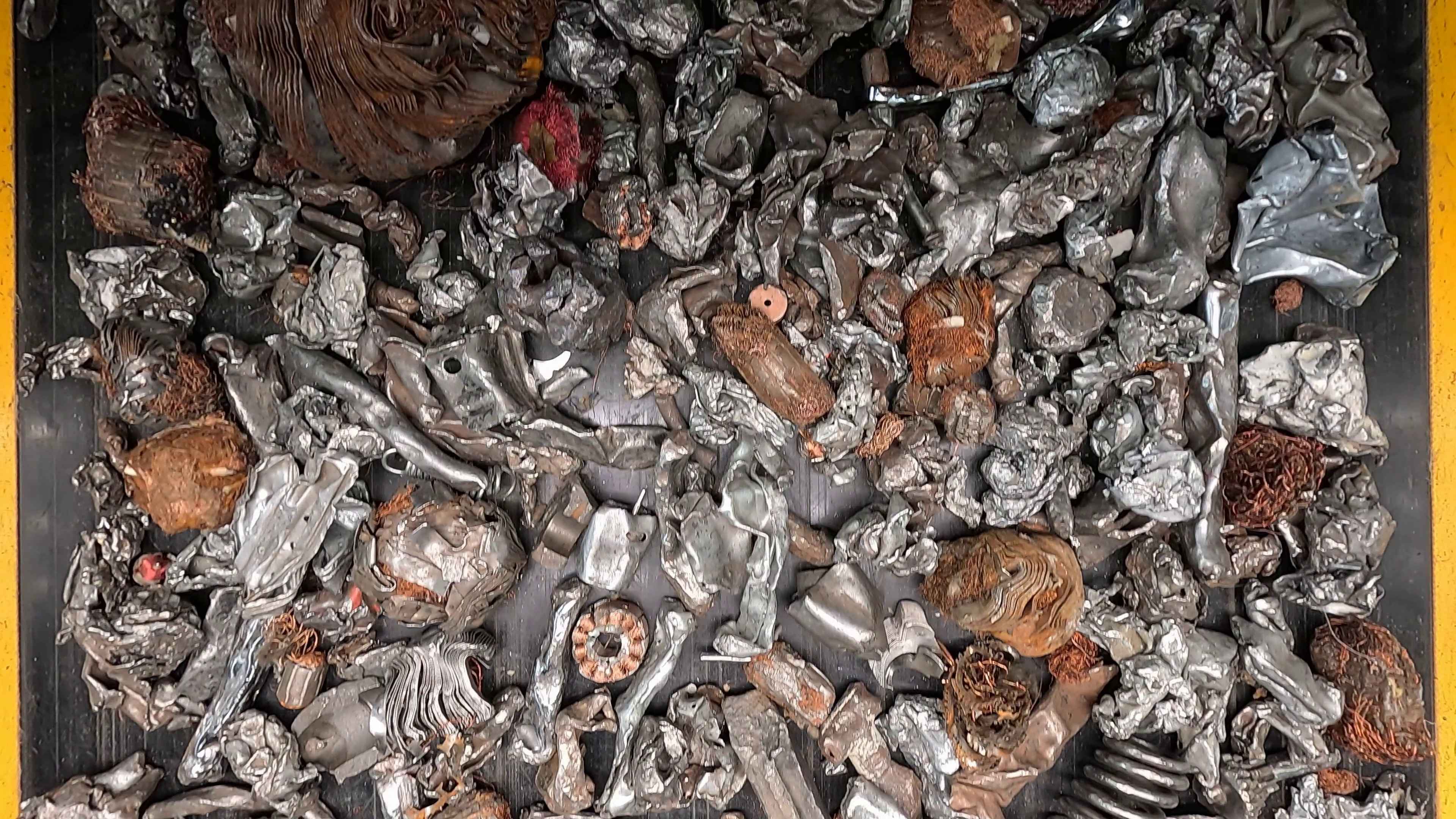}
    \caption{Subset a5 - Test 2}
  \end{subfigure}

  \caption{Unlabeled subsets to test or to use for unsupervised training.}
  \label{fig:dataset_examples2}
\end{figure}

Figure~\ref{fig:dataset_examples1} and~\ref{fig:dataset_examples2} show representative frames with annotated steel and copper fragments across all subsets.

To ensure robust machine learning evaluation and prevent data leakage, strict split isolation was enforced. 
Objects are physically distinct and never shared across training, validation, or test splits. 

Because each object appears in multiple consecutive frames, the dataset natively supports object-level tracking and temporal learning. 
Detailed statistical breakdowns of these splits are provided in Section~\ref{data_record} (Tables~\ref{tab:runs_distances} and~\ref{tab:dataset_detail_information}).

\begin{table}[t]
\centering  
\caption{Dataset specifications including the number of recording runs, approximate inter-particle distance, and the total counts of steel and copper particles per split. The number of runs indicates how often the complete set of objects passed over the conveyor belt, with each run featuring a newly randomized spatial arrangement. 'Distances' defines the approximate spatial separation between objects on the conveyor.}
\label{tab:runs_distances}
\begin{tabular}{
  p{1.6cm}  
  >{\centering\arraybackslash}p{1cm}
  >{\centering\arraybackslash}p{1cm}
  >{\centering\arraybackslash}p{1.5cm}
  >{\centering\arraybackslash}p{1.5cm}
  >{\centering\arraybackslash}p{1.5cm}
  >{\centering\arraybackslash}p{1.5cm}
}
\hline
        & \textbf{Set}  & \textbf{Runs} & \textbf{Distances} & \textbf{\# Steel} & \textbf{\# Copper} & \textbf{\# Frames}\\ 
        \hline

\textbf{Dataset a1} & Train & 5 & \textgreater1cm & 96 & 11 & 3831\\
(annotated)         & Val.  & 3 & \textgreater1cm & 22 & 3 & 907\\
                   & Test  & 3 & \textgreater1cm & 22 & 3 & 956\\
\hline
\textbf{Dataset a2} & Train & 5 & \textgreater1cm & 182 & 44 & 6308\\
 (annotated)        & Val.  & 3 & \textgreater1cm & 55 & 10 & 1266\\
                   & Test  & 3 & \textgreater1cm & 55 & 10 & 1187\\
\hline
\textbf{Dataset a3} & Train & 5 & \textgreater1cm & 246 & 69 & 6715\\
(annotated)         & Val.  & 3 & \textgreater1cm & 75 & 16 & 1528\\
                   & Test  & 3 & \textgreater1cm & 75 & 16 & 1599\\
\hline
\textbf{Dataset a4} & Train & 2 & No & 246 & 69 & 1025\\
                   & Val.  & 2 & No & 75 & 16 & 549\\
                   & Test  & 2 & No & 75 & 16 & 555\\
\hline
\textbf{Dataset a5} & Test & 2 & \textgreater1cm & 396 & 101 & 3977\\
                   &  Test & 2 & No & 396 & 101 & 1233\\
\hline

\end{tabular}
\end{table}

\section*{Data Records}
\label{data_record}

The complete dataset is publicly hosted on Zenodo and can be accessed via \href{https://doi.org/10.5281/zenodo.20271102}{https://doi.org/10.5281/zenodo.20271102} \cite{neubauer2026steelds}.
The data is released under the Creative Commons Attribution 4.0 International (CC BY 4.0) license.

\subsection*{Data Formats}

The dataset comprises raw video sequences, extracted image frames, and corresponding annotations. 
The original recording sequences, serving as the raw source material, are provided as \texttt{.mp4} video files captured at 100~fps. 
From these sequences, lossless image frames were extracted at a reduced sampling rate of 20~fps and saved in \texttt{.png} format at a resolution of $3680 \times 2160$ pixels. 
The corresponding instance segmentation masks are supplied as \texttt{.txt} files adhering to the standard PyTorch YOLO format. 
Within these files, each line defines a single object instance using the structure \texttt{<class-index> <x1> <y1> ... <xn> <yn>}. 
All polygon coordinates $(x_i, y_i)$ are normalized to a range between 0.0 and 1.0 relative to the image dimensions. 
The class indices are strictly mapped to 0 for \textit{Background}, 1 for \textit{Steel}  and 2 for \textit{Copper}.

\subsection*{Directory Structure}
\begin{figure}[h]
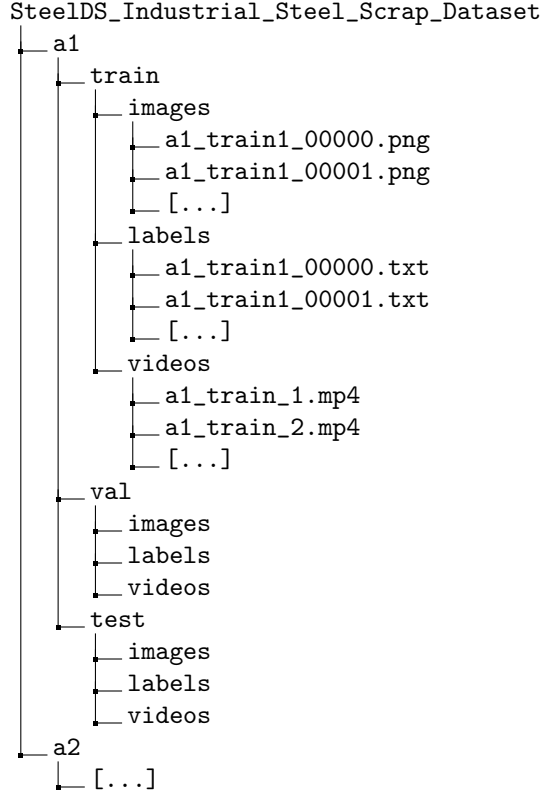

  \centering
  \resizebox{\textwidth}{!}{%
    \begin{minipage}{\textwidth}
    \dirtree{%
    .1 SteelDS\_Industrial\_Steel\_Scrap\_Dataset.
    .2 a1.
    .3 train.
    .4 images.
    .5 a1\_train1\_00000.png.
    .5 a1\_train1\_00001.png.
    .5 [...].
    .4 labels.
    .5 a1\_train1\_00000.txt.
    .5 a1\_train1\_00001.txt.
    .5 [...].
    .4 videos.
    .5 a1\_train\_1.mp4.
    .5 a1\_train\_2.mp4.
    .5 [...].
    .3 val.
    .4 images.
    .4 labels.
    .4 videos.
    .3 test.
    .4 images.
    .4 labels.
    .4 videos.
    .2 a2.
    .3 [...].
    }
    \end{minipage}%
  }
  \caption{Hierarchical directory structure of the dataset. Raw videos, extracted frames, and YOLO-formatted annotation files are organized by complexity subset (a1--a5) and functional split (train, val, test).}
  \label{fig:dir_tree}
\end{figure}

The dataset is organized hierarchically by subset (a1--a5) and subsequent data splits (train, val, test), as illustrated in Figure~\ref{fig:dir_tree}. The file naming convention links individual frames to their parent video sequence.

\subsection*{Dataset Statistics and Configurations}

The dataset is intended for multi-task learning applications, supporting object detection, instance segmentation, and classification tasks across varied levels of complexity. 
Each subset was generated through multiple recording runs, typically ranging from two to five per split, utilizing distinct particle layouts to simulate diverse sorting scenarios. 
During acquisition, particles were manually arranged on the conveyor to control spacing and ensure consistent object orientation.

Specifically, the training splits for subsets a1–a3 consist of five recording runs each, while subset a4 includes two. Validation and test splits comprise two to three independent runs to support robust and reproducible model evaluation. 
Table~\ref{tab:runs_distances} details the quantity of runs, inter-particle distances, and frame counts per subset.

The subsets progressively scale in material diversity and spatial complexity:

    \paragraph{Subsets a1 to a3 (Controlled Spacing):} These subsets maintain a strict minimum of 1~cm spacing between objects to prevent occlusions. Subset a1 includes only basic steel (S1–S2) and copper (C1) particles. Subset a2 expands the object pool, and a3 incorporates the complete range of materials, featuring all five steel types (S1–S5) and three copper types (C1–C3). Cumulatively, the training splits of these three subsets contain 315 unique particles, with frame counts ranging from 3,831 in a1 to 6,715 in a3.
    \paragraph{Subset a4 (Dense Clustering):} Introduces a more complex setting with no spacing between particles. This configuration increases natural occlusion and overlap, creating a challenging segmentation and tracking environment. The training split includes 246 steel and 69 copper particles distributed over 1,025 frames.
    \paragraph{Subset a5 (Holistic Evaluation):} Serves as a comprehensive evaluation benchmark containing all material types (S1–S5 and C1–C3). It combines both spatial configurations: standard inter-particle spacing (3,977 frames) and dense clustering with no spacing (1,233 frames). Together, a5 includes 396 steel and 101 copper particles, making it the most diverse portion of the dataset.

Table~\ref{tab:dataset_detail_information} illustrates the specific count of particles per dataset, where columns S1--S5 and C1--C3 represent the material categories as defined in Table~\ref{tab:comparison}.

\begin{table}[!ht]
\centering  
\caption{Detailed particle counts per dataset split, categorized by material and heuristic size classes (S1--S5 for steel, C1--C3 for copper). Note: These size subdivisions are provided solely to illustrate the physical variance and dimension of the objects for the reader. They are not explicitly encoded as distinct classes within the dataset's ground truth segmentation masks, which only differentiate between binary categories (Steel and Copper).}
\label{tab:dataset_detail_information}

\begin{tabular}{
  p{1.6cm}  
  >{\centering\arraybackslash}p{1cm}
  >{\centering\arraybackslash}p{0.7cm}
  >{\centering\arraybackslash}p{0.7cm}
  >{\centering\arraybackslash}p{0.7cm}
  >{\centering\arraybackslash}p{0.7cm}
  >{\centering\arraybackslash}p{0.7cm}
  >{\centering\arraybackslash}p{0.7cm}
  >{\centering\arraybackslash}p{0.7cm}
  >{\centering\arraybackslash}p{0.7cm}
  >{\centering\arraybackslash}p{0.7cm}
}

 \hline
          & \textbf{Set} & \textbf{S1} & \textbf{S2} & \textbf{S3} & \textbf{S4} & \textbf{S5} & \textbf{C1} & \textbf{C2} & \textbf{C3}\\ \hline
\textbf{Dataset a1} & Train & 31 & 65 &   &   &   & 11 &   &   \\
                   & Val.  & 7  & 15 &   &   &   & 3  &   &   \\
                   & Test  & 7  & 15 &   &   &   & 3  &   &   \\
\hline
\textbf{Dataset a2} & Train & 31 & 65 & 40  &  46 &   & 11 & 33  &   \\
                   & Val.  & 7  & 15 & 16  &  17 &   & 3  &  7  &   \\
                   & Test  & 7  & 15 & 16  &  17 &   & 3  &  7  &   \\

\hline
\textbf{Dataset a3} & Train & 31 & 65 & 40  &  46 & 64  & 11 & 33  &  25 \\
                   & Val.  & 7  & 15 & 16  &  17 & 20  & 3  &  7  &   6\\
                   & Test  & 7  & 15 & 16  &  17 & 20  & 3  &  7  &   6\\

\hline
\textbf{Dataset a4} & Train & 31 & 65 & 40  &  46 & 64  & 11 & 33  &  25 \\
                   & Val.  & 7  & 15 & 16  &  17 & 20  & 3  &  7  &   6\\
                   & Test  & 7  & 15 & 16  &  17 & 20  & 3  &  7  &   6\\

\hline
\textbf{Dataset a5} & All-1 & 45 & 95 & 72  &  80 & 104  & 17 & 47  &  37 \\
                   & All-2 & 45 & 95 & 72  &  80 & 104  & 17 & 47  &  37 \\

\hline
\end{tabular}
\end{table}

\section*{Technical Validation}
This section is structured as follows. First, we detail the baseline model training configurations and define the unified evaluation metrics. 
Second, we present a comprehensive evaluation of the dataset, analyzing the performance of state-of-the-art object detection and instance segmentation models alongside their computational efficiency. 
Subsequently, we outline previous applications of the dataset to demonstrate its practical utility in advanced perception tasks. 
Finally, we conclude with a critical discussion of open challenges in automated material perception and provide information on code availability to ensure full reproducibility.

To assess the practical value of the dataset for real-world applications in automated recycling, we conducted baseline experiments using a state-of-the-art object detection and instance segmentation framework. 
The goal of this evaluation was to demonstrate the learnability of key tasks such as material classification and segmentation. 
Based on the foundational YOLO architecture introduced by Redmon et al.~\cite{redmon2016you}, several modern iterations have been developed. In this work, the selected segmentation and classification baselines include YOLOv8n-seg~\cite{yolo8}, YOLO11n-seg~\cite{yolo11}, YOLO12n-seg~\cite{yolo12}, YOLO26n-seg~\cite{yolo26}, and Mask R-CNN~\cite{he2017mask}.

\subsection*{Model Training}

To establish robust baselines, models were trained across 5 different random seeds with deterministic execution enabled to ensure reproducibility on a standardized Ubuntu 24.04 environment equipped with an NVIDIA RTX 4090 GPU (24 GB VRAM).

The YOLO models were trained using the AdamW optimizer with a batch size of 8, limited by hardware VRAM constraints, and processed using 4 CPU worker threads. 
Automatic Mixed Precision was enabled to optimize memory consumption. 
The optimization schedule utilized an initial learning rate of 0.01 and a weight decay of 0.0005. 
Training included a warmup phase of 3 epochs with an initial bias learning rate of 0.00001. 
To prevent overfitting, early stopping was configured with a patience of 3 epochs.
Data augmentation was strictly controlled to reflect realistic conveyor belt conditions: left-right flipping was applied with a 50\% probability, while up-down flipping and mosaic augmentation were completely disabled to maintain the physical scale and orientation of the objects. 
For high-fidelity instance segmentation, mask overlap was disabled and masks were processed at full resolution.

The Mask R-CNN models (ResNet-50 backbone) were trained using the AdamW optimizer with a reduced batch size of 4, constrained by hardware VRAM limitations, and processed using 4 CPU worker threads. 
Deterministic execution was enforced via manual seed initialization to ensure reproducibility.
The optimization schedule utilized a constant learning rate of 0.001 and a weight decay of 0.0005. 
To stabilize the training process, gradient clipping was applied with a maximum norm of 5.0. 
During the training loop, execution stability was maintained by dynamically filtering out instances lacking valid bounding boxes and skipping batches that resulted in NaN losses.
Unlike the YOLO configuration, Automatic Mixed Precision, learning rate warmup, and early stopping mechanisms were not utilized. 
The models were trained for a fixed duration of 20 epochs. 
Data augmentation was completely disabled for the Mask R-CNN baseline to establish a raw performance benchmark.

\subsection*{Evaluation Metrics}

Reported metrics strictly adhere to the COCO evaluation standards~\cite{lin2014microsoft} and include Mean Average Precision at a threshold of 0.50 (mAP50), the primary challenge metric averaged across multiple thresholds from 0.50 to 0.95 (mAP50-95), and Mean Average Recall given 100 detections per image (MAR100). 
Performance is further disaggregated by object scale into Medium ($32^2 \leq \text{area} < 96^2$ pixels) and Large ($\text{area} \geq 96^2$ pixels) categories. 
Metrics for small objects (area $< 32^2$ pixels) are structurally omitted, as the dataset inherently lacks ground-truth instances in this dimension, rendering the metric mathematically undefined.

\paragraph{Methodological Adjustment for Area Calculation:} 
Standard area estimations based on bounding box heuristics can introduce discrepancies between frameworks. 
To guarantee exact geometric evaluation, the area of ground truth objects was explicitly computed by summing the True pixels of the Boolean masks.

\paragraph{Fitness Metric Formulation:} To provide a holistic scalar for model convergence and performance, we utilize a weighted fitness metric adapted from the Ultralytics framework. 
It balances lenient and strict precision requirements:
\begin{equation}
    Fitness = 0.1 \times mAP^{0.50} + 0.9 \times mAP^{0.50:0.95} \, .
\label{eq:fitness}
\end{equation}
In this evaluation, three variants are reported to isolate modality performance: Fitness (B) computed exclusively on bounding box precision, Fitness (M) for segmentation mask precision, and Fitness (C) representing the arithmetic sum of both modalities:
\begin{equation}
    Fitness (C) = Fitness (B) + Fitness (M) \, .
\end{equation}

\subsection*{Evaluation Results}
Results are reported as Mean $\pm$ Standard Deviation in percentages (\%) over 5 seeds. 
The empirical evaluation is summarized in three corresponding tables: Table \ref{tab:fitness_results} provides the aggregated Fitness scores (Eq.~\ref{eq:fitness}), Table~\ref{tab:avg_model_results} shows the average performance across all models, Table \ref{tab:det_results} presents the bounding box detection (B) metrics, and Table \ref{tab:seg_results} details the instance segmentation (M) performance. 
The evaluation covers overall and class-specific mean Average Precision (mAP) across various thresholds, alongside size-dependent breakdowns (Small, Medium, Large) to quantify spatial detection capabilities. Missing values (`---') signify scale categories absent from the ground-truth data, rendering the specific metric mathematically undefined.

\paragraph{Model Architecture Comparison}
As seen in Table~\ref{tab:fitness_results}, the Mask R-CNN (ResNet-50 backbone) consistently outperformed the single-stage YOLO nano architectures (YOLOv8n, YOLO11n, YOLO12n, YOLO26n) across all evaluated subsets. Specifically, as detailed in the `Fitness (C)' rows for datasets a1, a2, and a3, Mask R-CNN achieved $189.65$ on subset a1 (compared to a maximum of $168.74$ by YOLOv8n), $184.86$ on a2 (vs. $171.45$ by YOLO26n), and $182.55$ on a3 (vs. $168.17$ by YOLO26n). 
This performance gap is structurally expected due to architectural trade-offs in feature extraction and spatial precision. 
Regarding feature extraction, Mask R-CNN utilizes a large, two-stage ResNet-50 backbone, capturing richer hierarchical features. 
In contrast, the YOLO nano variants are explicitly parameterized for high-throughput edge inference, sacrificing representational capacity for speed and lower VRAM consumption. 
Furthermore, in terms of spatial precision, the Mask R-CNN preserves exact spatial feature map alignments. 
Conversely, YOLO architectures rely on single-stage, prototype-based mask generation, which systematically struggles with fine boundary details at strict IoU thresholds, resulting in significantly lower mAP50-95 scores.
Additionally, a steady decline in the YOLO models' `Fitness (M)' scores is clearly visible when moving sequentially from subset a1 to a2, and down to a3, where the performance reaches its lowest point. 
This continuous drop in instance segmentation precision demonstrates that the structural complexity of the sub-datasets increases progressively from a1 through to a3.

\begin{table*}[t]
\centering
\caption{Fitness Results (Mean $\pm$ SD over 5 Seeds in \%). 
Evaluated metrics: 
(C)~Combined, 
(B)~Bounding Box, 
(M)~Instance Segmentation.}
\label{tab:fitness_results}
\resizebox{\textwidth}{!}{%
\begin{tabular}{@{}cccccc@{}}
\toprule
 & \textbf{YOLOv8n} & \textbf{YOLO11n} & \textbf{YOLO12n} & \textbf{YOLO26n} & \textbf{Mask R-CNN} \\
\midrule
\multicolumn{6}{c}{\textbf{Dataset a1}} \\
\midrule
Fitness (B) & 88.29 $\pm$ 4.28 & 85.28 $\pm$ 6.87 & 85.40 $\pm$ 5.90 & 84.52 $\pm$ 6.62 & 92.98 $\pm$ 0.65 \\
Fitness (M) & 80.45 $\pm$ 3.36 & 78.10 $\pm$ 4.91 & 77.43 $\pm$ 4.97 & 81.55 $\pm$ 2.81 & 96.66 $\pm$ 0.23 \\
Fitness (C) & 168.74 $\pm$ 7.18 & 163.38 $\pm$ 11.55 & 162.82 $\pm$ 10.85 & 166.07 $\pm$ 9.35 & \textbf{189.65 $\pm$ 0.86} \\
\midrule
\multicolumn{6}{c}{\textbf{Dataset a2}} \\
\midrule
Fitness (B) & 91.68 $\pm$ 0.52 & 87.82 $\pm$ 2.70 & 90.27 $\pm$ 0.91 & 91.77 $\pm$ 2.60 & 93.47 $\pm$ 1.03 \\
Fitness (M) & 78.13 $\pm$ 0.55 & 75.93 $\pm$ 1.66 & 77.92 $\pm$ 0.34 & 79.68 $\pm$ 1.51 & 91.39 $\pm$ 0.91 \\
Fitness (C) & 169.80 $\pm$ 1.00 & 163.75 $\pm$ 4.34 & 168.19 $\pm$ 1.20 & 171.45 $\pm$ 4.07 & \textbf{184.86 $\pm$ 1.91} \\
\midrule
\multicolumn{6}{c}{\textbf{Dataset a3}} \\
\midrule
Fitness (B) & 88.08 $\pm$ 1.52 & 87.78 $\pm$ 0.52 & 87.32 $\pm$ 2.82 & 91.78 $\pm$ 2.31 & 90.14 $\pm$ 0.60 \\
Fitness (M) & 73.01 $\pm$ 1.74 & 72.67 $\pm$ 0.85 & 73.33 $\pm$ 1.82 & 76.39 $\pm$ 0.44 & 92.41 $\pm$ 0.56 \\
Fitness (C) & 161.09 $\pm$ 3.20 & 160.45 $\pm$ 1.24 & 160.65 $\pm$ 4.62 & 168.17 $\pm$ 2.61 & \textbf{182.55 $\pm$ 1.11} \\
\bottomrule
\end{tabular}
}
\end{table*}

\paragraph{Performance and Class-Specific Disparities}
As detailed in Table~\ref{tab:avg_model_results}, bounding box detection (Box) systematically outperforms instance segmentation (Seg) across all subsets, a structurally expected outcome given the higher complexity of pixel-level masking. For instance, the overall mAP50-95 on Subset a1 is 86.63\% for detection compared to 81.67\% for segmentation. 
Furthermore, object scale dictates model efficacy. Large objects are reliably processed across all tasks (e.g., Box mAP Large ranging from 87.15\% to 90.65\%), whereas medium-sized objects cause severe performance drops (e.g., Seg mAP Medium dropping to 29.03\% on Subset a1).

Regarding material classes, performance disparities are highly task-dependent. 
In bounding box detection, Steel consistently outperforms Copper (e.g., Subset a1: 90.60\% vs. 82.66\% mAP50-95). However, in instance segmentation, both classes perform similarly, with marginal differences (e.g., Subset a3: 75.90\% for Steel vs. 75.53\% for Copper). 
This behavior is driven by a combination of dataset imbalance and distinct physical properties. 
The dataset contains a disproportionate amount of bulkier, rigid steel structures, establishing a strong optimization bias and facilitating reliable bounding box proposals. 
Conversely, copper surfaces exhibit high specular reflectance under industrial lighting, disrupting gradient-based feature extraction. 
Morphologically, copper waste frequently appears as tangled wire bundles with thin, multidirectional strands. The presence of partial plastic insulation or mixed-material assemblies (e.g., steel pins embedded in copper nets) further degrades mask coherence. This complexity explains why bounding box detection succeeds on copper, while exact boundary delineation during segmentation fails, lowering copper's segmentation metrics to match steel.

Despite superior detection rates, the models still exhibit specific failure modes on steel objects. Topological errors occur where internal background spaces of ring-shaped objects are erroneously masked as part of the object. Additionally, long, twisted rods and crumpled metal sheets with torn edges are frequently misclassified or fragmented into multiple smaller instances.

\begin{table*}[t]
\centering
\caption{Average Performance Across All Models (YOLOv8n, YOLO11n, YOLO12n, YOLO26n, Mask R-CNN) in \%.}
\label{tab:avg_model_results}
\resizebox{\textwidth}{!}{%
\begin{tabular}{@{}llcccccc@{}}
\toprule
 & \textbf{Dataset} & \textbf{mAP50-95} & \textbf{mAP50-95} & \textbf{mAP50-95} & \textbf{mAP} & \textbf{mAP} & \textbf{Recall} \\
& & \textbf{(Overall)} & \textbf{(Steel)} & \textbf{(Copper)} & \textbf{(Medium)} & \textbf{(Large)} & \\
\midrule
\multirow{3}{*}{\rotatebox[origin=c]{90}{Box (B)}}
& Subset a1 & 86.63 & 90.60 & 82.66 & 44.44 & 87.15 & 90.67 \\
& Subset a2 & \textbf{90.02} & 91.56 & \textbf{88.49} & 59.64 & \textbf{90.65} & \textbf{93.02} \\
& Subset a3 & 88.43 & \textbf{92.09} & 84.77 & \textbf{69.07} & 89.75 & 91.79 \\
\midrule
\multirow{3}{*}{\rotatebox[origin=c]{90}{Seg (M)}}
& Subset a1 & \textbf{81.67} & \textbf{82.27} & \textbf{81.07} & 29.03 & \textbf{82.40} & \textbf{85.16} \\
& Subset a2 & 79.39 & 77.80 & 80.98 & 37.65 & 80.28 & 82.69 \\
& Subset a3 & 75.72 & 75.90 & 75.53 & \textbf{41.16} & 77.83 & 79.03 \\
\bottomrule
\end{tabular}
}
\end{table*}

\begin{table*}[t]
\centering
\caption{Bounding Box Detection Results (B) (Mean $\pm$ SD over 5 Seeds in \%).}
\label{tab:det_results}
\resizebox{\textwidth}{!}{%
\begin{tabular}{@{}lccccc@{}}
\toprule
\textbf{Metric} & \textbf{YOLOv8n} & \textbf{YOLO11n} & \textbf{YOLO12n} & \textbf{YOLO26n} & \textbf{Mask R-CNN} \\
\midrule
\multicolumn{6}{c}{\textbf{Dataset a1}} \\
\midrule
mAP50 (Overall) & 92.90 $\pm$ 3.71 & 90.59 $\pm$ 5.17 & 89.74 $\pm$ 6.32 & 94.46 $\pm$ 2.83 & 98.72 $\pm$ 0.09 \\
mAP50-95 (Overall) & 87.78 $\pm$ 4.51 & 84.69 $\pm$ 7.12 & 84.91 $\pm$ 5.86 & 83.42 $\pm$ 7.06 & 92.35 $\pm$ 0.72 \\
\cmidrule{1-6}
mAP50 (Steel) & 98.41 $\pm$ 0.43 & 97.00 $\pm$ 1.69 & 97.29 $\pm$ 1.46 & 97.27 $\pm$ 1.17 & 98.83 $\pm$ 0.05 \\
mAP50 (Copper) & 87.39 $\pm$ 7.01 & 84.17 $\pm$ 9.06 & 82.19 $\pm$ 11.32 & 91.65 $\pm$ 4.83 & 98.62 $\pm$ 0.15 \\
mAP50-95 (Steel) & 92.66 $\pm$ 3.97 & 90.44 $\pm$ 6.27 & 91.95 $\pm$ 1.26 & 85.13 $\pm$ 5.90 & 92.82 $\pm$ 1.06 \\
mAP50-95 (Copper) & 82.90 $\pm$ 6.68 & 78.95 $\pm$ 9.58 & 77.88 $\pm$ 10.65 & 81.70 $\pm$ 8.59 & 91.87 $\pm$ 0.72 \\
\cmidrule{1-6}
mAP Small & --- & --- & --- & --- & --- \\
mAP Medium & 39.42 $\pm$ 8.74 & 38.31 $\pm$ 4.82 & 45.19 $\pm$ 5.80 & 35.73 $\pm$ 12.00 & 63.56 $\pm$ 4.21 \\
mAP Large & 88.31 $\pm$ 4.51 & 85.23 $\pm$ 7.13 & 85.50 $\pm$ 5.77 & 84.08 $\pm$ 6.99 & 92.62 $\pm$ 0.66 \\
\cmidrule{1-6}
Recall & 90.24 $\pm$ 4.10 & 88.53 $\pm$ 6.49 & 87.93 $\pm$ 5.57 & 91.20 $\pm$ 3.82 & 95.46 $\pm$ 0.42 \\
\midrule
\multicolumn{6}{c}{\textbf{Dataset a2}} \\
\midrule
mAP50 (Overall) & 94.62 $\pm$ 0.89 & 92.93 $\pm$ 1.91 & 95.51 $\pm$ 0.79 & 97.51 $\pm$ 0.99 & 97.69 $\pm$ 0.52 \\
mAP50-95 (Overall) & 91.35 $\pm$ 0.48 & 87.25 $\pm$ 2.82 & 89.69 $\pm$ 1.06 & 91.13 $\pm$ 2.79 & 90.69 $\pm$ 0.95 \\
\cmidrule{1-6}
mAP50 (Steel) & 96.89 $\pm$ 0.68 & 96.88 $\pm$ 0.54 & 97.06 $\pm$ 0.41 & 97.85 $\pm$ 0.72 & 97.41 $\pm$ 0.39 \\
mAP50 (Copper) & 92.35 $\pm$ 1.20 & 88.98 $\pm$ 3.42 & 93.96 $\pm$ 1.24 & 97.18 $\pm$ 1.34 & 97.97 $\pm$ 0.68 \\
mAP50-95 (Steel) & 93.90 $\pm$ 0.41 & 90.98 $\pm$ 2.01 & 90.85 $\pm$ 1.81 & 91.48 $\pm$ 2.78 & 90.57 $\pm$ 0.62 \\
mAP50-95 (Copper) & 88.80 $\pm$ 0.86 & 83.52 $\pm$ 3.93 & 88.53 $\pm$ 0.67 & 90.79 $\pm$ 2.87 & 90.81 $\pm$ 1.39 \\
\cmidrule{1-6}
mAP Small & --- & --- & --- & --- & --- \\
mAP Medium & 62.38 $\pm$ 4.27 & 50.85 $\pm$ 6.77 & 57.14 $\pm$ 4.45 & 58.66 $\pm$ 6.02 & 69.18 $\pm$ 1.43 \\
mAP Large & 91.95 $\pm$ 0.58 & 88.03 $\pm$ 2.94 & 90.39 $\pm$ 0.98 & 91.77 $\pm$ 2.69 & 91.12 $\pm$ 0.98 \\
\cmidrule{1-6}
Recall & 93.59 $\pm$ 0.41 & 90.05 $\pm$ 2.32 & 92.77 $\pm$ 0.51 & 94.68 $\pm$ 1.44 & 93.99 $\pm$ 0.65 \\
\midrule
\multicolumn{6}{c}{\textbf{Dataset a3}} \\
\midrule
mAP50 (Overall) & 92.06 $\pm$ 2.25 & 92.67 $\pm$ 0.90 & 92.84 $\pm$ 2.65 & 97.25 $\pm$ 0.57 & 96.68 $\pm$ 0.58 \\
mAP50-95 (Overall) & 87.63 $\pm$ 1.45 & 87.24 $\pm$ 0.48 & 86.71 $\pm$ 2.85 & 91.17 $\pm$ 2.54 & 89.42 $\pm$ 0.62 \\
\cmidrule{1-6}
mAP50 (Steel) & 96.94 $\pm$ 0.14 & 97.44 $\pm$ 0.41 & 97.65 $\pm$ 0.83 & 98.32 $\pm$ 0.42 & 98.60 $\pm$ 0.12 \\
mAP50 (Copper) & 87.19 $\pm$ 4.59 & 87.90 $\pm$ 1.53 & 88.03 $\pm$ 4.64 & 96.18 $\pm$ 1.02 & 94.76 $\pm$ 1.12 \\
mAP50-95 (Steel) & 92.81 $\pm$ 1.35 & 91.85 $\pm$ 0.75 & 91.74 $\pm$ 1.05 & 92.44 $\pm$ 3.26 & 91.63 $\pm$ 0.49 \\
mAP50-95 (Copper) & 82.46 $\pm$ 3.82 & 82.63 $\pm$ 1.25 & 81.68 $\pm$ 4.77 & 89.90 $\pm$ 1.93 & 87.20 $\pm$ 1.04 \\
\cmidrule{1-6}
mAP Small & --- & --- & --- & --- & --- \\
mAP Medium & 67.95 $\pm$ 7.47 & 70.26 $\pm$ 4.36 & 62.34 $\pm$ 8.75 & 73.06 $\pm$ 2.78 & 71.74 $\pm$ 2.79 \\
mAP Large & 89.13 $\pm$ 1.26 & 88.52 $\pm$ 0.60 & 88.46 $\pm$ 2.50 & 92.22 $\pm$ 2.47 & 90.42 $\pm$ 0.60 \\
\cmidrule{1-6}
Recall & 90.64 $\pm$ 1.83 & 90.21 $\pm$ 0.82 & 90.08 $\pm$ 2.37 & 94.88 $\pm$ 1.70 & 93.12 $\pm$ 0.45 \\
\bottomrule
\end{tabular}
}
\end{table*}

\begin{table*}[t]
\centering
\caption{Instance Segmentation Results (M) (Mean $\pm$ SD over 5 Seeds in \%).}
\label{tab:seg_results}
\resizebox{\textwidth}{!}{%
\begin{tabular}{@{}lccccc@{}}
\toprule
\textbf{Metric} & \textbf{YOLOv8n} & \textbf{YOLO11n} & \textbf{YOLO12n} & \textbf{YOLO26n} & \textbf{Mask R-CNN} \\
\midrule
\multicolumn{6}{c}{\textbf{Dataset a1}} \\
\midrule
mAP50 (Overall) & 92.99 $\pm$ 3.76 & 90.60 $\pm$ 5.17 & 89.74 $\pm$ 6.31 & 94.50 $\pm$ 2.80 & 98.82 $\pm$ 0.28 \\
mAP50-95 (Overall) & 79.06 $\pm$ 3.32 & 76.71 $\pm$ 4.88 & 76.06 $\pm$ 4.82 & 80.11 $\pm$ 2.82 & 96.42 $\pm$ 0.26 \\
\cmidrule{1-6}
mAP50 (Steel) & 98.43 $\pm$ 0.43 & 97.03 $\pm$ 1.68 & 97.29 $\pm$ 1.43 & 97.30 $\pm$ 1.15 & 99.01 $\pm$ 0.43 \\
mAP50 (Copper) & 87.56 $\pm$ 7.11 & 84.17 $\pm$ 9.06 & 82.19 $\pm$ 11.33 & 91.71 $\pm$ 4.76 & 98.62 $\pm$ 0.15 \\
mAP50-95 (Steel) & 79.93 $\pm$ 1.23 & 78.86 $\pm$ 2.43 & 78.80 $\pm$ 0.42 & 78.35 $\pm$ 1.53 & 95.42 $\pm$ 0.44 \\
mAP50-95 (Copper) & 78.19 $\pm$ 6.73 & 74.56 $\pm$ 8.14 & 73.32 $\pm$ 9.63 & 81.86 $\pm$ 4.27 & 97.42 $\pm$ 0.33 \\
\cmidrule{1-6}
mAP Small & --- & --- & --- & --- & --- \\
mAP Medium & 20.38 $\pm$ 6.77 & 16.98 $\pm$ 4.35 & 21.36 $\pm$ 4.31 & 23.99 $\pm$ 6.91 & 62.46 $\pm$ 2.87 \\
mAP Large & 79.88 $\pm$ 3.24 & 77.54 $\pm$ 4.92 & 76.91 $\pm$ 4.85 & 80.95 $\pm$ 2.81 & 96.74 $\pm$ 0.29 \\
\cmidrule{1-6}
Recall & 81.49 $\pm$ 3.32 & 80.40 $\pm$ 4.98 & 79.13 $\pm$ 4.70 & 86.73 $\pm$ 0.38 & 98.06 $\pm$ 0.15 \\
\midrule
\multicolumn{6}{c}{\textbf{Dataset a2}} \\
\midrule
mAP50 (Overall) & 94.64 $\pm$ 0.87 & 92.93 $\pm$ 1.92 & 95.60 $\pm$ 0.79 & 97.63 $\pm$ 0.94 & 97.91 $\pm$ 0.60 \\
mAP50-95 (Overall) & 76.29 $\pm$ 0.54 & 74.04 $\pm$ 1.65 & 75.96 $\pm$ 0.45 & 77.69 $\pm$ 1.59 & 92.97 $\pm$ 1.08 \\
\cmidrule{1-6}
mAP50 (Steel) & 96.92 $\pm$ 0.64 & 96.87 $\pm$ 0.52 & 97.23 $\pm$ 0.45 & 98.08 $\pm$ 0.57 & 97.83 $\pm$ 0.55 \\
mAP50 (Copper) & 92.35 $\pm$ 1.20 & 88.99 $\pm$ 3.42 & 93.96 $\pm$ 1.24 & 97.18 $\pm$ 1.33 & 97.98 $\pm$ 0.69 \\
mAP50-95 (Steel) & 75.36 $\pm$ 0.54 & 73.91 $\pm$ 0.71 & 73.71 $\pm$ 0.80 & 75.11 $\pm$ 1.29 & 90.92 $\pm$ 1.18 \\
mAP50-95 (Copper) & 77.23 $\pm$ 1.03 & 74.18 $\pm$ 2.83 & 78.21 $\pm$ 0.36 & 80.26 $\pm$ 1.95 & 95.02 $\pm$ 1.05 \\
\cmidrule{1-6}
mAP Small & --- & --- & --- & --- & --- \\
mAP Medium & 29.06 $\pm$ 4.73 & 25.24 $\pm$ 4.89 & 27.97 $\pm$ 6.78 & 35.48 $\pm$ 3.94 & 70.48 $\pm$ 1.05 \\
mAP Large & 77.33 $\pm$ 0.53 & 75.10 $\pm$ 1.67 & 77.01 $\pm$ 0.50 & 78.58 $\pm$ 1.60 & 93.37 $\pm$ 1.10 \\
\cmidrule{1-6}
Recall & 79.41 $\pm$ 0.30 & 77.32 $\pm$ 1.57 & 79.51 $\pm$ 0.53 & 81.57 $\pm$ 0.59 & 95.63 $\pm$ 0.71 \\
\midrule
\multicolumn{6}{c}{\textbf{Dataset a3}} \\
\midrule
mAP50 (Overall) & 91.96 $\pm$ 2.25 & 92.31 $\pm$ 0.90 & 92.64 $\pm$ 2.46 & 97.23 $\pm$ 0.57 & 96.69 $\pm$ 0.58 \\
mAP50-95 (Overall) & 70.91 $\pm$ 1.68 & 70.49 $\pm$ 0.86 & 71.18 $\pm$ 1.74 & 74.08 $\pm$ 0.45 & 91.93 $\pm$ 0.55 \\
\cmidrule{1-6}
mAP50 (Steel) & 96.73 $\pm$ 0.35 & 96.74 $\pm$ 0.49 & 97.26 $\pm$ 0.55 & 98.30 $\pm$ 0.41 & 98.62 $\pm$ 0.10 \\
mAP50 (Copper) & 87.18 $\pm$ 4.59 & 87.89 $\pm$ 1.53 & 88.02 $\pm$ 4.63 & 96.16 $\pm$ 1.02 & 94.76 $\pm$ 1.12 \\
mAP50-95 (Steel) & 71.88 $\pm$ 0.29 & 70.89 $\pm$ 0.58 & 71.75 $\pm$ 0.68 & 72.27 $\pm$ 0.94 & 92.72 $\pm$ 0.22 \\
mAP50-95 (Copper) & 69.93 $\pm$ 3.33 & 70.08 $\pm$ 1.58 & 70.62 $\pm$ 2.90 & 75.89 $\pm$ 0.51 & 91.15 $\pm$ 0.98 \\
\cmidrule{1-6}
mAP Small & --- & --- & --- & --- & --- \\
mAP Medium & 34.24 $\pm$ 4.68 & 33.62 $\pm$ 3.82 & 29.70 $\pm$ 4.98 & 38.19 $\pm$ 2.93 & 70.04 $\pm$ 3.82 \\
mAP Large & 73.26 $\pm$ 1.57 & 72.84 $\pm$ 0.83 & 73.74 $\pm$ 1.55 & 76.20 $\pm$ 0.47 & 93.09 $\pm$ 0.51 \\
\cmidrule{1-6}
Recall & 74.19 $\pm$ 1.91 & 73.50 $\pm$ 1.28 & 74.74 $\pm$ 1.70 & 77.91 $\pm$ 0.38 & 94.83 $\pm$ 0.46 \\
\bottomrule
\end{tabular}
}
\end{table*}

\subsection*{Previous Applications of the Dataset}

To demonstrate the broader applicability and complexity of the dataset, it has been utilized as an evaluation benchmark in several applied machine learning studies. 
Specific applications include PASTA~\cite{neubauer2026pasta}, a system for evaluating weakly supervised anomaly and target detection pipelines utilizing Vision Transformer patch aggregation, IBIS~\cite{neubauer2025ibis}, a framework for the assessment of sparse instance segmentation models, and FOST~\cite{neubauer2024semi} for validating a semi-autonomous video annotation and tracking tool utilizing optical flow for rapid object segmentation.

\subsection*{Discussion of Open Challenges and Issues}

\begin{figure}[t]
  \centering
  \includegraphics[width=0.9\textwidth]{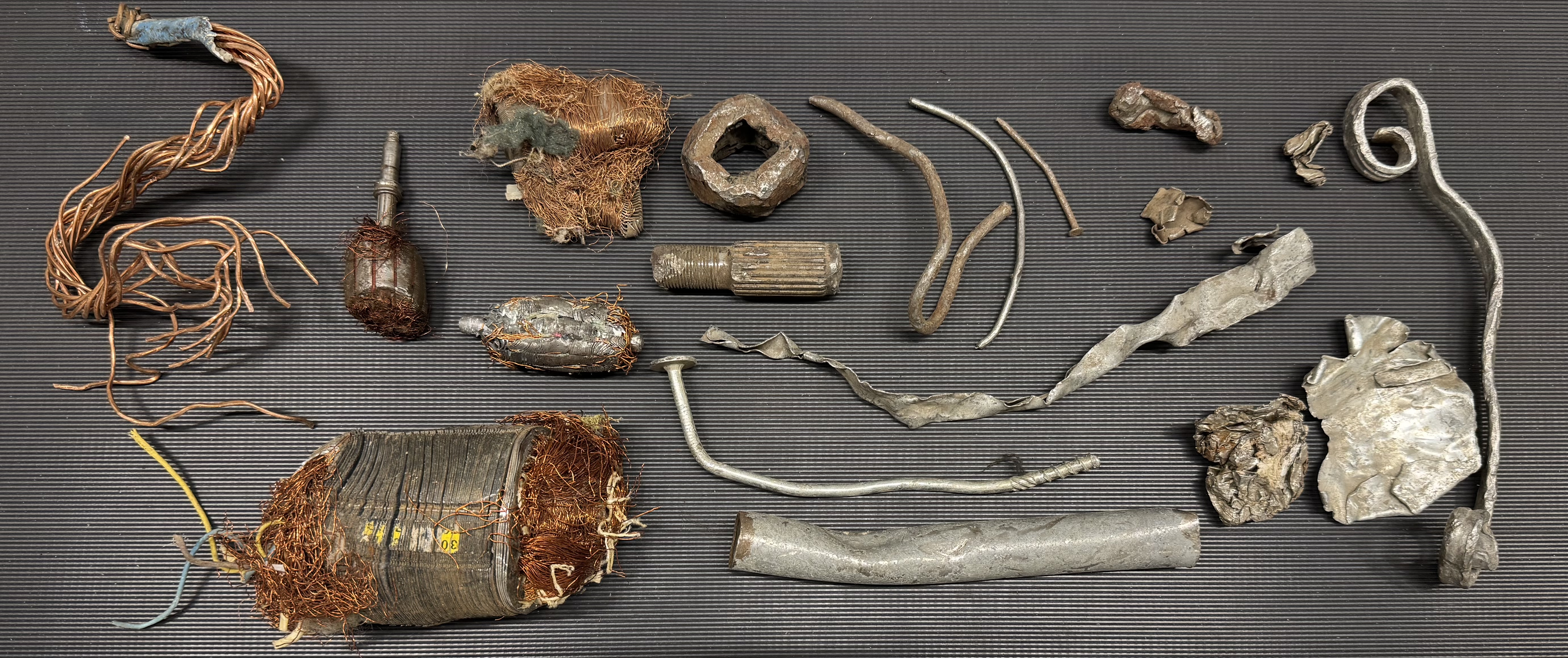} 
  \caption{Primary open challenges in material segmentation. Left: Copper objects with complex wires, plastic insulation, and mixed materials. Right: Steel objects with long shapes, holes, and torn edges.}
  \label{fig:open_challenges}
\end{figure}

While baseline evaluations confirm the dataset's validity, the automated perception of shredded E40 scrap presents specific challenges for current instance segmentation models. 
These issues, illustrated in Figure~\ref{fig:open_challenges}, are categorized as follows:

\begin{itemize}
    \item \textbf{Copper Objects (Fig.~\ref{fig:open_challenges}, left):}
    \begin{itemize}
        \item \textit{Complex shapes:} Wire bundles have thin strands pointing in all directions, making it hard to draw exact masks.
        \item \textit{Hidden parts:} Copper is often covered by plastic insulation.
        \item \textit{Mixed materials:} Some objects are mostly steel but have thin copper wires inside, or have steel pins sticking out of a copper net.
    \end{itemize}
    \item \textbf{Steel Objects (Fig.~\ref{fig:open_challenges}, right):}
    \begin{itemize}
        \item \textit{Long and thin shapes:} Thin, twisted rods or nails are often wrongly split into multiple smaller objects by the model.
        \item \textit{Holes:} Ring-shaped objects with empty spaces inside are often segmented only by their outer edge. This means the background inside the hole is wrongly marked as part of the object.
        \item \textit{Broken structures:} Thin metal sheets are often crumpled into lumps or have highly torn edges.
    \end{itemize}
    \item \textbf{High Size Variance:} Both steel and copper pieces vary greatly in size. Very small scrap pieces are hard to detect on the conveyor belt because they reach the limits of the camera resolution.
\end{itemize}

\begin{figure}[t]
  \centering
  \begin{tabular}{c c c c}
    & \textbf{Subset a1} & \textbf{Subset a2} & \textbf{Subset a3} \\[0.15cm] 
    
    \smash{\rotatebox[origin=c]{90}{\textbf{Ground Truth}}} &
    \includegraphics[width=0.32\textwidth, valign=c]{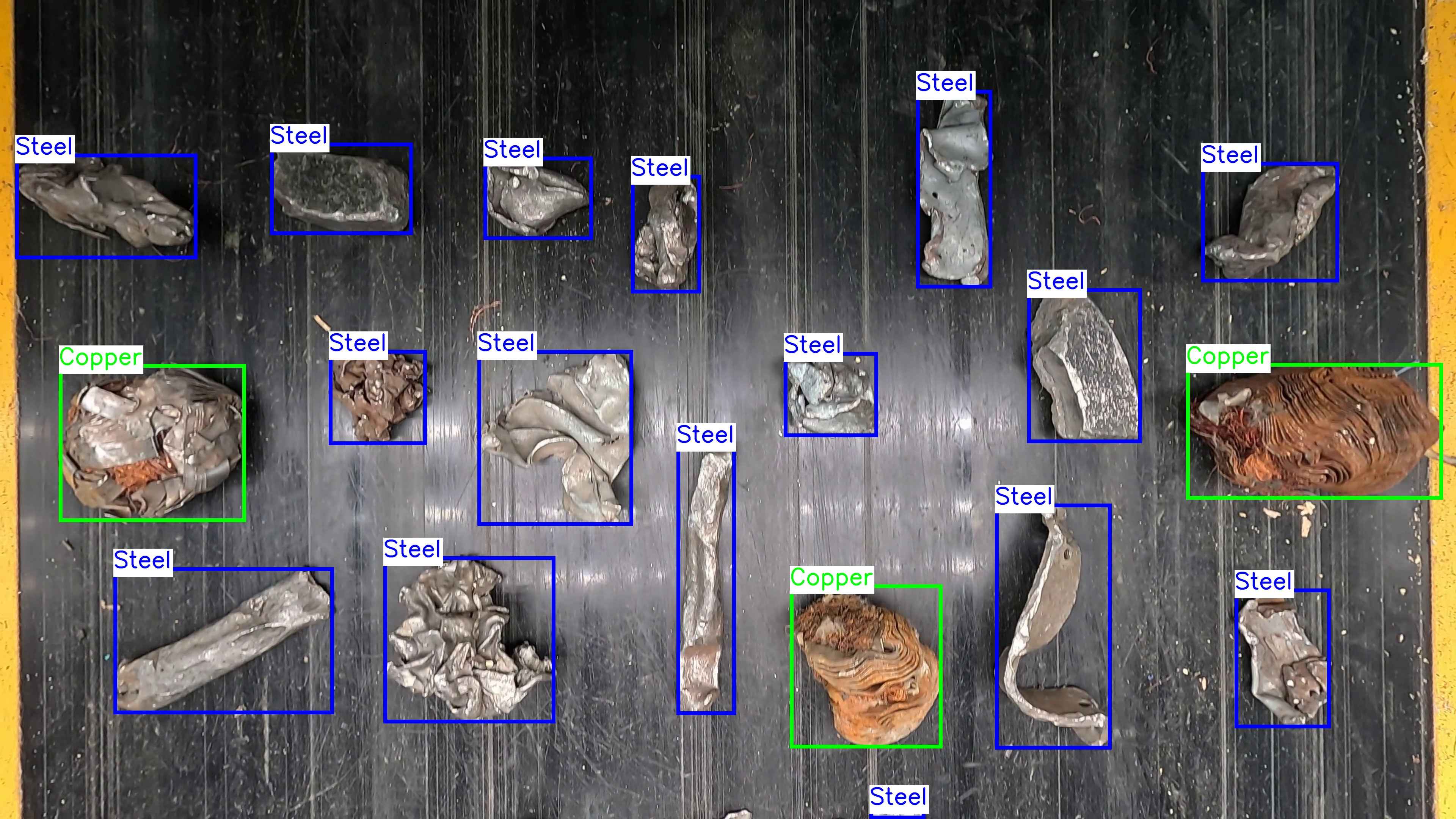} &
    \includegraphics[width=0.32\textwidth, valign=c]{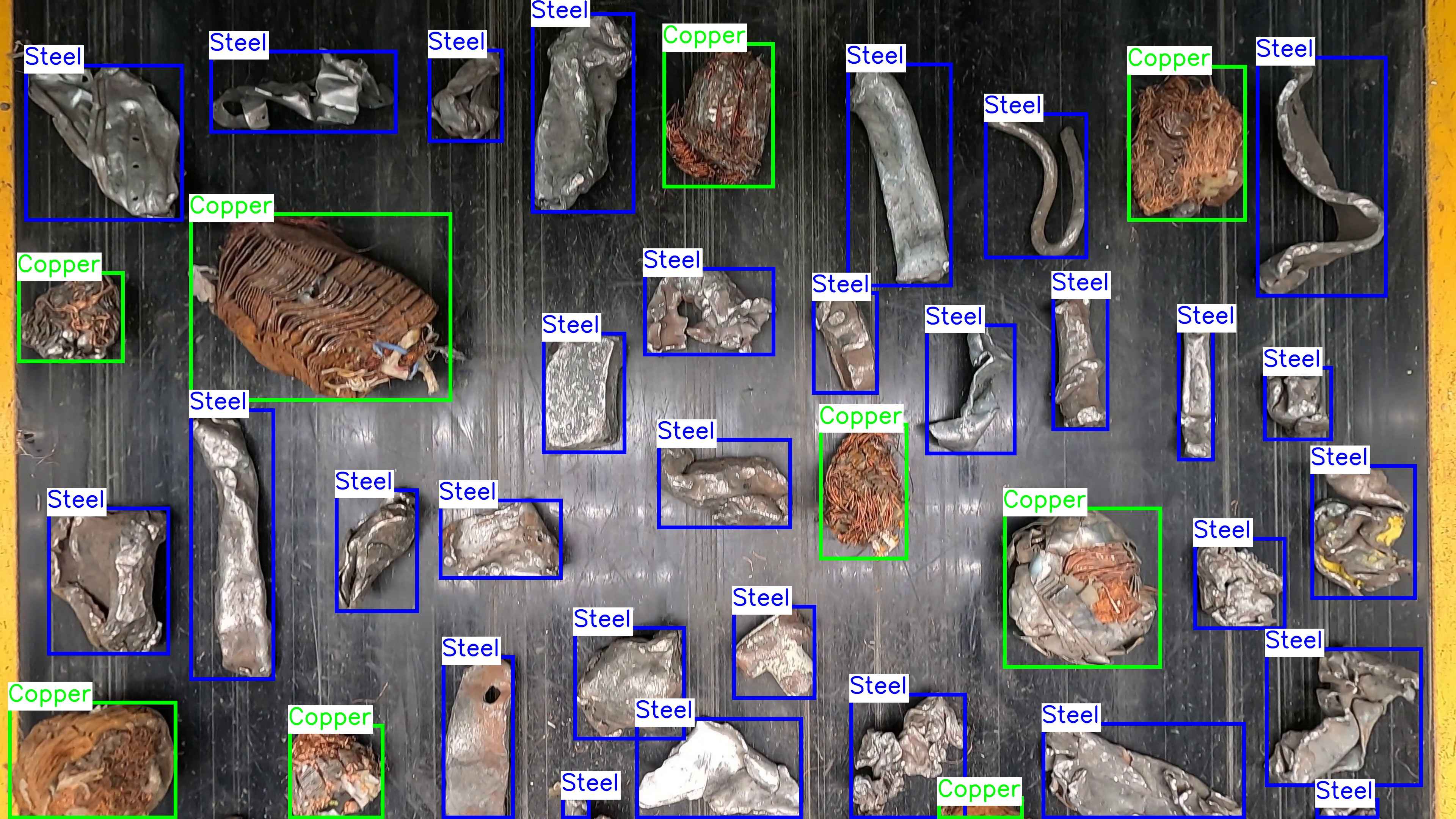} &
    \includegraphics[width=0.32\textwidth, valign=c]{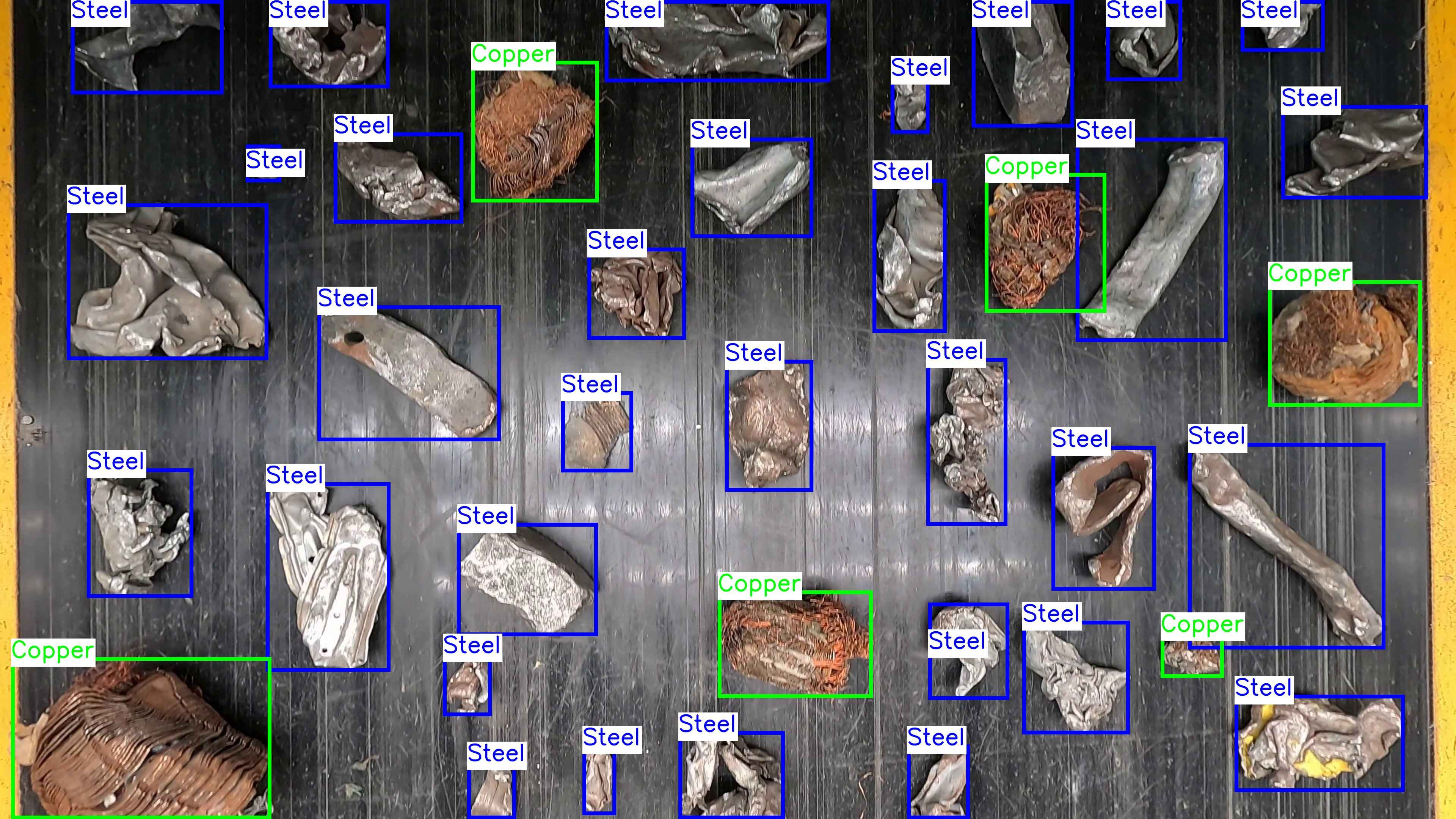} \\[1.5cm]
    
    \smash{\rotatebox[origin=c]{90}{\textbf{YOLOv8n}}} &
    \includegraphics[width=0.32\textwidth, valign=c]{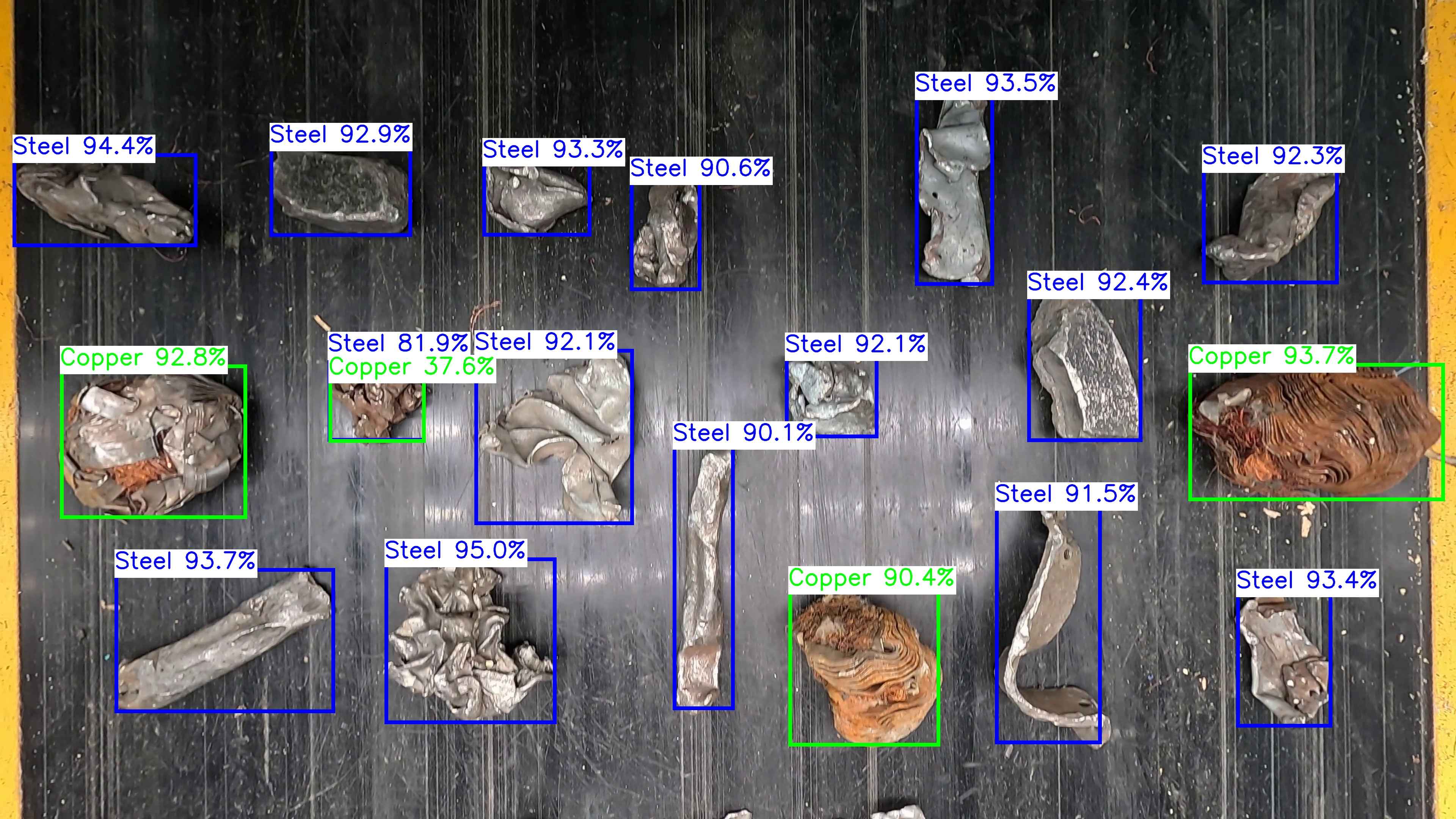} &
    \includegraphics[width=0.32\textwidth, valign=c]{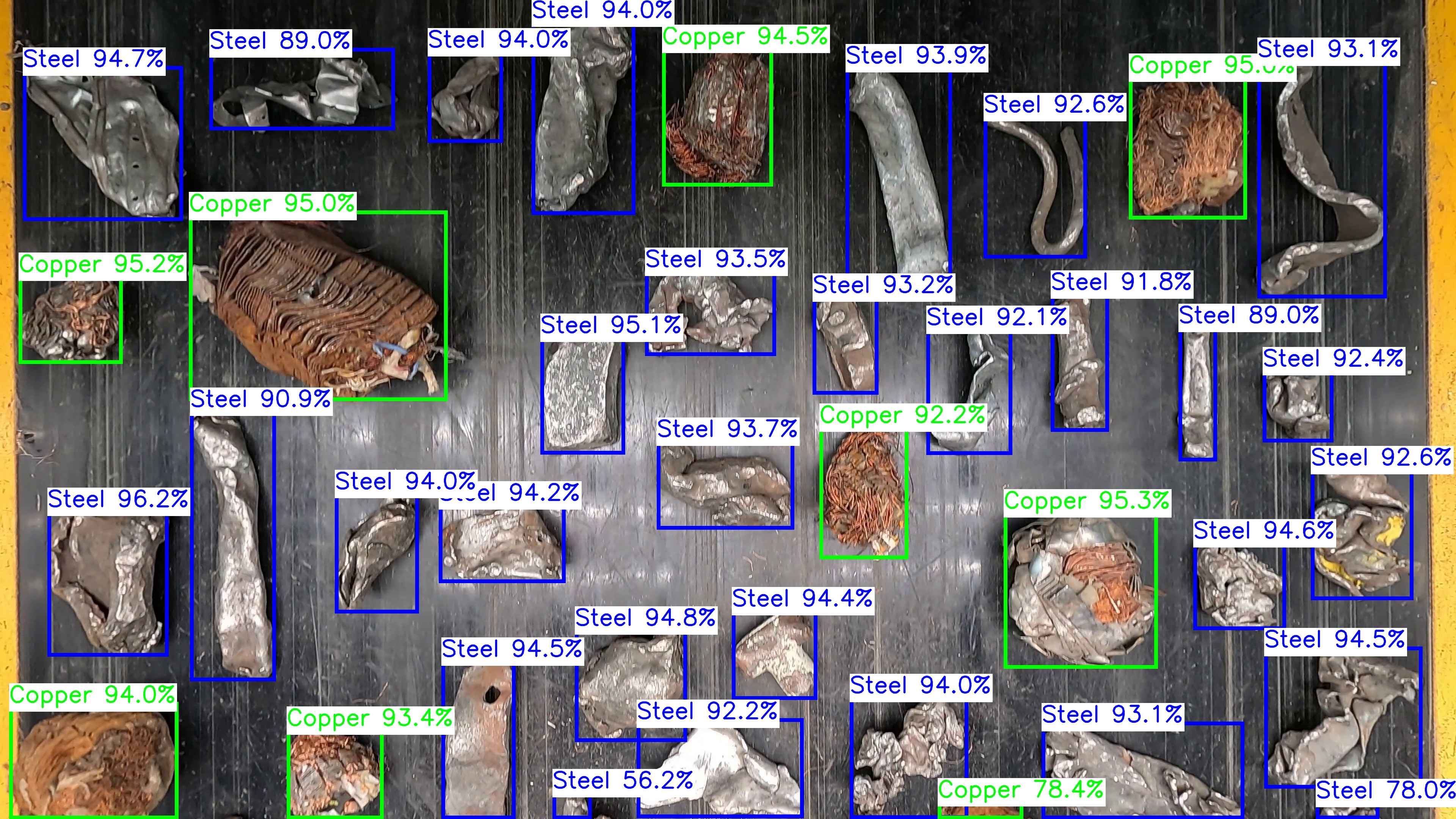} &
    \includegraphics[width=0.32\textwidth, valign=c]{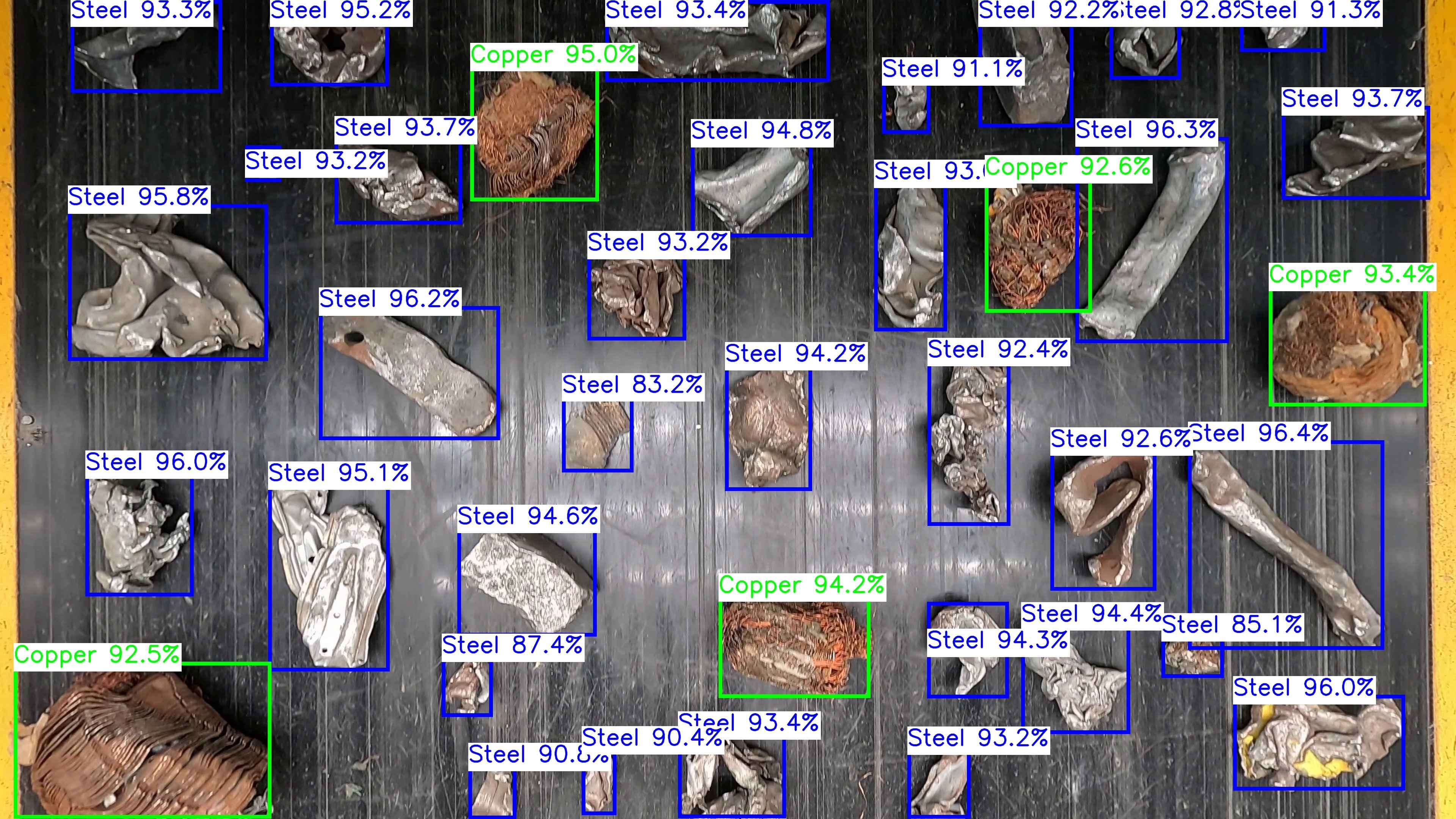} \\
    [1.5cm]
    
    \smash{\rotatebox[origin=c]{90}{\textbf{YOLO11n}}} &
    \includegraphics[width=0.32\textwidth, valign=c]{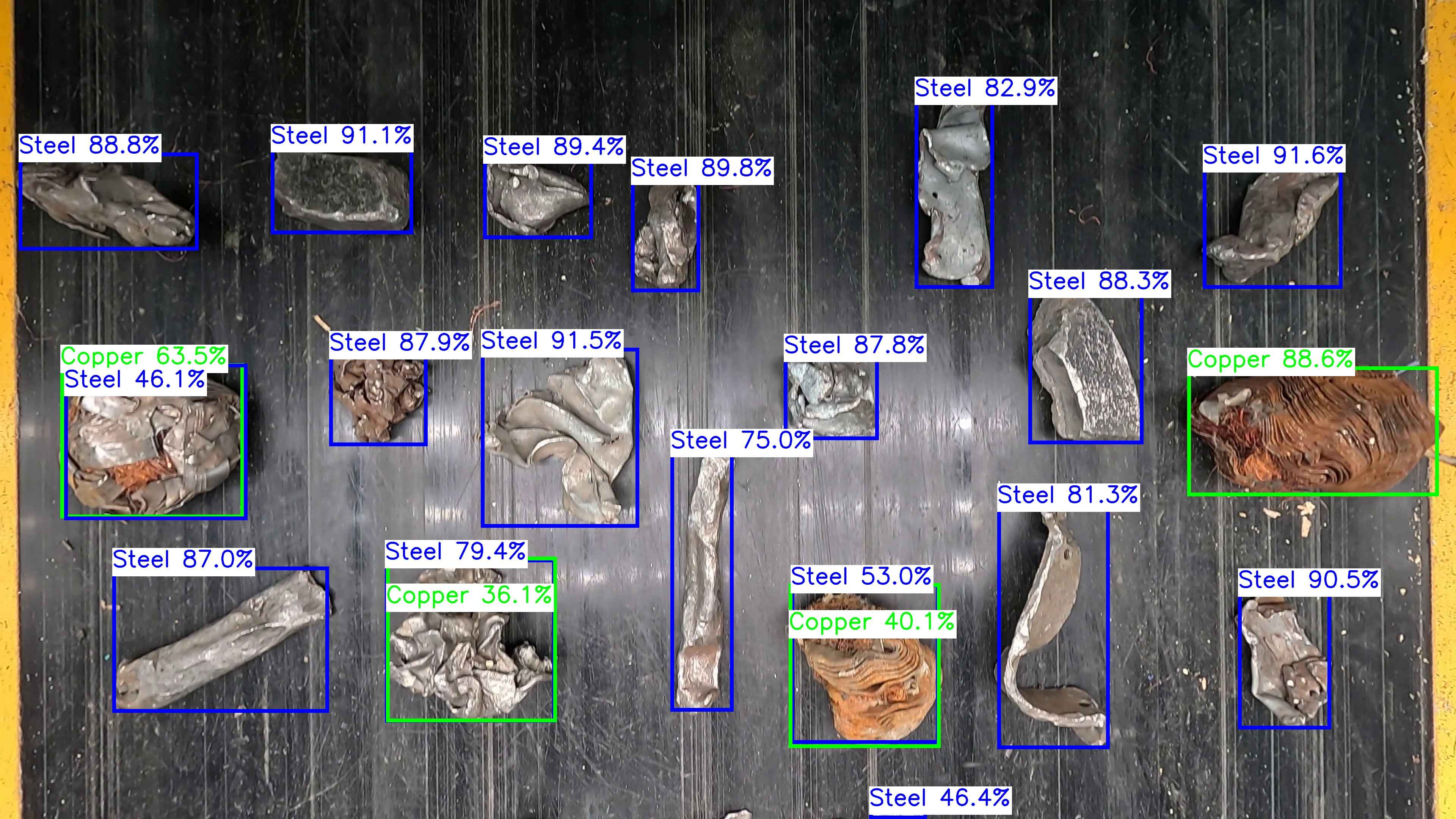} &
    \includegraphics[width=0.32\textwidth, valign=c]{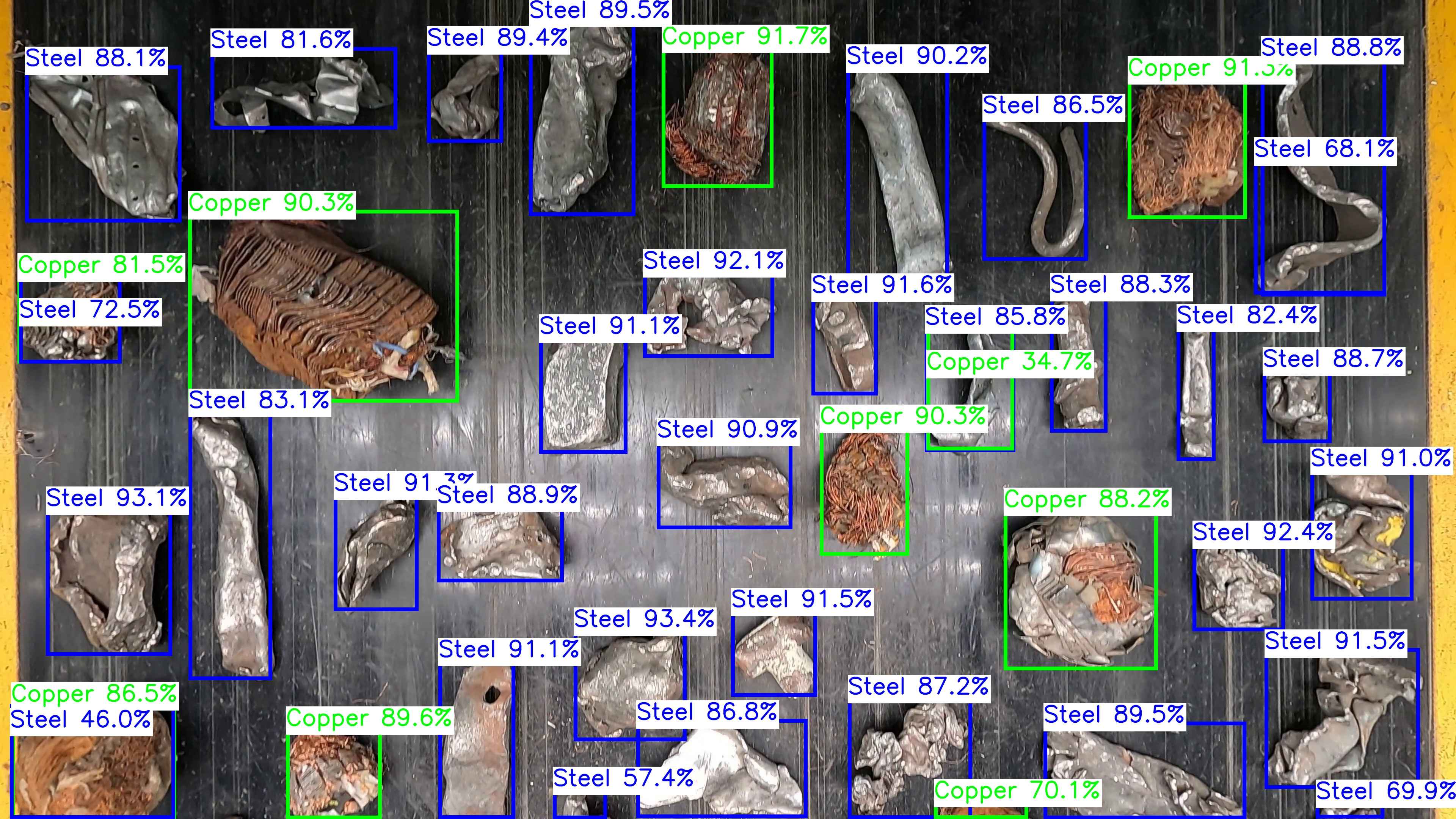} &
    \includegraphics[width=0.32\textwidth, valign=c]{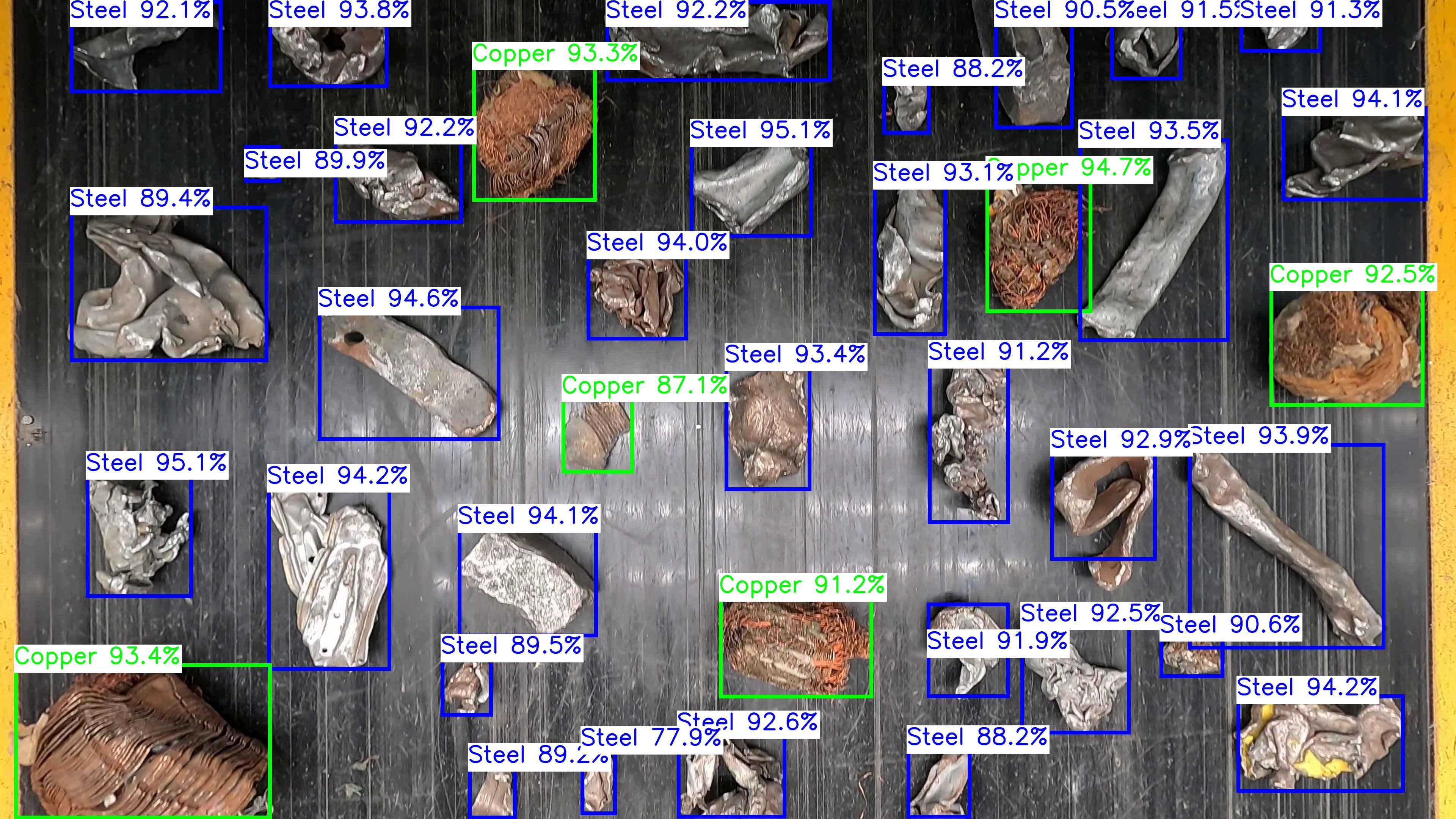} \\
    [1.5cm]
    
    \smash{\rotatebox[origin=c]{90}{\textbf{YOLO12n}}} &
    \includegraphics[width=0.32\textwidth, valign=c]{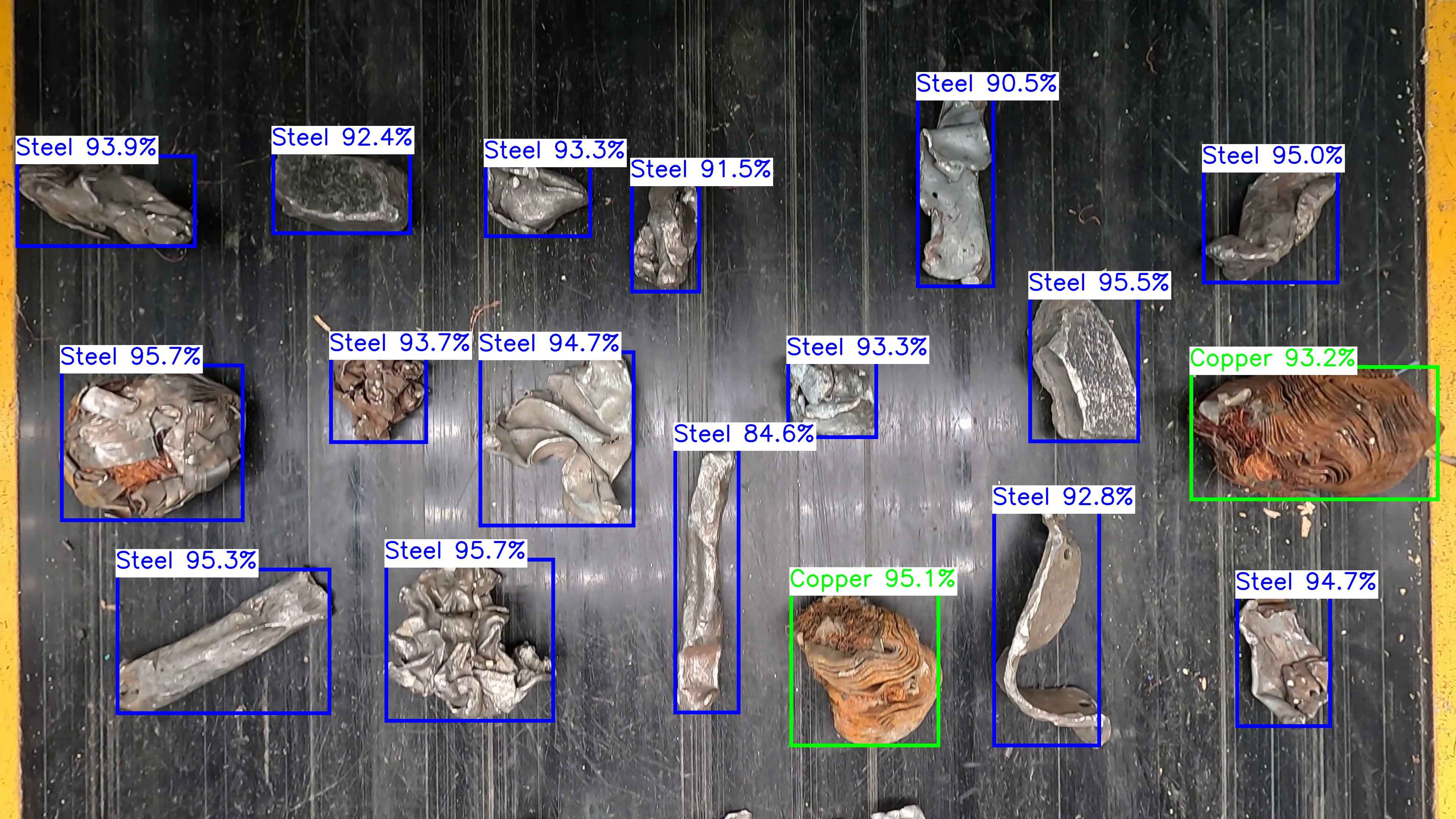} &
    \includegraphics[width=0.32\textwidth, valign=c]{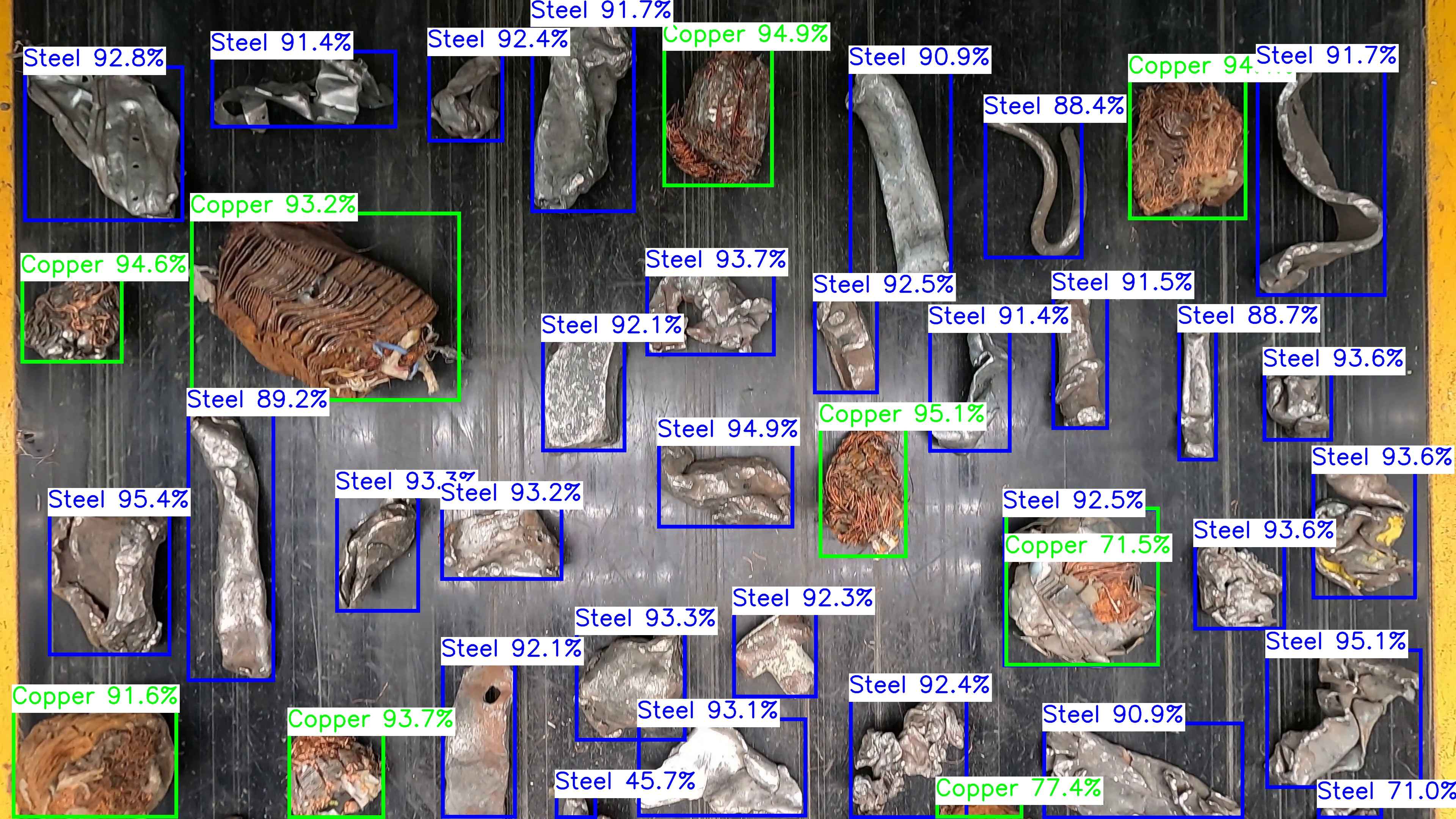} &
    \includegraphics[width=0.32\textwidth, valign=c]{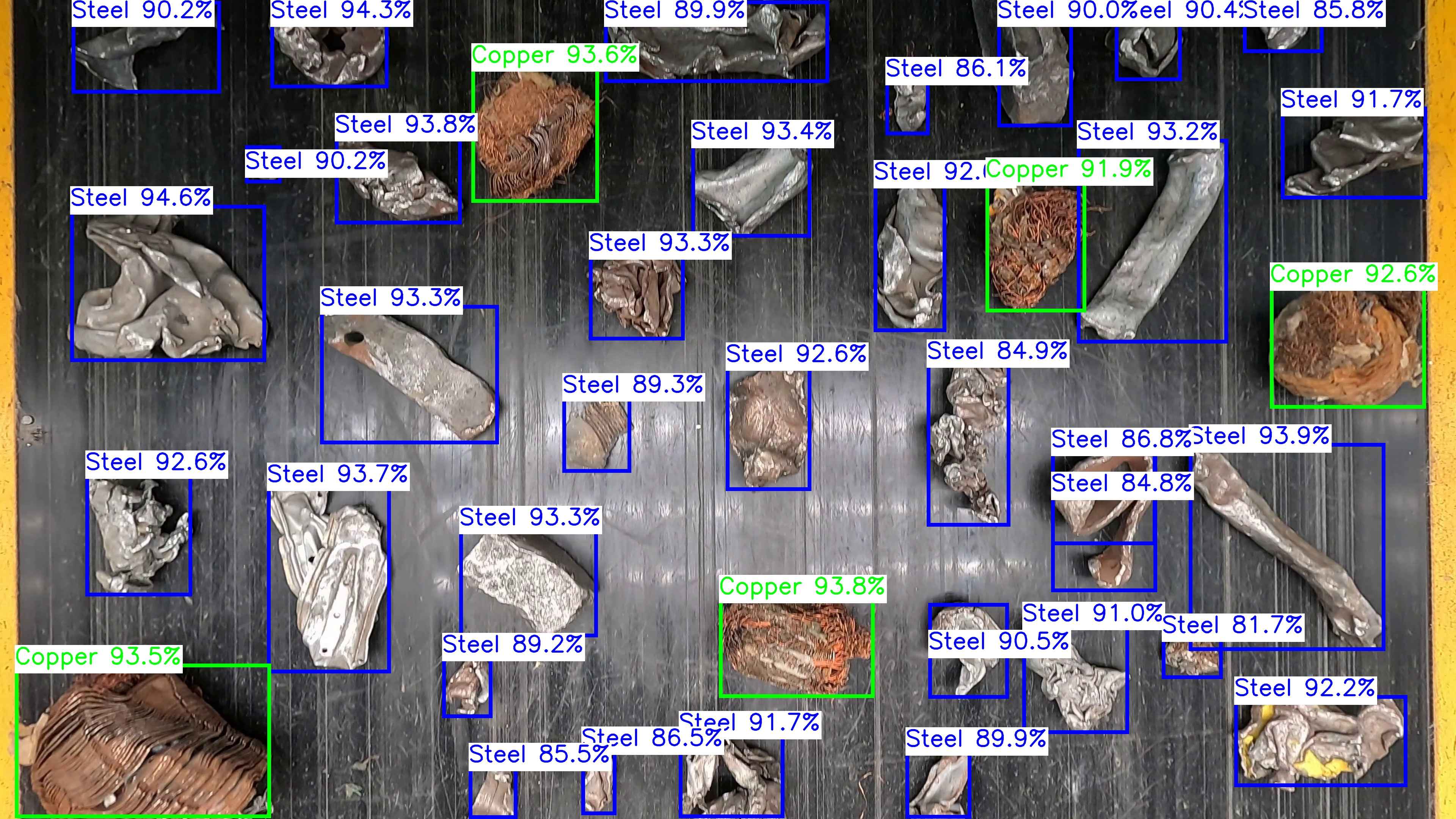} \\
    [1.5cm]
    
    \smash{\rotatebox[origin=c]{90}{\textbf{YOLO26n}}} &
    \includegraphics[width=0.32\textwidth, valign=c]{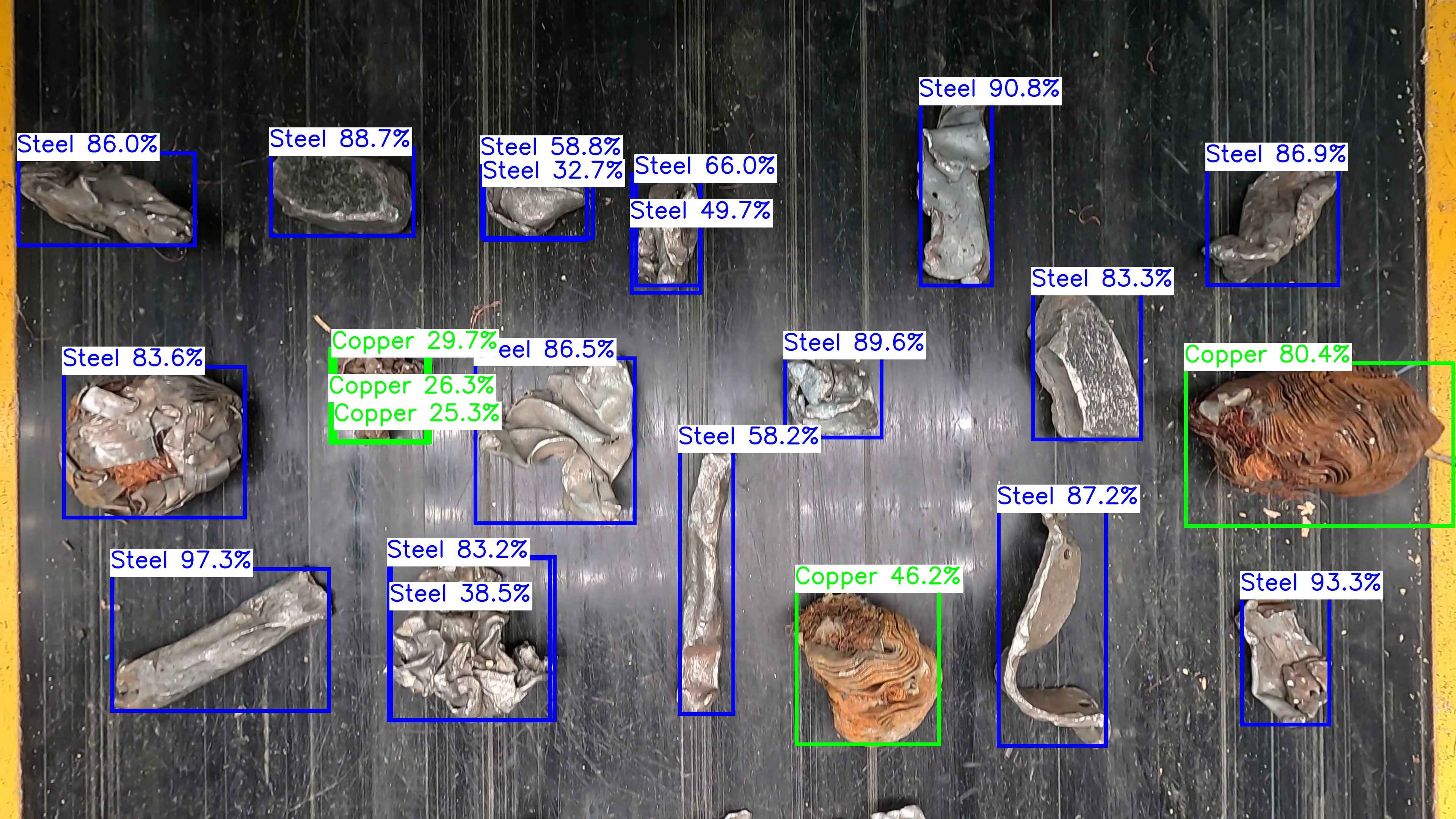} &
    \includegraphics[width=0.32\textwidth, valign=c]{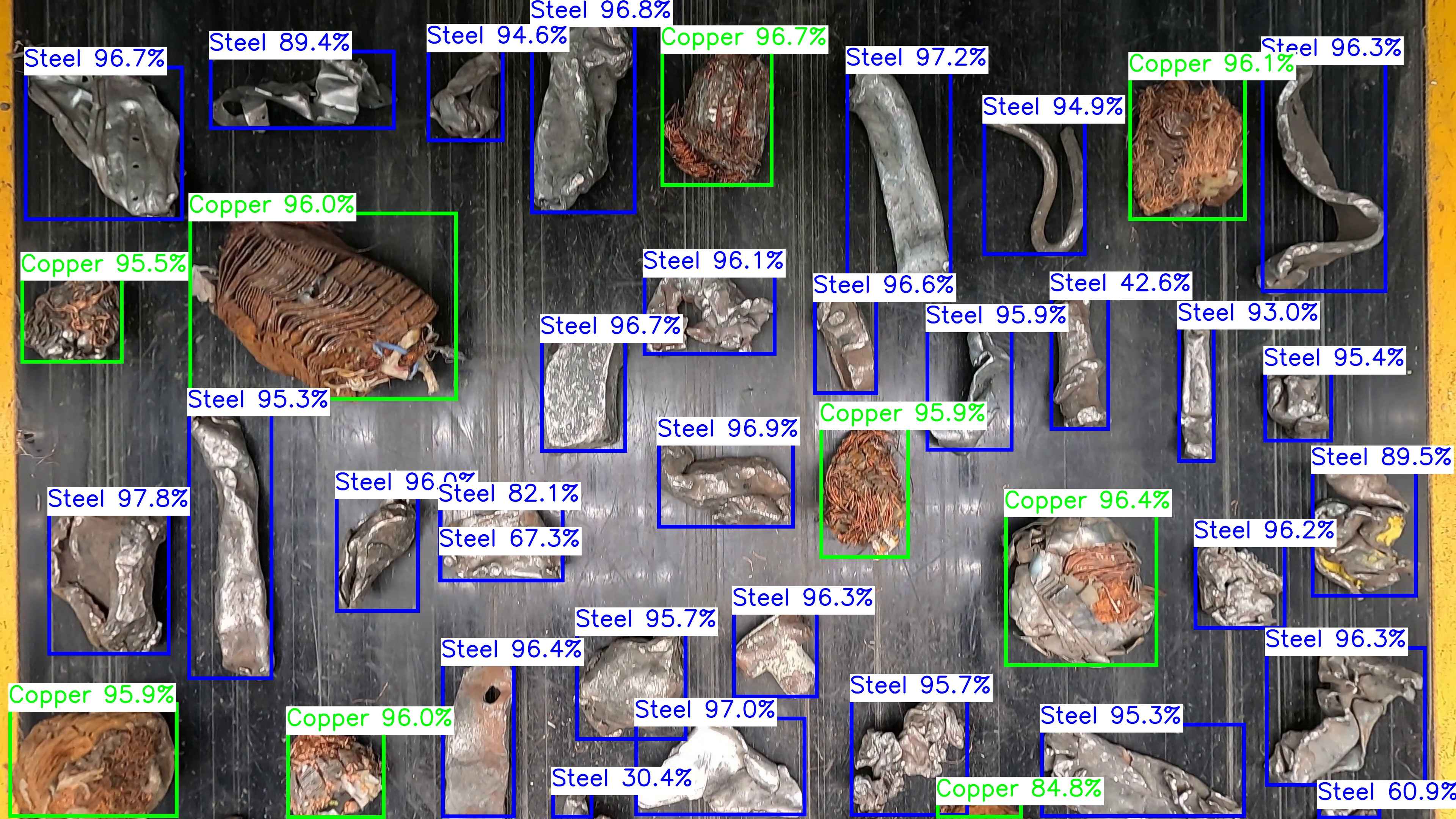} &
    \includegraphics[width=0.32\textwidth, valign=c]{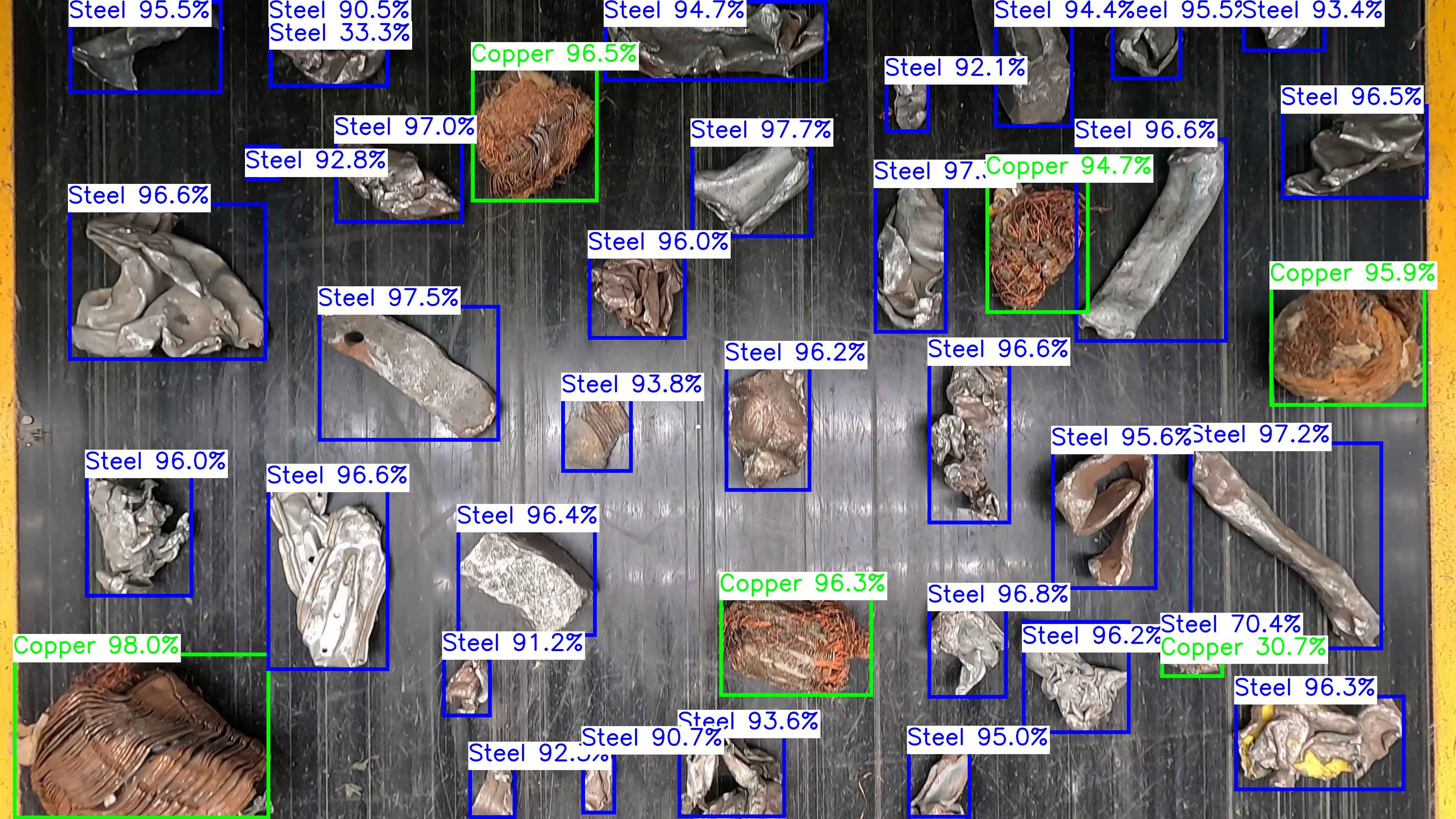} \\
    [1.5cm]
    
    \smash{\rotatebox[origin=c]{90}{\textbf{Mask R-CNN}}} &
    \includegraphics[width=0.32\textwidth, valign=c]{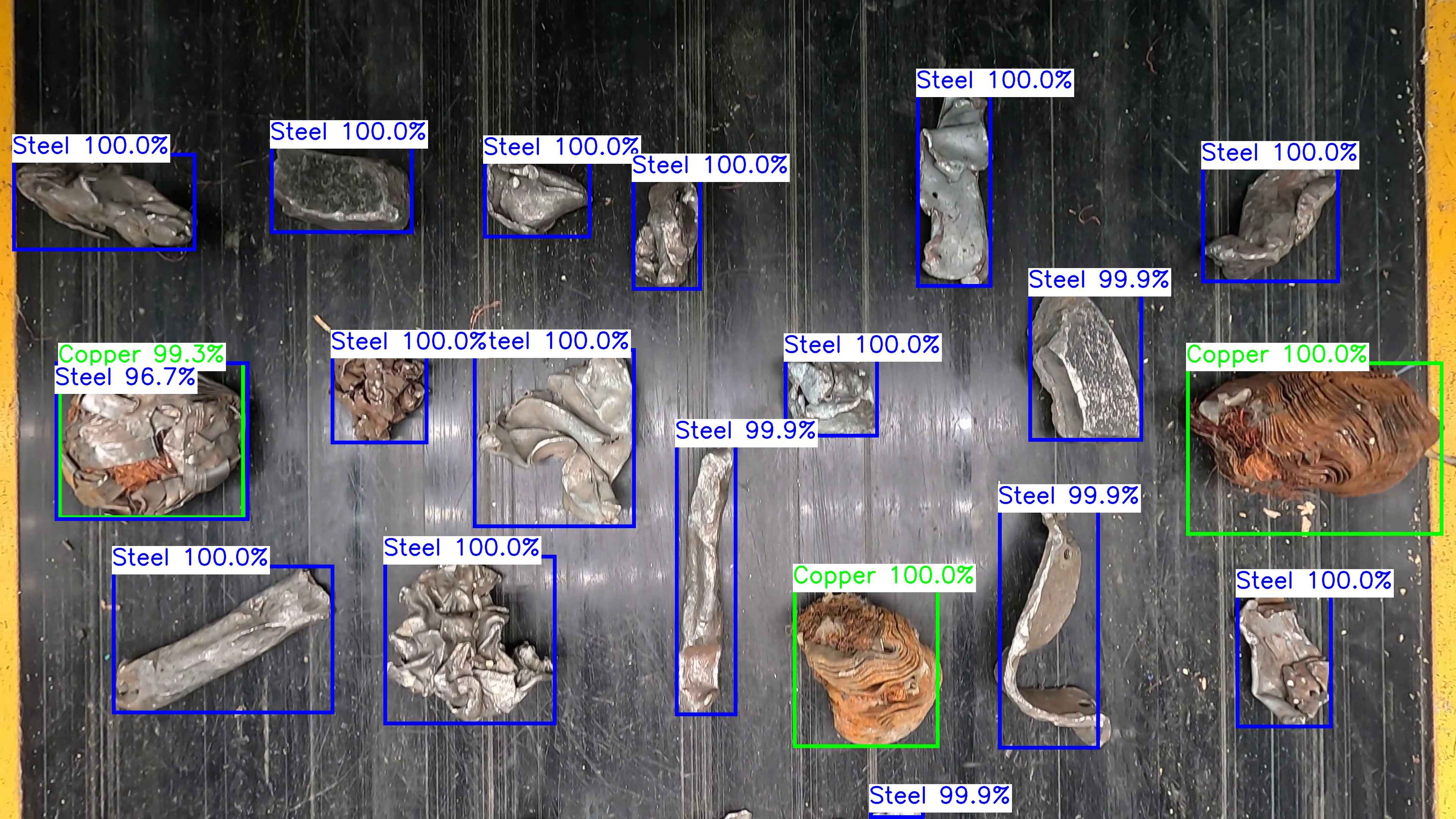} &
    \includegraphics[width=0.32\textwidth, valign=c]{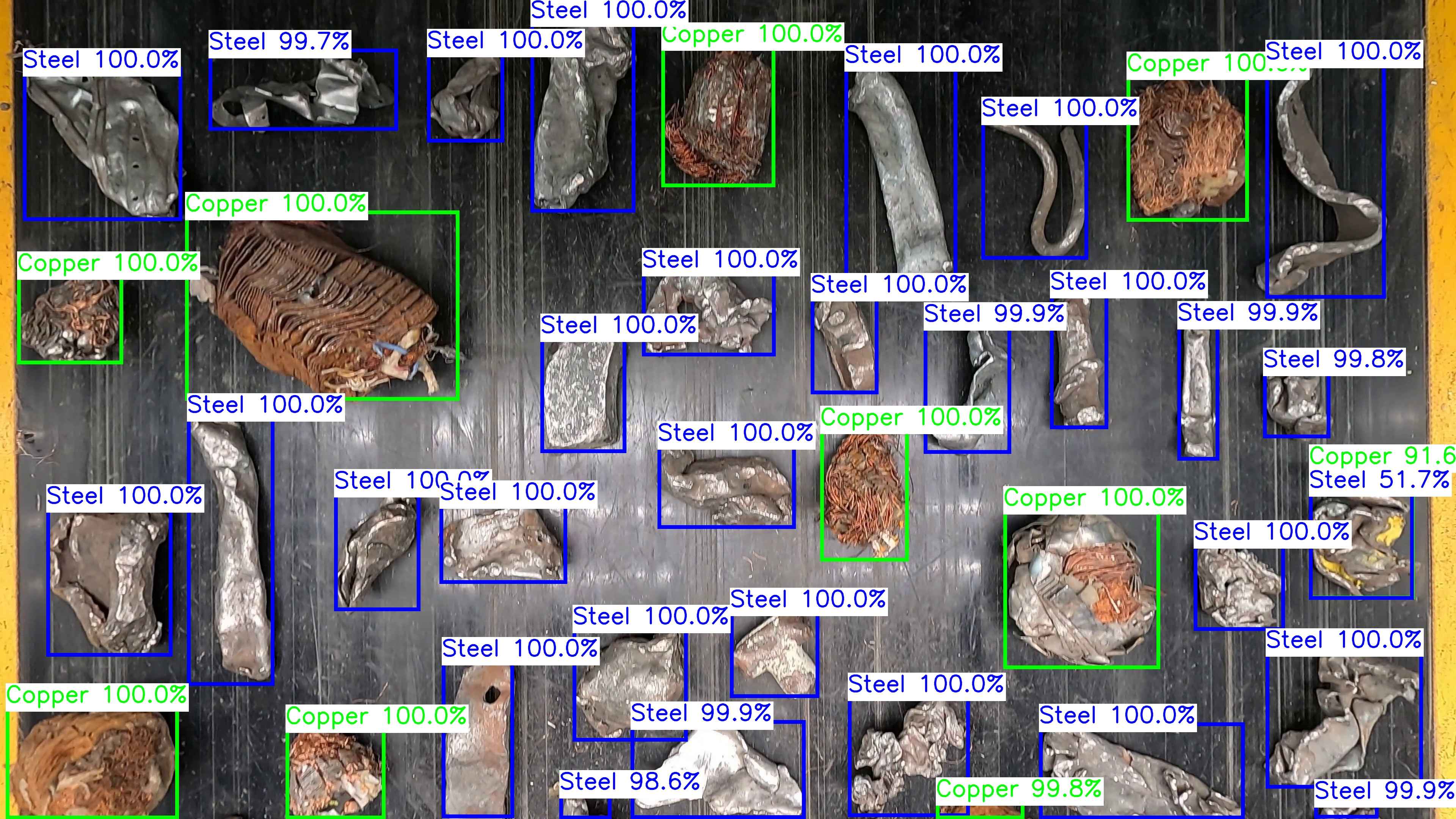} &
    \includegraphics[width=0.32\textwidth, valign=c]{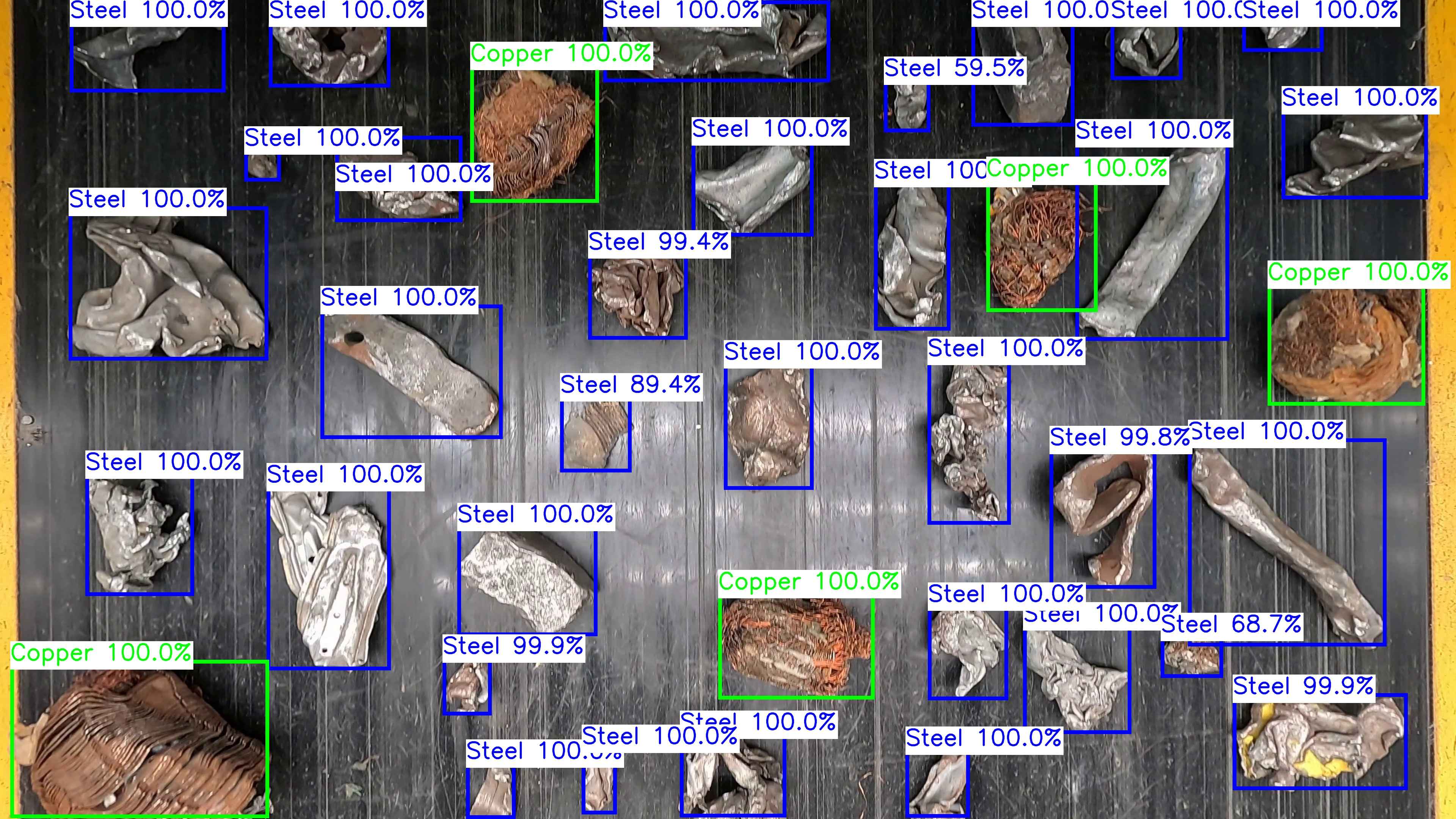} \\
    [1.1cm]
  \end{tabular}
  \caption{Visual results for bounding box detection. Green and blue boxes denote steel and copper objects, respectively. The top row shows ground truth annotations, while each column corresponds to a distinct sub-dataset.}
  \label{fig:vis_results_bb}
\end{figure}

\begin{figure}[t]
  \centering
  \begin{tabular}{c c c c}
    & \textbf{Subset a1} & \textbf{Subset a2} & \textbf{Subset a3} \\[0.15cm] 
    
    \smash{\rotatebox[origin=c]{90}{\textbf{Ground Truth}}} &
    \includegraphics[width=0.32\textwidth, valign=c]{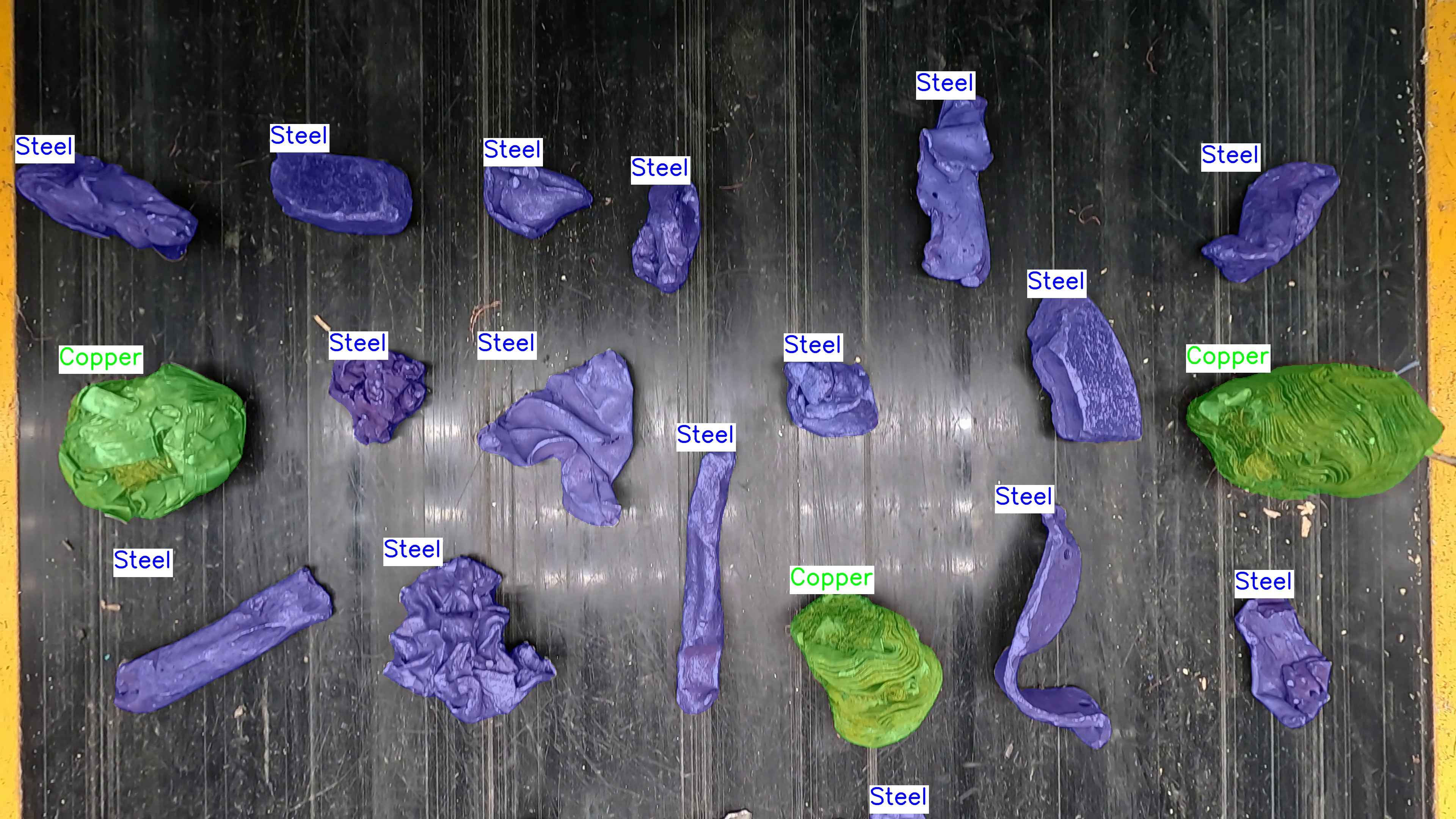} &
    \includegraphics[width=0.32\textwidth, valign=c]{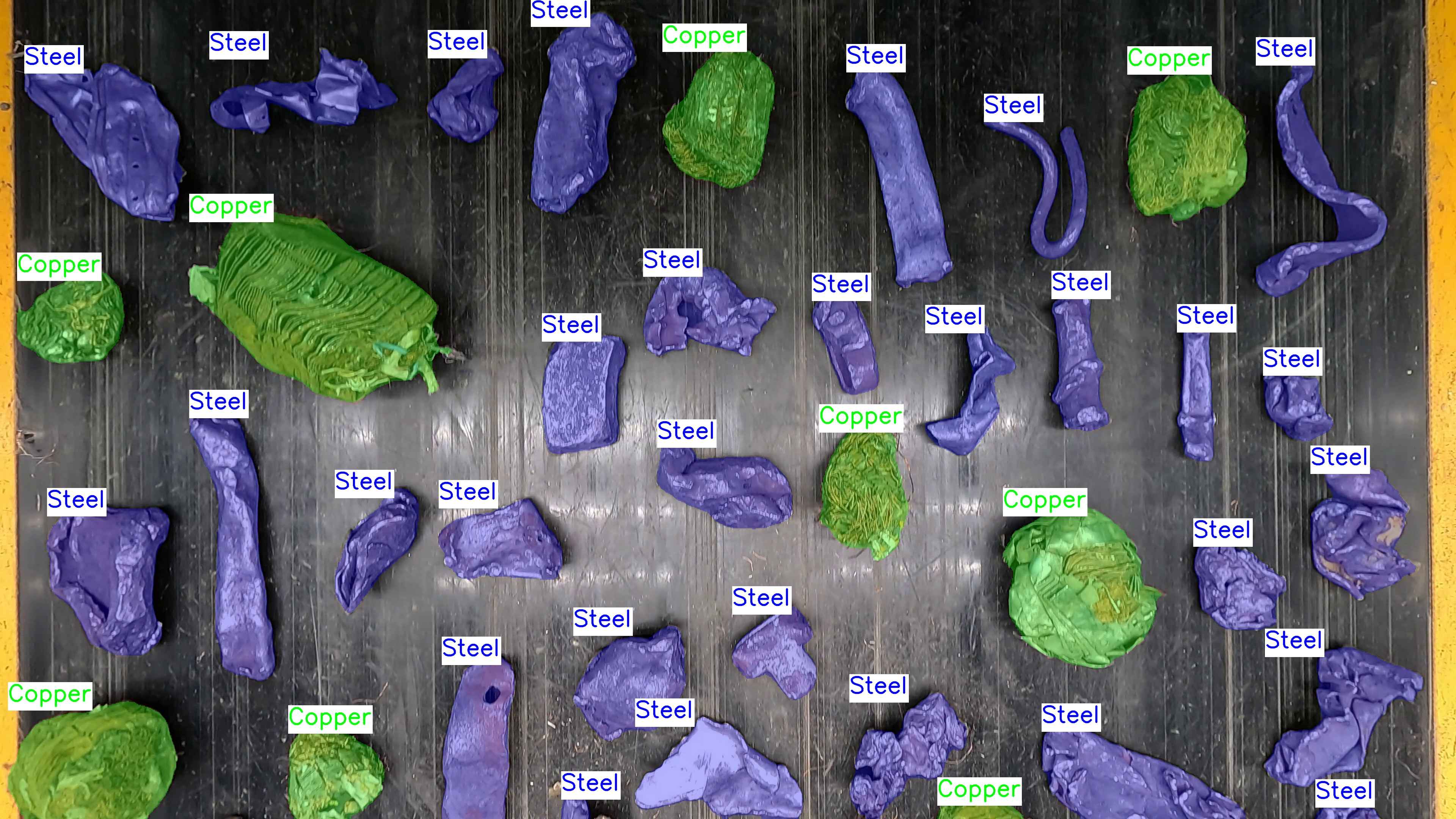} &
    \includegraphics[width=0.32\textwidth, valign=c]{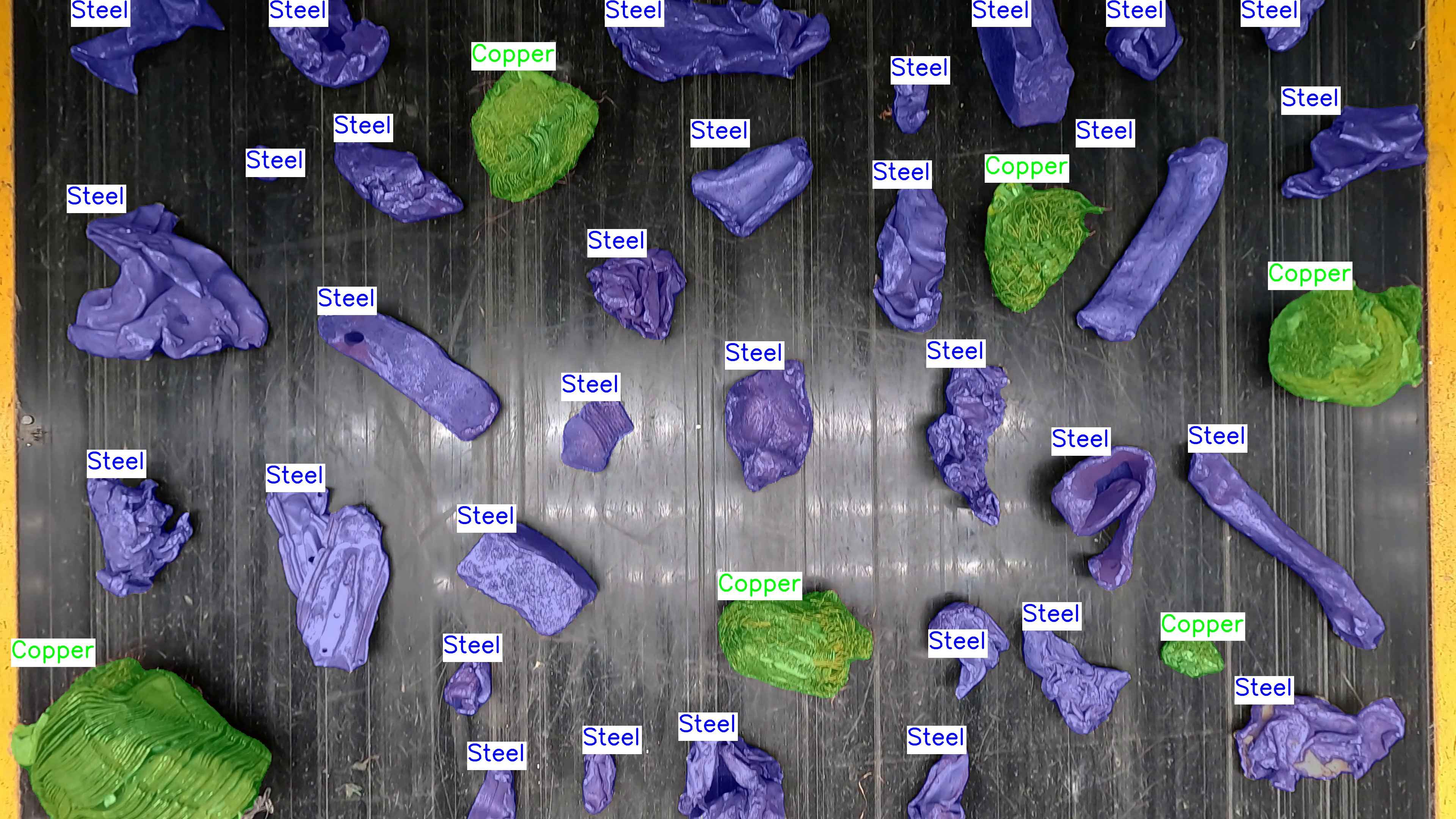} \\[1.5cm]
    
    \smash{\rotatebox[origin=c]{90}{\textbf{YOLOv8n}}} &
    \includegraphics[width=0.32\textwidth, valign=c]{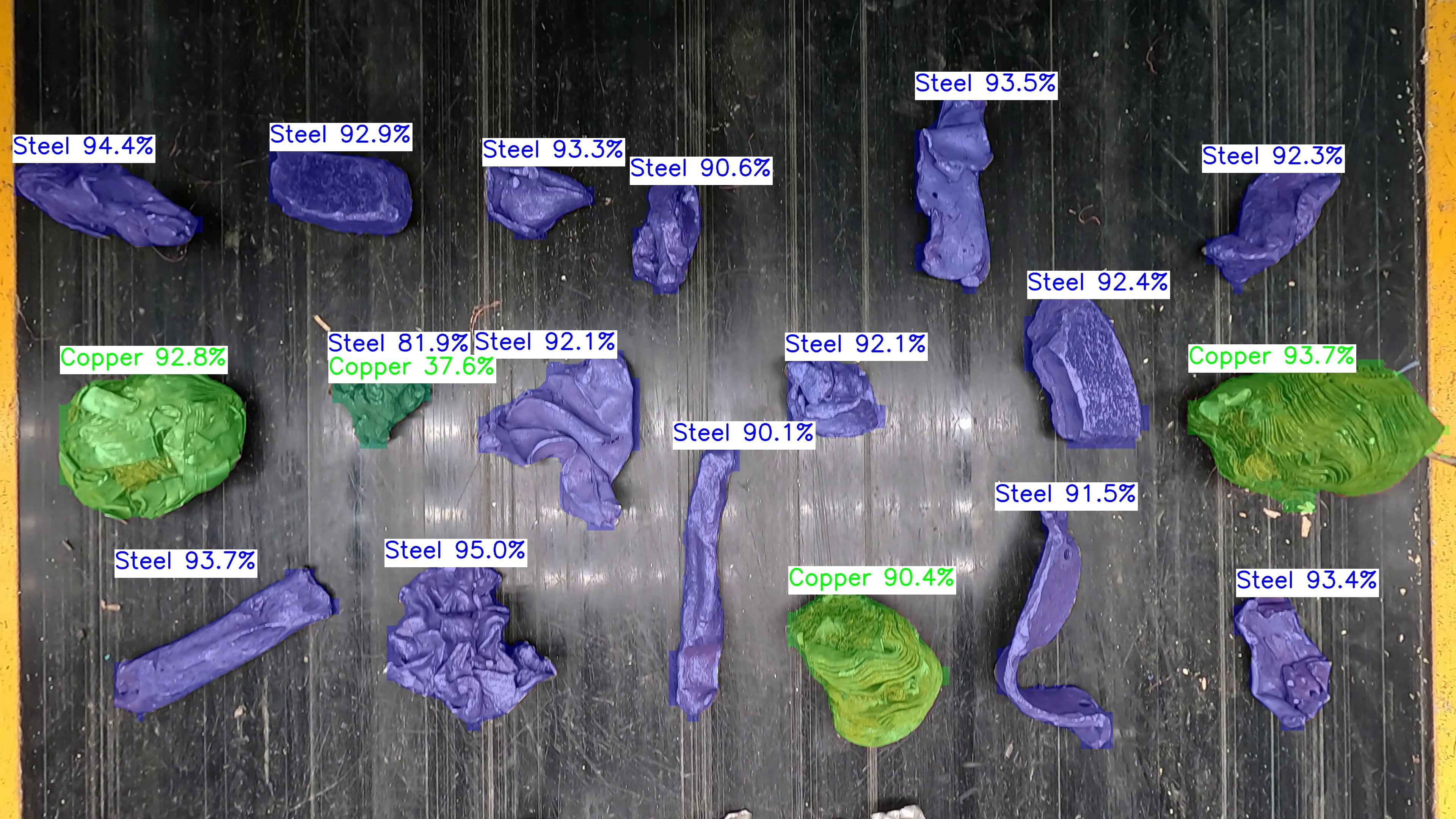} &
    \includegraphics[width=0.32\textwidth, valign=c]{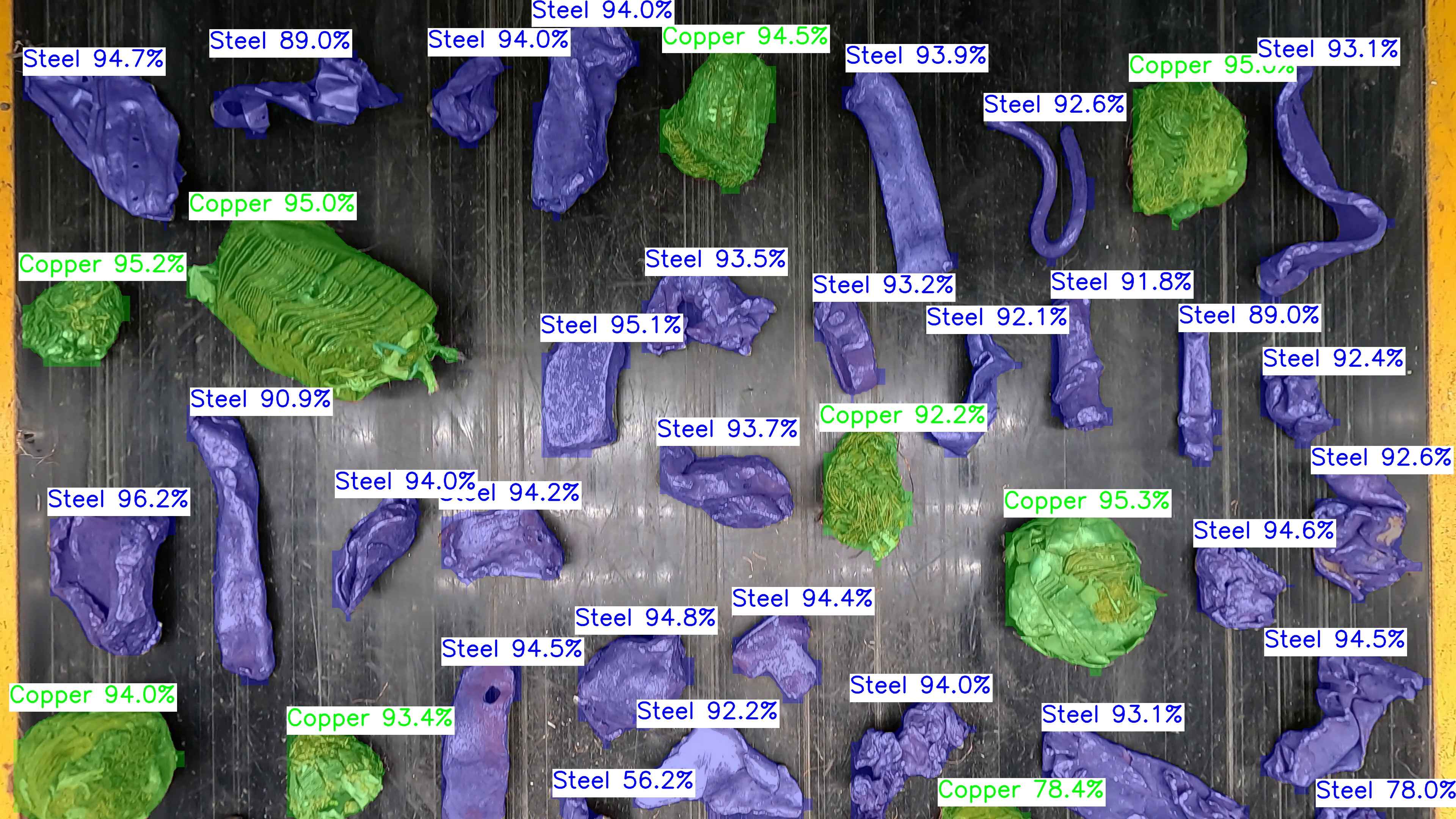} &
    \includegraphics[width=0.32\textwidth, valign=c]{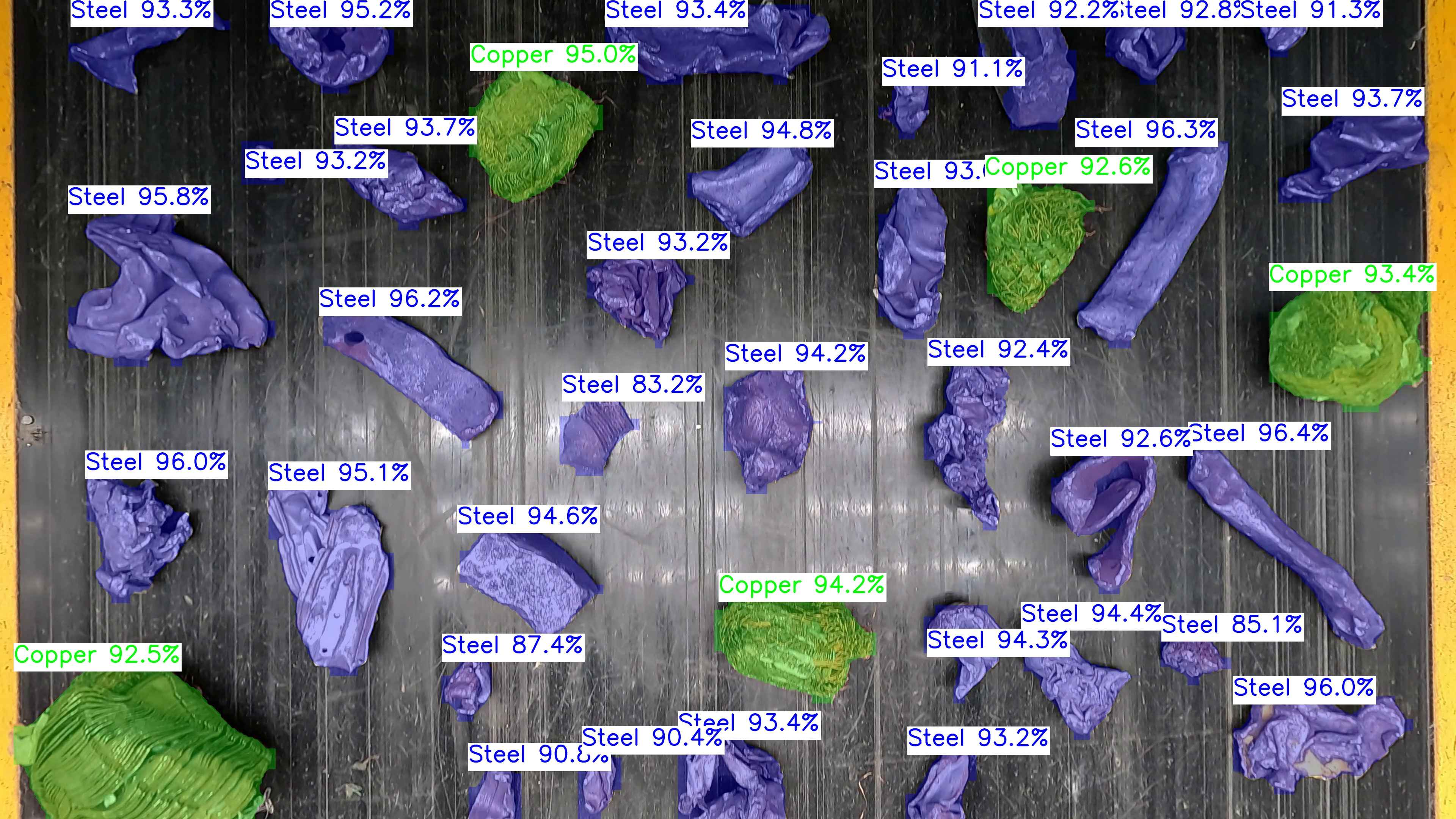} \\
    [1.5cm]
    
    \smash{\rotatebox[origin=c]{90}{\textbf{YOLO11n}}} &
    \includegraphics[width=0.32\textwidth, valign=c]{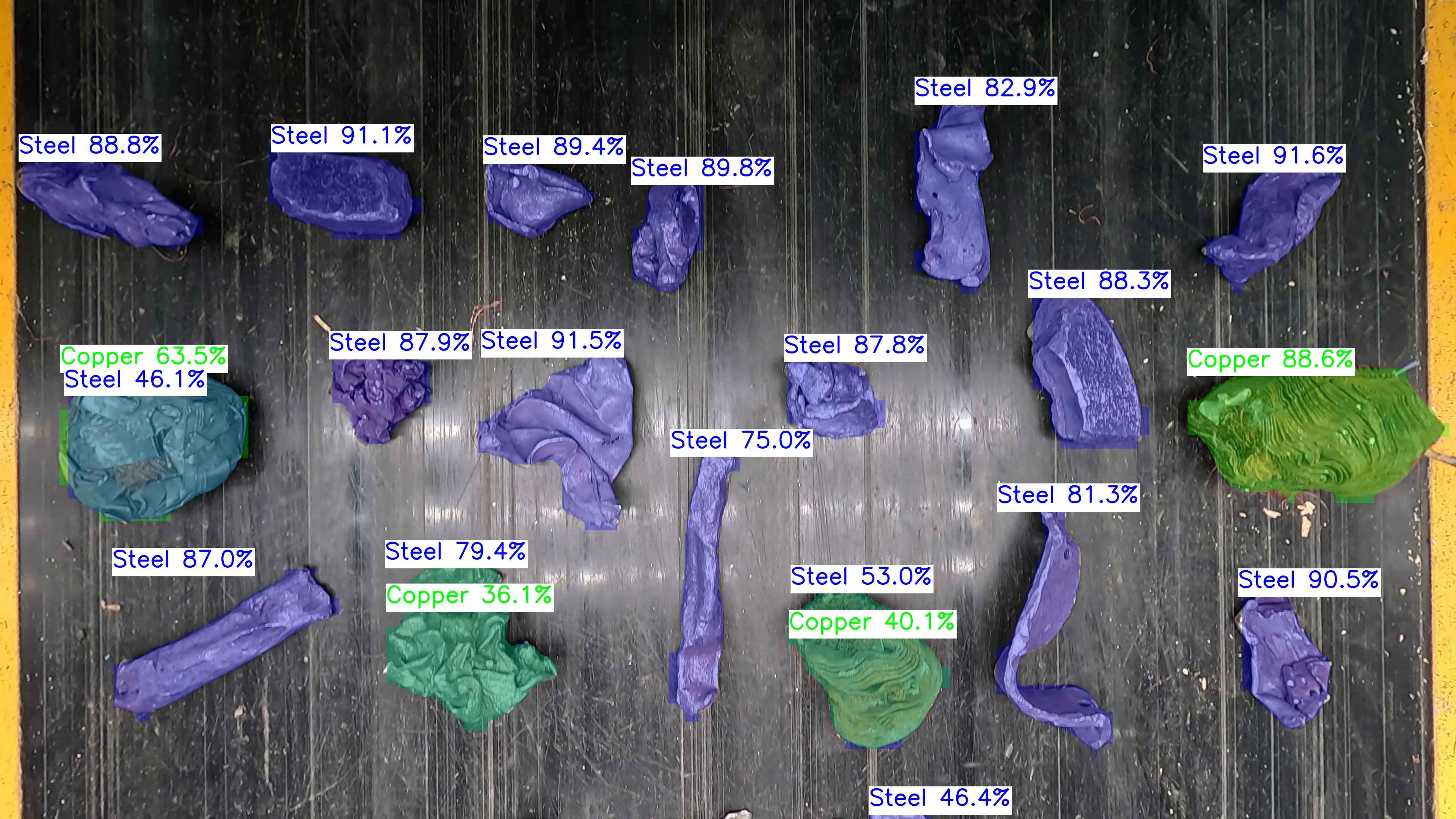} &
    \includegraphics[width=0.32\textwidth, valign=c]{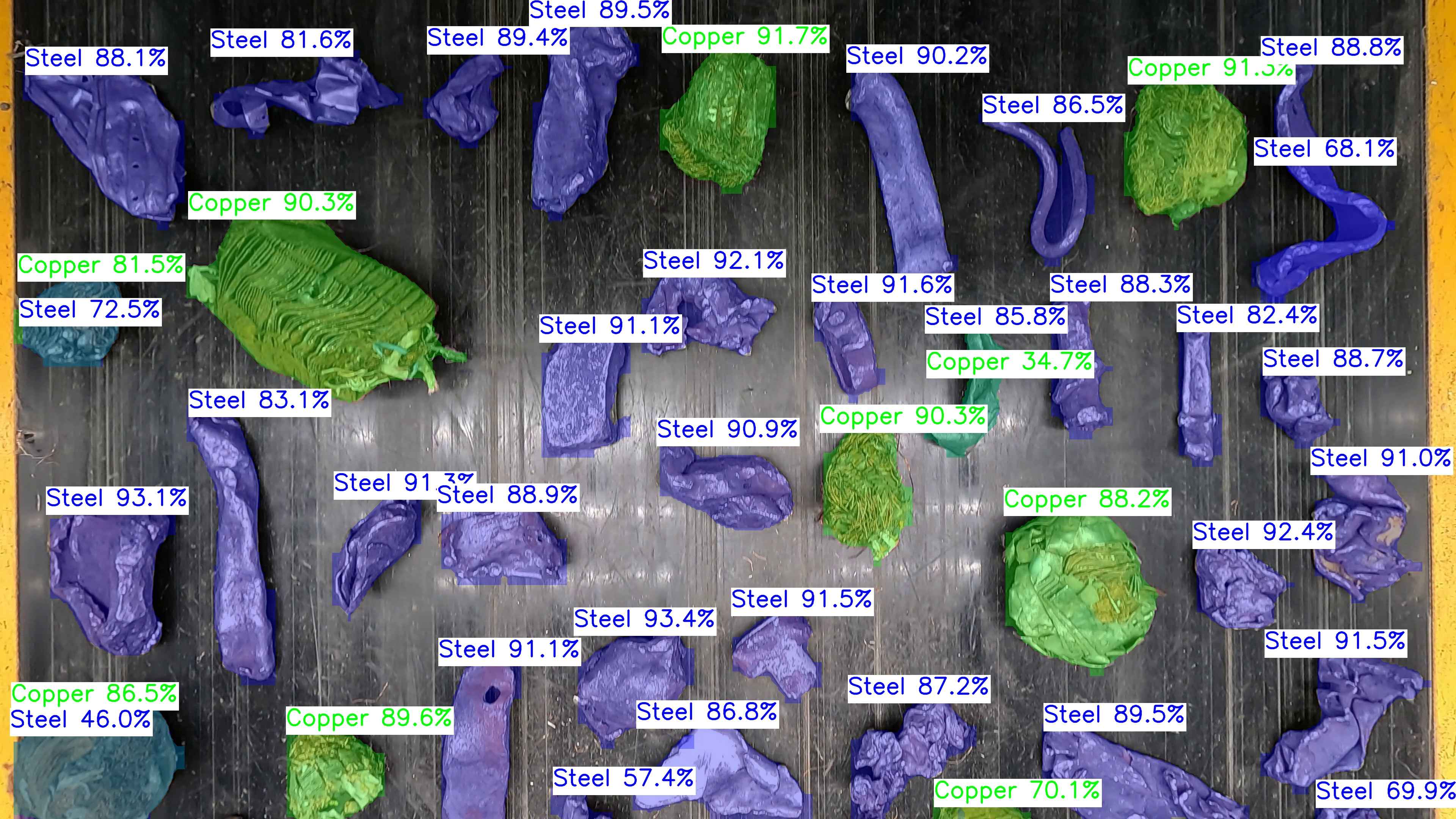} &
    \includegraphics[width=0.32\textwidth, valign=c]{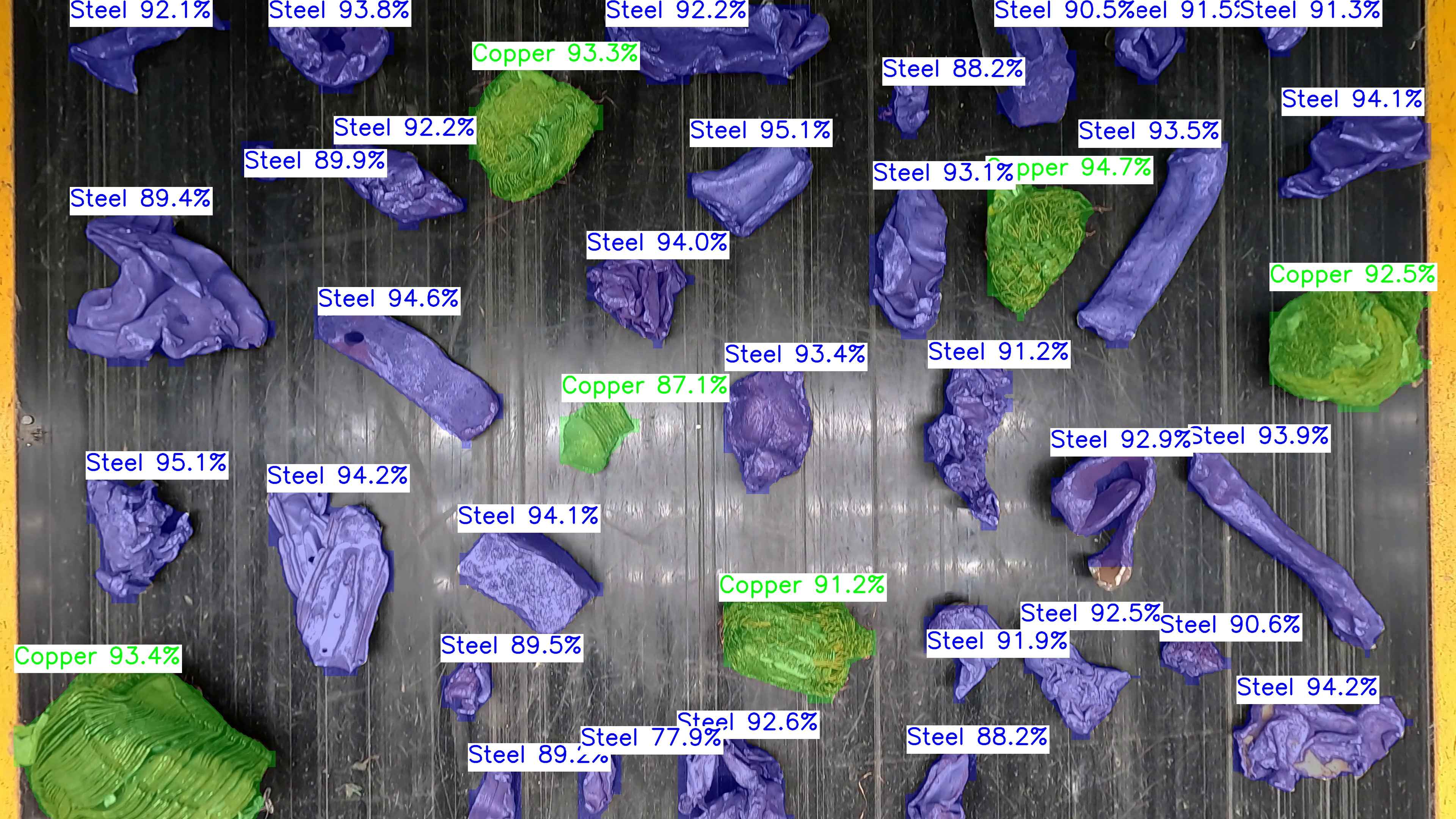} \\
    [1.5cm]
    
    \smash{\rotatebox[origin=c]{90}{\textbf{YOLO12n}}} &
    \includegraphics[width=0.32\textwidth, valign=c]{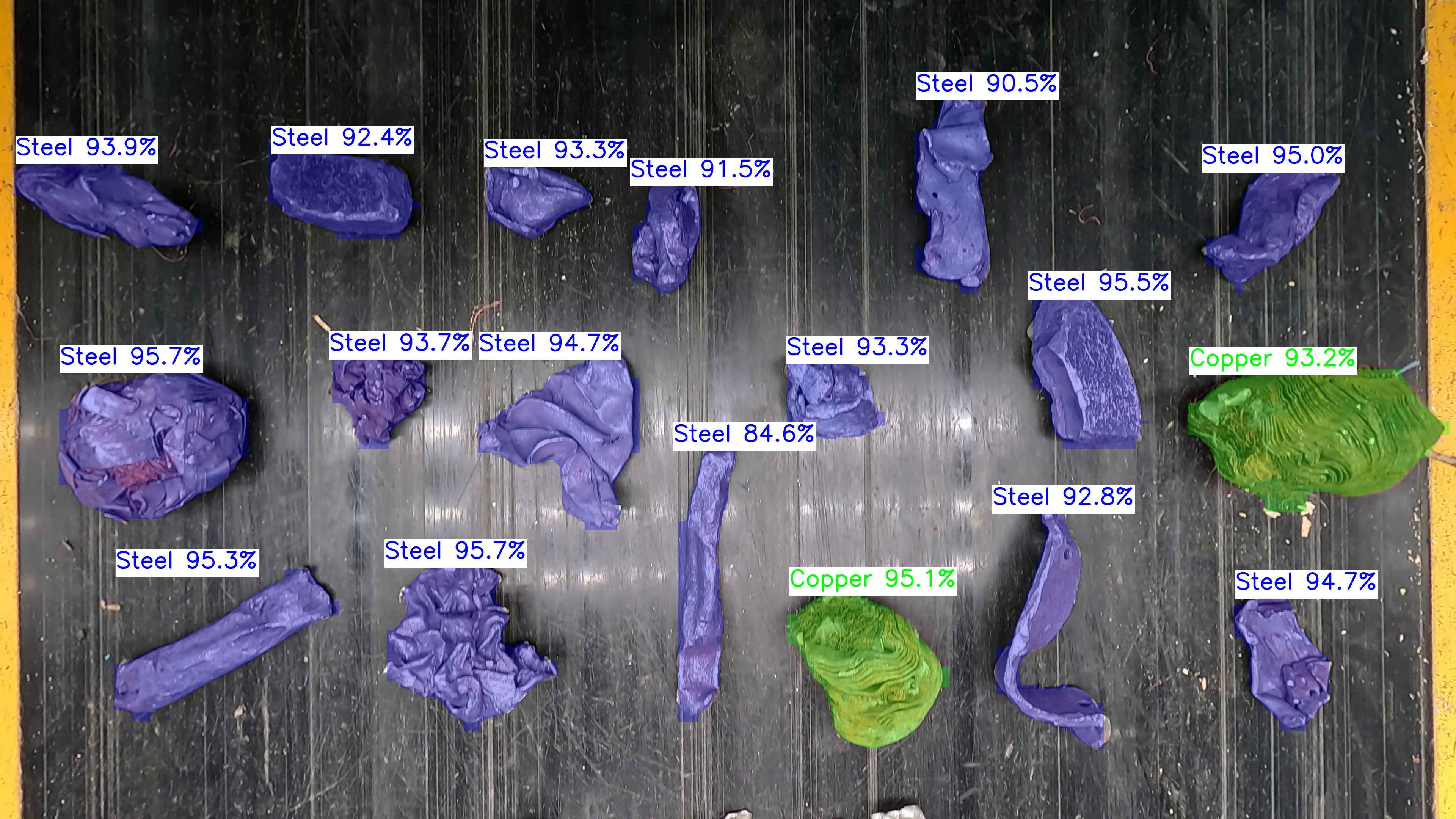} &
    \includegraphics[width=0.32\textwidth, valign=c]{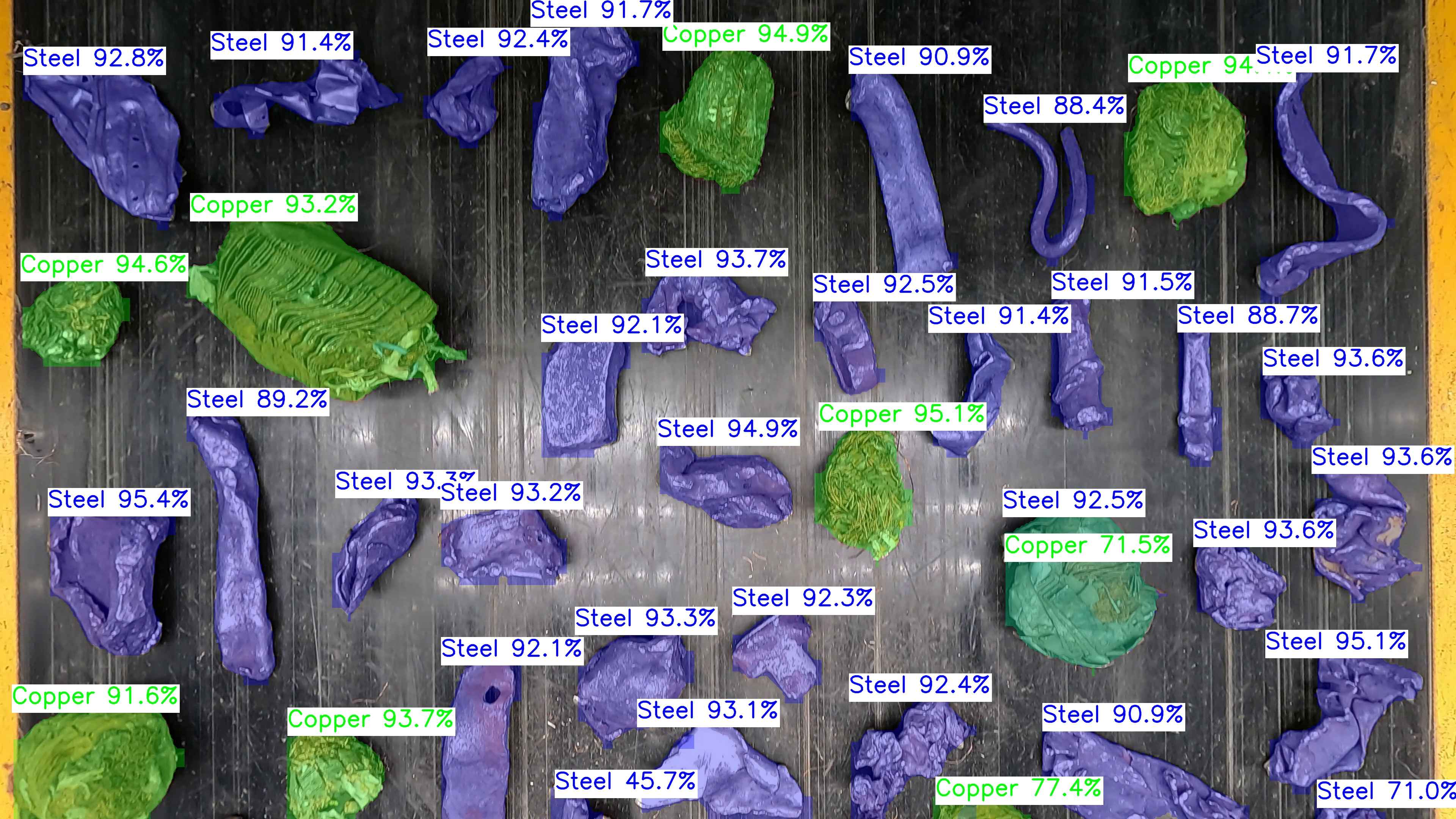} &
    \includegraphics[width=0.32\textwidth, valign=c]{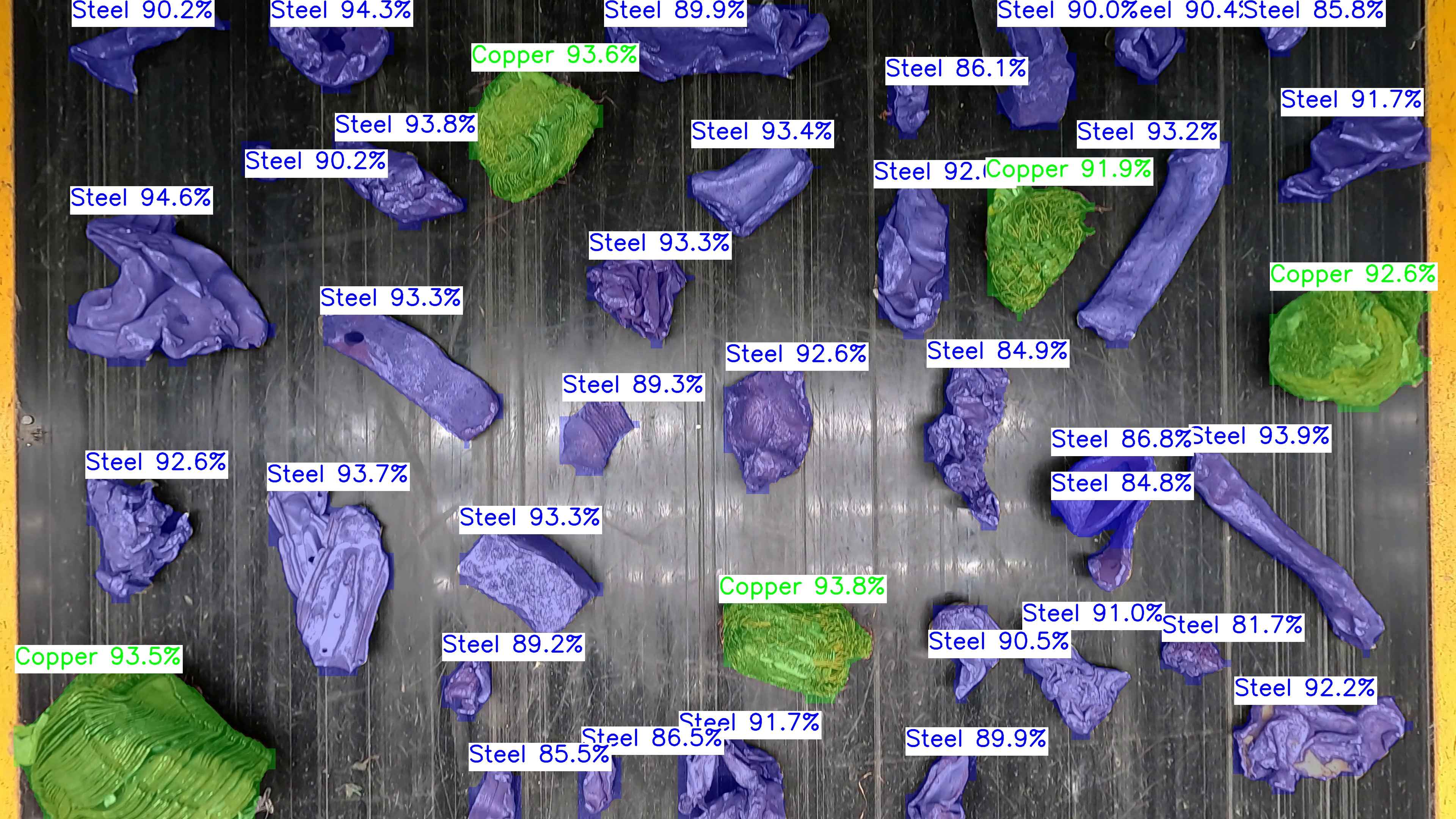} \\
    [1.5cm]
    
    \smash{\rotatebox[origin=c]{90}{\textbf{YOLO26n}}} &
    \includegraphics[width=0.32\textwidth, valign=c]{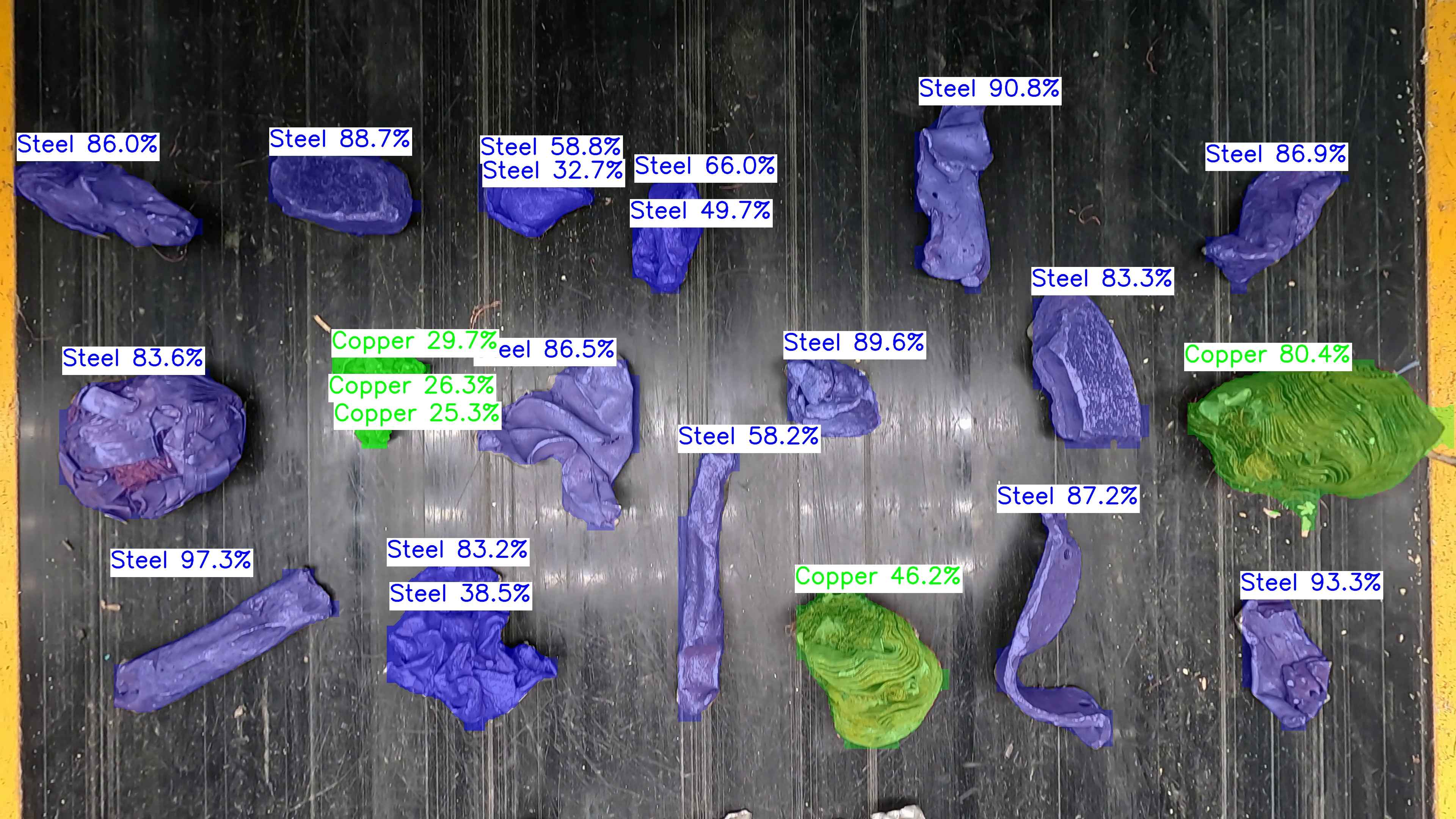} &
    \includegraphics[width=0.32\textwidth, valign=c]{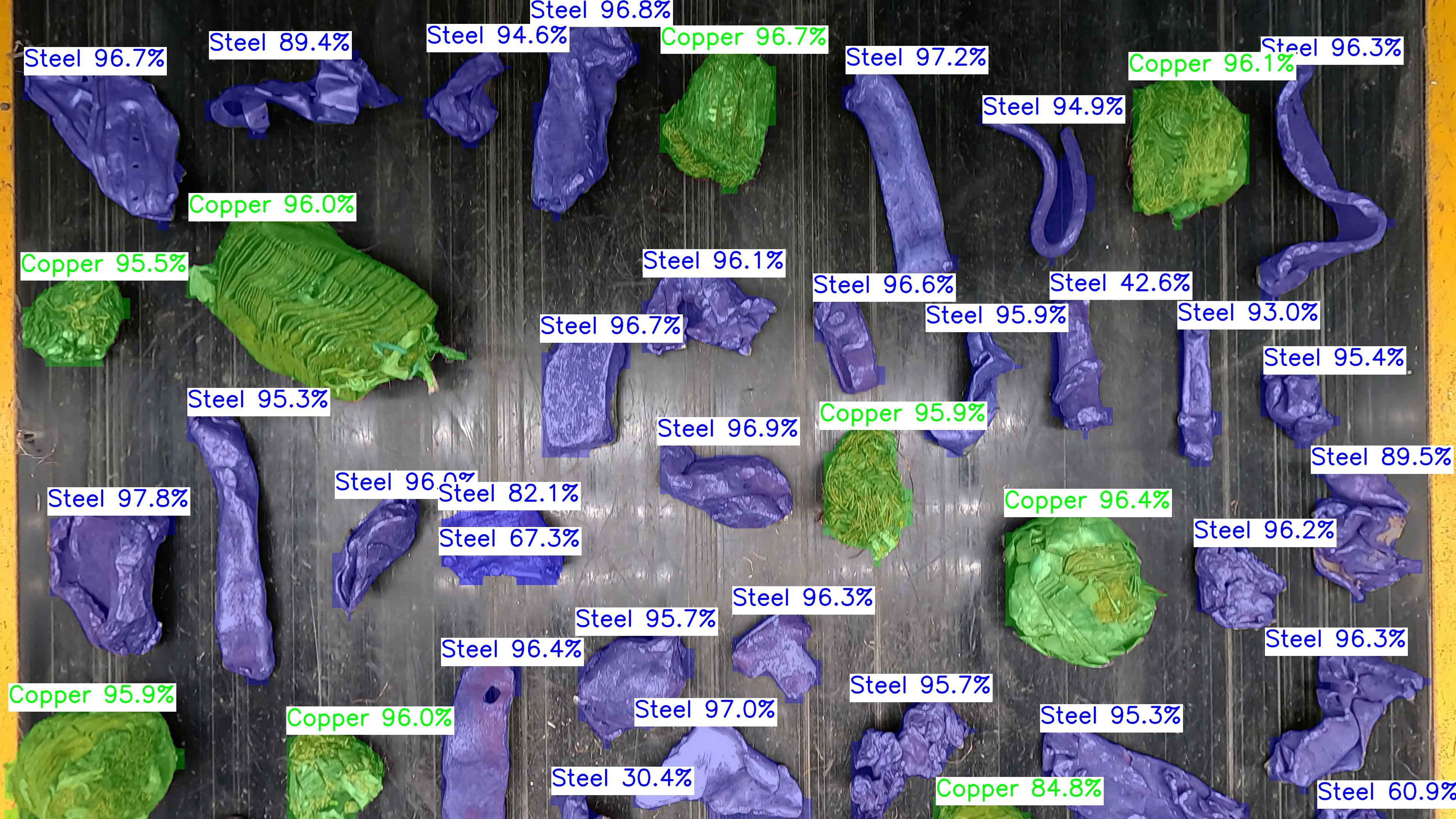} &
    \includegraphics[width=0.32\textwidth, valign=c]{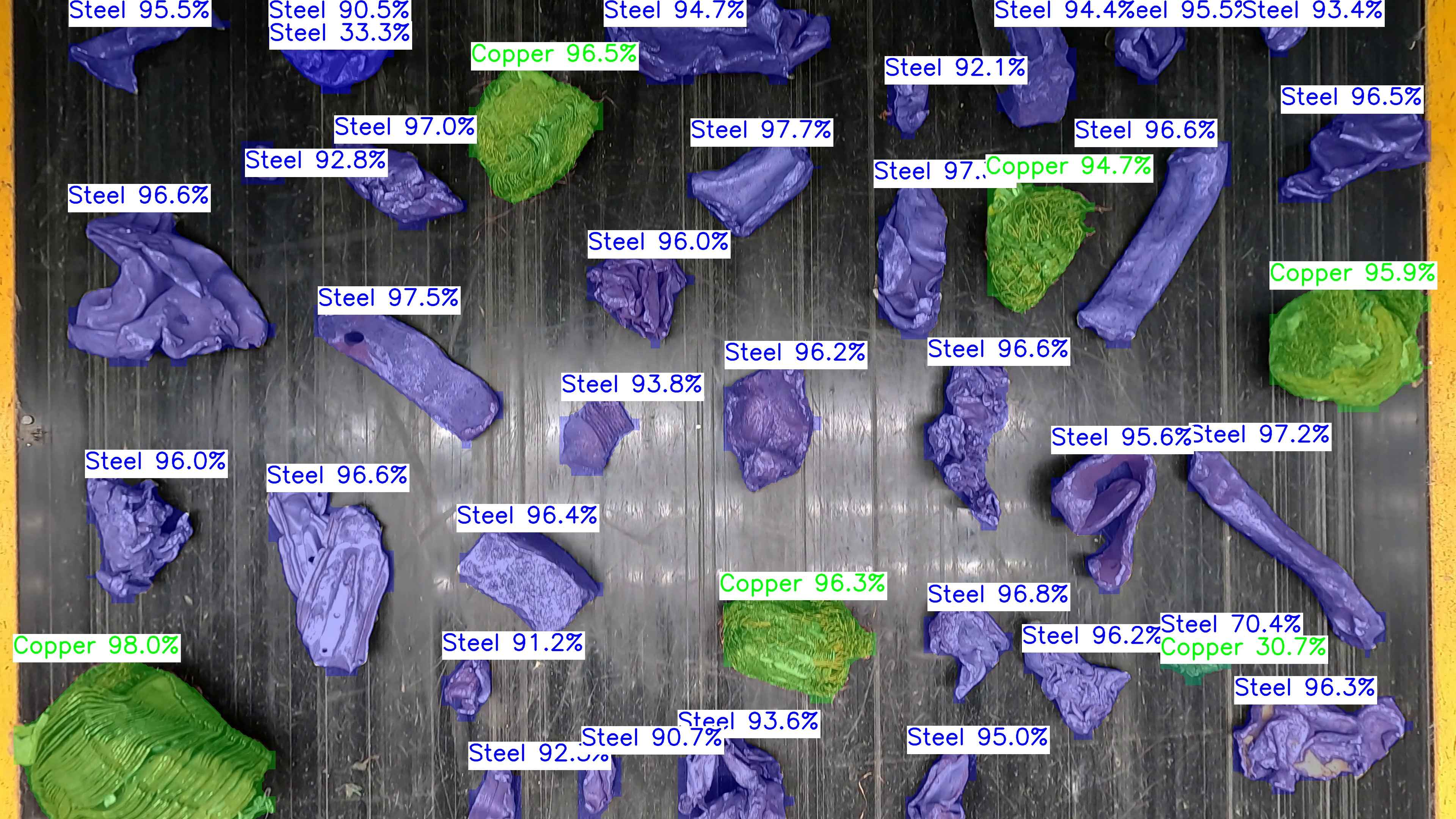} \\
    [1.5cm]
    
    \smash{\rotatebox[origin=c]{90}{\textbf{Mask R-CNN}}} &
    \includegraphics[width=0.32\textwidth, valign=c]{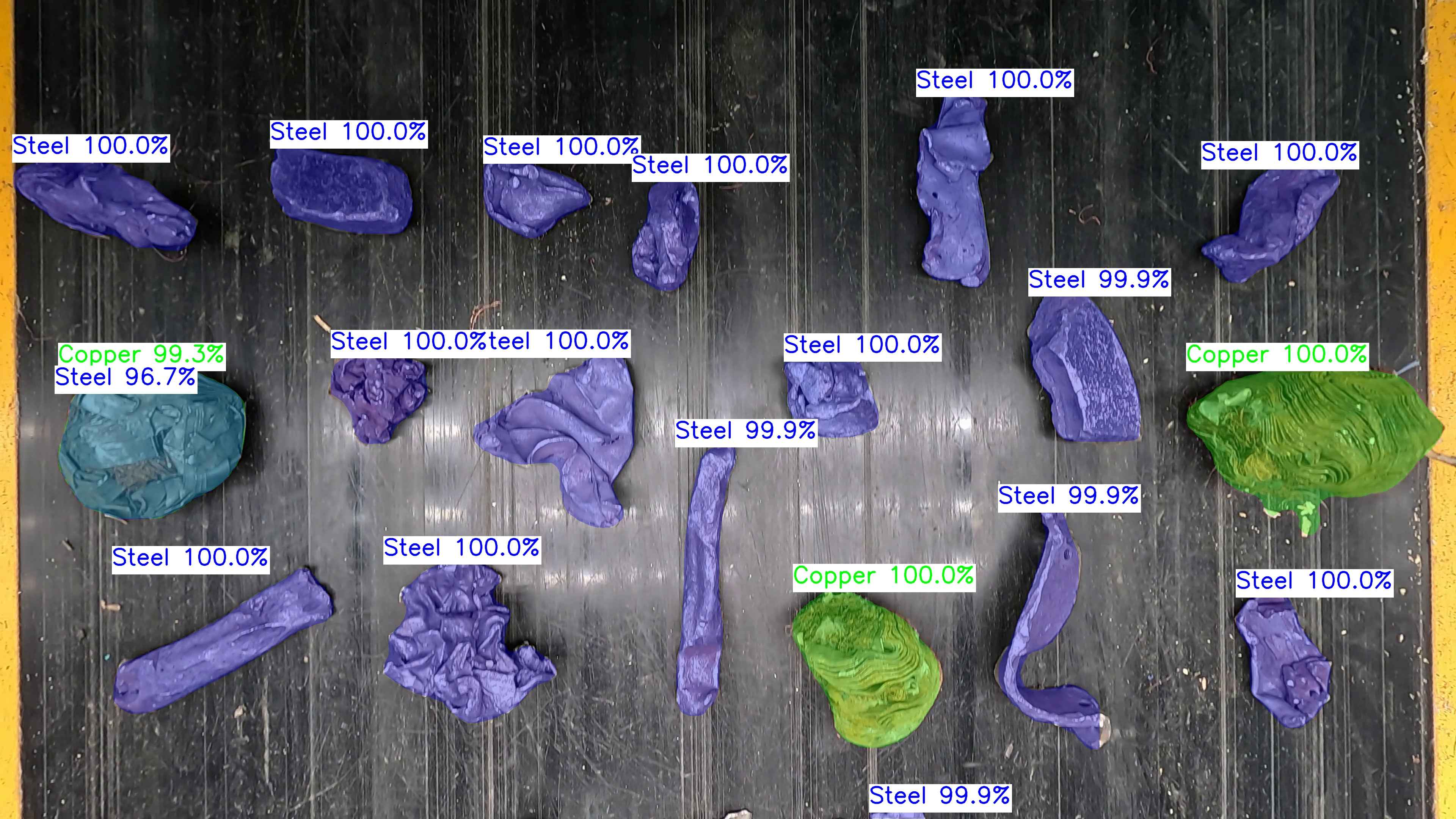} &
    \includegraphics[width=0.32\textwidth, valign=c]{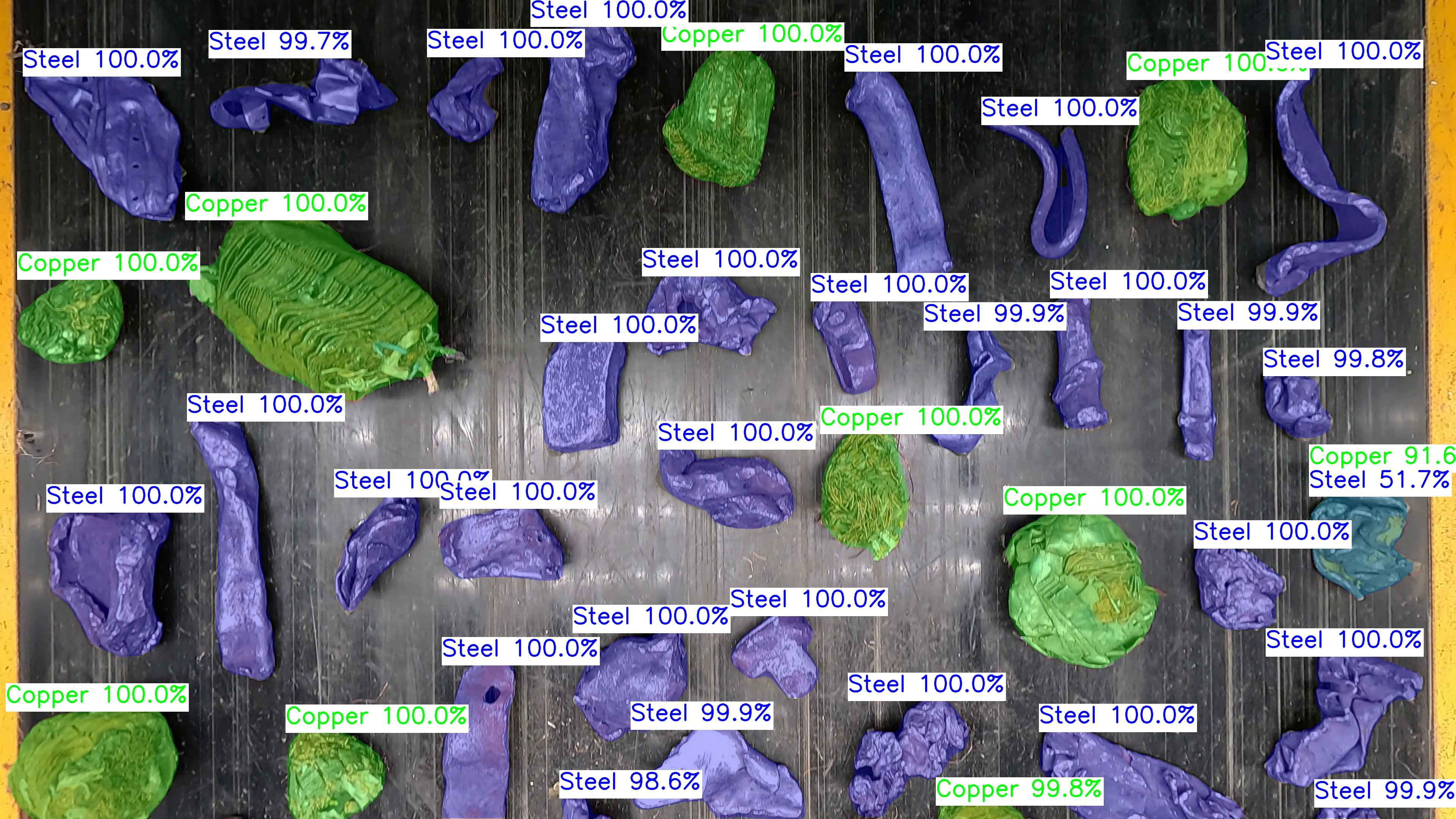} &
    \includegraphics[width=0.32\textwidth, valign=c]{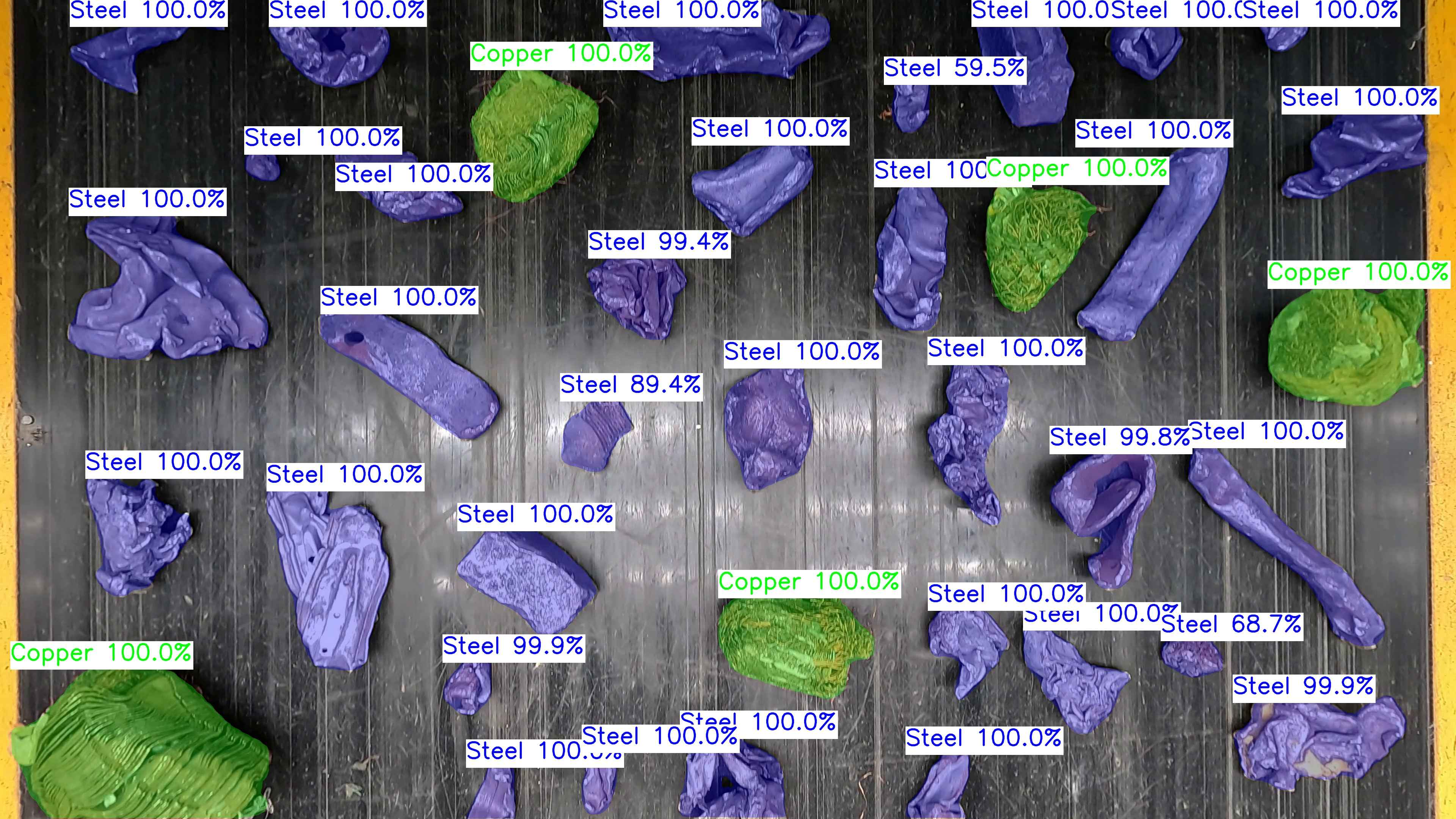} \\
    [1.1cm]
  \end{tabular}
  \caption{Visual results for instance segmentation. Green and blue segments denote steel and copper objects, respectively. The top row shows ground truth annotations, while each column corresponds to a distinct sub-dataset.}
  \label{fig:vis_results_seg}
\end{figure}

\section*{Code Availability}
The dataset is freely available as described in \cite{neubauer2026steelds}. 
The custom code to generate or process these data as well as the source code of the validation model can be found in the following GitHub repository: \url{https://github.com/MelanieNeubauer/SteelDS_Baseline.git}.

\section*{Acknowledgements}
The authors thank Scholz Austria GmbH in Laxenburg, and specifically Walter Martinelli and Yves Radmann, providing scrap test specimens and supporting the research activities. 

This work was carried out within the projects KIRAMET – KI-based Recycling of Metal Compound Waste (FO999899661) and MUTAVIA (FO999922732), funded by the Austrian Research Promotion Agency (FFG) under the programme of the Federal Ministry for Climate Action, Environment, Energy, Mobility, Innovation and Technology (BMK). 
Additional support was provided by the Steirische Wirtschaftsförderungsgesellschaft mbH (SFG) within the project BaKIRoS (1000077791).

\section*{Author Contributions}
Melanie Neubauer is the primary author and did the sample collection, planning of work, execution of work, data annotation, data analysis, preparation and writing of the manuscript. 
Gerald Koinig contributed to the sample collection.
Christian Rauch, Alexia Tischberger-Aldrian and Roland Pomberger contributed to the review of the manuscript.
Elmar Rueckert contributed to the supervision and review of the manuscript.

\section*{Competing Interests}
The authors declare no competing interests.

\bibliography{sn-bibliography}

\end{document}